\pdfoutput=1
\documentclass[10pt,logo,copyright]{nvidiatechreport}
\usepackage[authoryear,sort&compress,round]{natbib}

\usepackage[utf8]{inputenc} 
\usepackage[T1]{fontenc}    

\usepackage{parskip}        
\usepackage{url}            
\usepackage{hyperref}       
\usepackage{booktabs}       
\usepackage{amsfonts}       
\usepackage{nicefrac}       
\usepackage{microtype}      
\usepackage{xcolor}         
\usepackage[dvipsnames]{xcolor} 
\usepackage{graphicx}
\usepackage{animate}        
\usepackage{subcaption}
\usepackage{tabularx}
\usepackage{makecell}
\usepackage{adjustbox}
\usepackage{setspace}
\newcolumntype{M}[1]{>{\centering\arraybackslash}m{#1}}
\usepackage{float}
\usepackage{tikz}
\usetikzlibrary{positioning,shapes,arrows}
\usepackage{amsmath,amsfonts,bm, bbm,leftindex}
\usepackage{multirow}
\usepackage{comment}
\usepackage{gensymb}
\usepackage{lipsum}
\usetikzlibrary{arrows.meta, positioning, fit}
\usepackage[para]{threeparttable}
\usepackage{tikz}
\usetikzlibrary{tikzmark}

\def\eqref#1{equation~\ref{#1}}

\def\1{\bm{1}}

\DeclareMathAlphabet{\mathsfit}{\encodingdefault}{\sfdefault}{m}{sl}
\SetMathAlphabet{\mathsfit}{bold}{\encodingdefault}{\sfdefault}{bx}{n}

\makeatletter
\let\save@mathaccent\mathaccent
\newcommand*\if@single[3]{%
  \setbox0\hbox{${\mathaccent"0362{#1}}^H$}%
  \setbox2\hbox{${\mathaccent"0362{\kern0pt#1}}^H$}%
  \ifdim\ht0=\ht2 #3\else #2\fi
  }
\newcommand*\rel@kern[1]{\kern#1\dimexpr\macc@kerna}
\newcommand*\widebar[1]{\@ifnextchar^{{\wide@bar{#1}{0}}}{\wide@bar{#1}{1}}}
\newcommand*\wide@bar[2]{\if@single{#1}{\wide@bar@{#1}{#2}{1}}{\wide@bar@{#1}{#2}{2}}}
\newcommand*\wide@bar@[3]{%
  \begingroup
  \def\mathaccent##1##2{%
    \let\mathaccent\save@mathaccent
    \if#32 \let\macc@nucleus\first@char \fi
    \setbox\z@\hbox{$\macc@style{\macc@nucleus}_{}$}%
    \setbox\tw@\hbox{$\macc@style{\macc@nucleus}{}_{}$}%
    \dimen@\wd\tw@
    \advance\dimen@-\wd\z@
    \divide\dimen@ 3
    \@tempdima\wd\tw@
    \advance\@tempdima-\scriptspace
    \divide\@tempdima 10
    \advance\dimen@-\@tempdima
    \ifdim\dimen@>\z@ \dimen@0pt\fi
    \rel@kern{0.6}\kern-\dimen@
    \if#31
      \overline{\rel@kern{-0.6}\kern\dimen@\macc@nucleus\rel@kern{0.4}\kern\dimen@}%
      \advance\dimen@0.4\dimexpr\macc@kerna
      \let\final@kern#2%
      \ifdim\dimen@<\z@ \let\final@kern1\fi
      \if\final@kern1 \kern-\dimen@\fi
    \else
      \overline{\rel@kern{-0.6}\kern\dimen@#1}%
    \fi
  }%
  \macc@depth\@ne
  \let\math@bgroup\@empty \let\math@egroup\macc@set@skewchar
  \mathsurround\z@ \frozen@everymath{\mathgroup\macc@group\relax}%
  \macc@set@skewchar\relax
  \let\mathaccentV\macc@nested@a
  \if#31
    \macc@nested@a\relax111{#1}%
  \else
    \def\gobble@till@marker##1\endmarker{}%
    \futurelet\first@char\gobble@till@marker#1\endmarker
    \ifcat\noexpand\first@char A\else
      \def\first@char{}%
    \fi
    \macc@nested@a\relax111{\first@char}%
  \fi
  \endgroup
}
\makeatother

\definecolor{darkred}{rgb}{0.7, 0.0, 0.0}

\usepackage{pifont}
\usepackage{wrapfig}
\usepackage{caption}
\usepackage{algorithm}
\usepackage{algpseudocode}
\usepackage{amsmath}
\usepackage{amssymb}
 
\usepackage[nameinlink]{cleveref}
\crefname{equation}{Eq.}{Eqs.}
\crefname{figure}{Fig.}{Figs.}
\crefname{section}{Sec.}{Sec.}
\crefname{appendix}{App.}{App.}
\crefname{table}{Tab.}{Tabs.}
\crefname{algorithm}{Algo}{Algo}
\crefname{thm}{Thm}{Thm}
\Crefname{thm}{Thm}{Thm}
\crefname{prop}{Prop}{Prop}
\usepackage{longtable}

\newcommand{\ourmethod}{\mbox{\textsc{DreamZero}}\xspace}

\newcommand{\crefnames}[3]{%
  \@for\next:=#1\do{%
    \expandafter\crefname\expandafter{\next}{#2}{#3}%
  }%
}

\title{World Action Models are Zero-shot Policies}

\newcommand{\symbolfootnotetext}[2]{%
  \begingroup%
  \renewcommand{\thefootnote}{#1}%
  \let\oldfootnoterule\footnoterule%
  \renewcommand{\footnoterule}{\oldfootnoterule}%
  \footnotetext{#2}%
  \endgroup%
}

\author{
  \textbf{Seonghyeon Ye}\textsuperscript{$\dagger$} \quad
  \textbf{Yunhao Ge}\textsuperscript{*} \quad
  \textbf{Kaiyuan Zheng}\textsuperscript{*} \quad
  \textbf{Shenyuan Gao}\textsuperscript{*} \quad
  \textbf{Sihyun Yu}\textsuperscript{*} \quad
  \textbf{George Kurian}\textsuperscript{*} \quad \quad
  \textbf{Suneel Indupuru}\textsuperscript{*} \quad 
  \textbf{You Liang Tan}\textsuperscript{*} \quad 
  \textbf{Chuning Zhu}\quad 
  \textbf{Jiannan Xiang} \quad
  \textbf{Ayaan Malik} \quad
  \textbf{Kyungmin Lee} \newline
  \textbf{William Liang} \quad 
  \textbf{Nadun Ranawaka} \quad
  \textbf{Jiasheng Gu} \quad
  \textbf{Yinzhen Xu} \quad
  \textbf{Guanzhi Wang} \quad
  \textbf{Fengyuan Hu} \newline 
  \textbf{Avnish Narayan} \quad
  \textbf{Johan Bjorck} \quad
  \textbf{Jing Wang} \quad
  \textbf{Gwanghyun Kim} \quad
  \textbf{Dantong Niu} \quad
  \textbf{Ruijie Zheng} \quad
  \textbf{Yuqi Xie} \quad
  \textbf{Jimmy Wu} \quad
  \textbf{Qi Wang} \quad
  \textbf{Ryan Julian} \quad 
  \textbf{Danfei Xu} \quad
  \textbf{Yilun Du} \quad
  \textbf{Yevgen Chebotar} \quad
  \textbf{Scott Reed} \quad
  \textbf{Jan Kautz} \quad \newline
  \textbf{Yuke Zhu}\textsuperscript{$\dagger$} \quad
  \textbf{Linxi ``Jim'' Fan}\textsuperscript{$\dagger$} \quad
  \textbf{Joel Jang}\textsuperscript{$\dagger$} \\
  \small NVIDIA \\
  \small $^{\dagger}$Project Leads \quad
         $^{\ast}$Core Contributors \\
  {\small \url{https://dreamzero0.github.io}}
}

\begin{abstract}
State-of-the-art Vision-Language-Action (VLA) models excel at semantic generalization but struggle to generalize to unseen physical motions in novel environments.
We introduce \ourmethod, a World Action Model (WAM) built upon a pretrained video diffusion backbone. Unlike VLAs, WAMs learn physical dynamics by predicting future world states and actions, using video as a dense representation of how the world evolves.
By jointly modeling video and action, \ourmethod learns diverse skills effectively from heterogeneous robot data without relying on repetitive demonstrations. This results in over 2$\times$ improvement in generalization to new tasks and environments compared to state-of-the-art VLAs in real-robot experiments.
Crucially, through model and system optimizations, we enable a 14B autoregressive video diffusion model to perform real-time closed-loop control at 7Hz. Finally, we demonstrate two forms of cross-embodiment transfer: video-only demonstrations from other robots or humans yield a relative improvement of over 42\% on unseen task performance with just 10–20 minutes of data. More surprisingly, \ourmethod enables few-shot embodiment adaptation, transferring to a new embodiment with only 30 minutes of play data while retaining zero-shot generalization.
\end{abstract}

\begin{document}

\maketitle
\begin{figure}[ht!]
    \centering
    \includegraphics[width=1\textwidth]{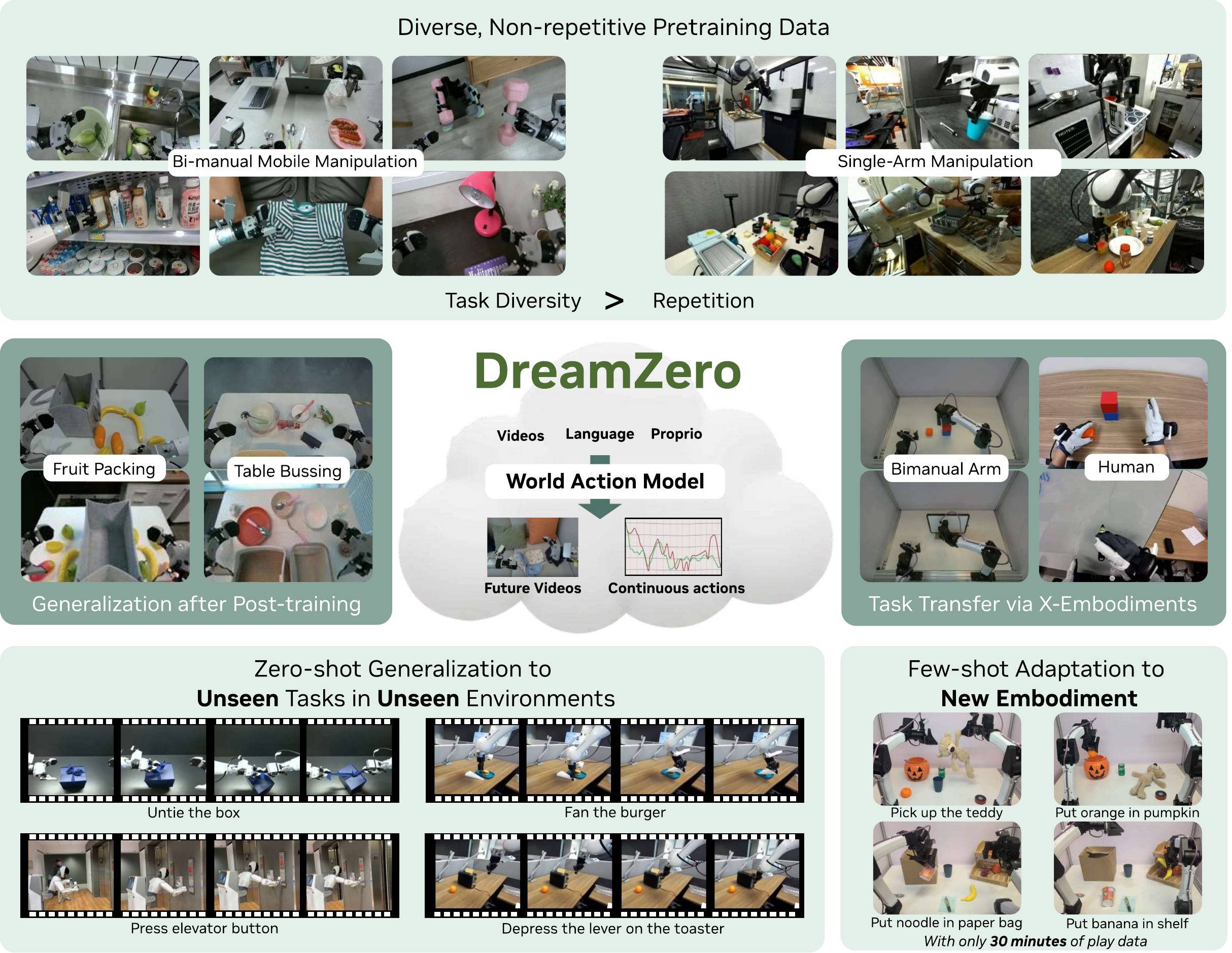}
    \caption{\textbf{Overview}. By jointly predicting video and action, World Action Models (WAMs) inherit world physics priors that enable 1) effective learning from diverse, non-repetitive data, 2) open-world generalization, 3) cross-embodiment learning from video-only data, and 4) few-shot adaptation to new robots.}
    \label{fig:fig1_preview}
\end{figure}
\clearpage
\abscontent

\section{Introduction}
\label{sec:intro}
Recent robotic foundation models, termed Vision-Language Action models (VLAs), extend pretrained Vision-Language Models (VLMs) to predict motor actions~\citep{kim2024openvla, black2410pi0, bjorck2025gr00t, team2025gemini, rt22023arxiv}. While VLAs successfully inherit linguistic priors to generalize across diverse language instructions, especially manipulating diverse objects~\citep{rt22023arxiv}, their generalization to novel environments and, more critically, to new motions or skills remains limited \citep{zhou2025exploring, guruprasad2025benchmarking}. For example, VLAs can successfully execute ``move coke can to Taylor Swift'' \citep{rt22023arxiv} by leveraging the web knowledge acquired during VLM pretraining to identify the target location, and connecting it to the learned move skill from the robot data. However, they fail at a task like ``untie the shoelace'' if that specific skill was not present in the robot training data.  
Although VLM priors encode \textit{what} to do at a semantic level, they lack representations of \textit{how} actions should be executed with precise spatial awareness, aligned with geometry, dynamics, and motor control \citep{chen2024spatialvlm, feng2025seeing}. 
As a result, VLAs often struggle to adapt to new environments or generalize to novel tasks beyond the distribution of expert demonstrations, without explicitly collecting large-scale task- and environment-specific action data. 

In this paper, we present \ourmethod, a 14B robot foundation model built upon a pretrained image-to-video diffusion backbone~\citep{wan2025wan}. We term this architecture a \textit{World Action Model (WAM)}—a foundation model designed to predict both actions and visual future states in an aligned manner. Initialized from video diffusion models trained on web-scale video data, WAMs leverage rich spatiotemporal priors to jointly generate future frames and actions conditioned on language instructions and observations. This shifts action learning from dense state–action imitation to inverse dynamics—aligning motor commands with predicted visual futures. Consequently, we observe that this enables (1) effective learning from robot data that are heterogeneous trajectories collected during the execution of useful behaviors in real-world settings, rather than relying solely on carefully repeated demonstrations (2) zero-shot generalization to new tasks in new environments, and (3) efficient cross-embodiment transfer.

\begin{figure}[t!]
    \vspace{-0.05in}
    \centering
    \includegraphics[width=1\textwidth]{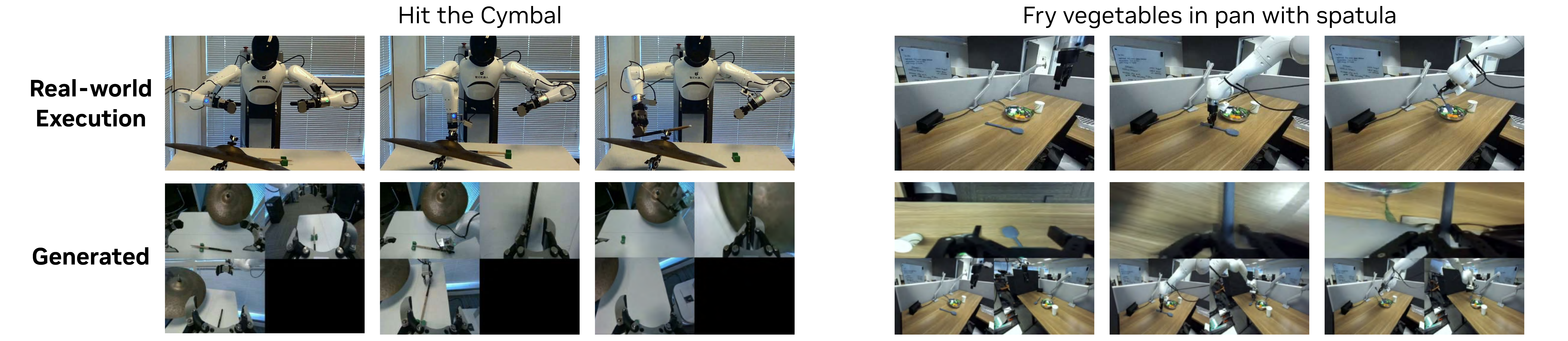}
    \caption{
    \textbf{Joint Video and Action Prediction}. \ourmethod jointly generates video and action. We observe that the predicted actions closely align with the generated video. The examples are from totally unseen tasks.
    }
    \vspace{-0.13in}
    \label{fig:dreamzero_gen}
\end{figure}
 
\begin{figure}[t!]
    \centering
    \includegraphics[width=1\textwidth]{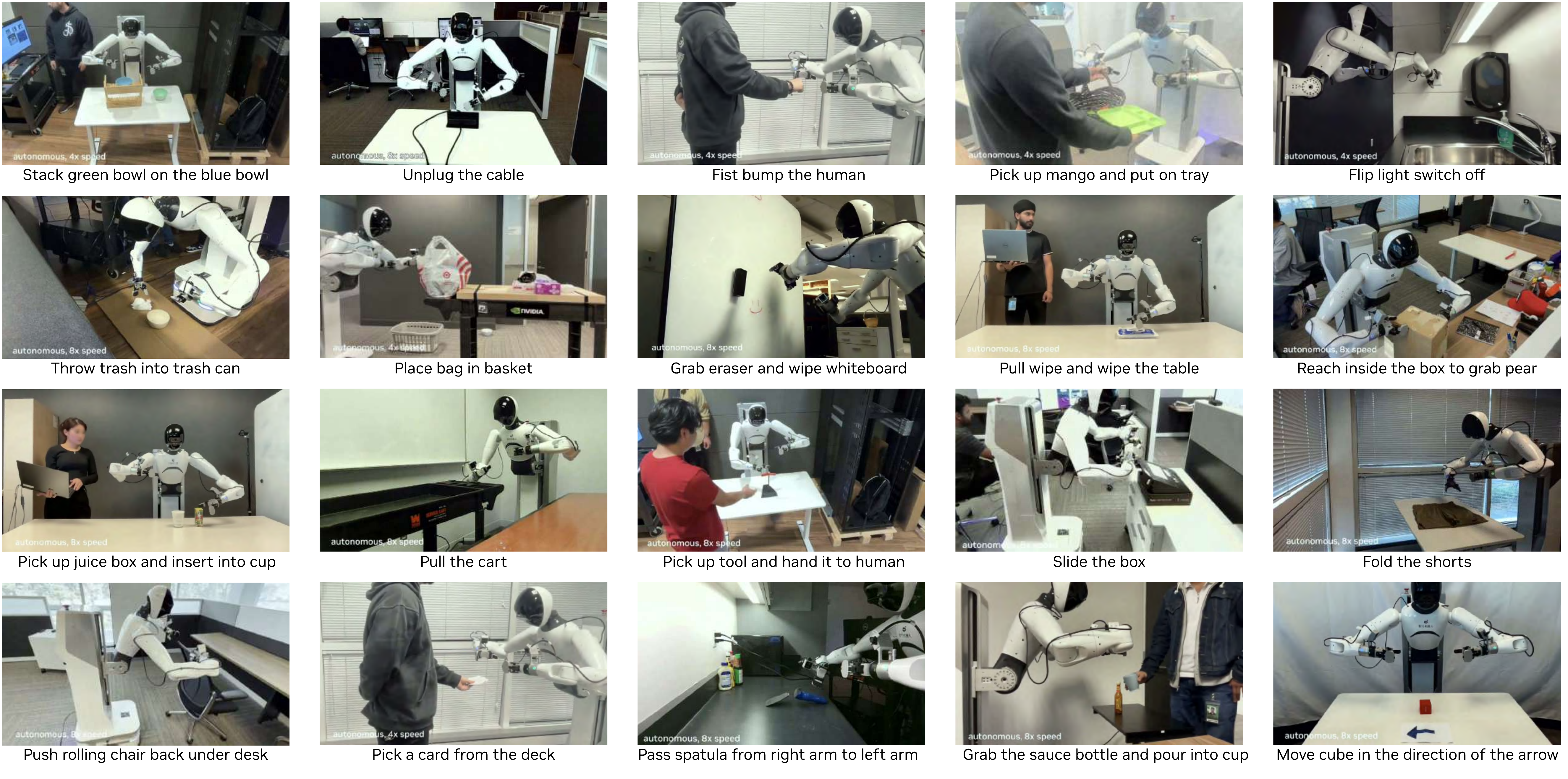}
    \caption{\textbf{Free-form Evaluation.} \ourmethod performs a diverse range of tasks when conditioned on natural language instructions, including object manipulation, tool use, and human-robot interaction.}
    \label{fig:free_form}
\end{figure}

This approach yields three core advancements that distinguish \ourmethod from prior work, including other WAMs~\citep{kim2026cosmos,liang2025video,pai2025mimic}. First, \ourmethod unlocks new generalization capabilities beyond traditional VLAs and previous WAMs—across environments, across tasks, and across embodiments (Figure \ref{fig:dreamzero_gen} and Figure \ref{fig:free_form}). Compared to the state-of-the-art pretrained VLAs, we observe more than a 2$\times$ improvement in average task progress on environment and task generalization benchmarks. Second, \ourmethod demonstrates that generalist policies can be learned effectively from diverse, heterogeneous data, breaking away from the conventional wisdom that generalist robot policies require multiple repeated demonstrations per task. Although other WAMs show that priors learned from videos prediction improves sample efficiency for action learning compared to VLAs \citep{pai2025mimic, liao2025genie}, most works still focus on repeated demonstrations. 
Moreover, the environment generalization of \ourmethod is retained even after task-specific post-training, outperforming state-of-the-art VLAs by 10\% on average task progress. Lastly, we demonstrate two forms of cross-embodiment transfer. First, video-only demonstrations from another robot (YAM) or humans yield a relative improvement of over 42\% on unseen task performance for the target robot (AgiBot G1) with just 10–20 minutes of data. Second, and more surprisingly, we show that \ourmethod enables \textit{few-shot embodiment adaptation}: a model pretrained on AgiBot G1 adapts to an entirely new robot (YAM) with only 30 minutes of play data, retaining zero-shot generalization. To the best of our knowledge, this sets a new benchmark for data-efficient embodiment adaptation.

\ourmethod is a 14B autoregressive diffusion transformer trained with a teacher-forcing chunk-wise video denoising objective.
Our architectural analysis reveals that larger pretrained video diffusion models produce higher-quality video predictions, which directly translates to superior downstream action execution—indicating that policy performance is fundamentally tied to video generation quality. We further find that diverse distribution of the training data is essential for generalization, outperforming multi-task repetitive data with the same amount of hours. Furthermore, we observe that autoregressive architectures lead to smoother robot motions and higher modality alignment between predicted videos and executed actions.

To address the computational overhead inherent to video diffusion models, we introduce a suite of optimizations spanning three categories: (1) algorithmic improvements, including decoupled video and action denoising schedules (\ourmethod-Flash); (2) system-level parallelism and caching strategies; and (3) low-level optimizations such as quantization, and CUDA kernel tuning. Collectively, these techniques achieve a 38× inference speedup without degrading performance, enabling \ourmethod to generate action chunks at approximately 7Hz for smooth, real-time robotic control.

Our main contributions are:
\begin{itemize}[leftmargin=*, itemsep=2pt, topsep=2pt]
    \item We introduce \ourmethod, a 14B WAM that jointly predicts video and actions, enabling effective learning from diverse, non-repetitive robot data.
    \item We demonstrate over 2$\times$ improvement in zero-shot generalization to \textbf{unseen verbs and motions} compared to state-of-the-art VLAs, while retaining generalization across objects and environments.
    \item We present model and system optimizations achieving \textbf{38$\times$ inference speedup}, enabling real-time closed-loop control at \textbf{7Hz}.
    \item We demonstrate cross-embodiment transfer: video-only data from humans (12 minutes) or other robots (20 minutes) yields a relative improvement of over 42\% on unseen tasks, and introduce \textbf{few-shot embodiment adaptation}—\ourmethod pretrained on AgiBot G1 adapts to an entirely new robot (YAM) with only 30 minutes of play data, enabling zero-shot generalization.
    \item We \textbf{open-source} our model weights, inference code, and code to run publicly available real-world (RoboArena) and simulation benchmarks (PolaRiS and Genie Sim 3.0)\footnote{Despite only being trained on $\sim$500 hours of real-world data, \ourmethod shows non-trivial performance on Genie Sim 3.0, which is a simulation benchmark comprised of 100 different tasks \textit{without} being explicitly trained on the 10k hours of simulation training data.} at \url{https://github.com/dreamzero0/dreamzero}.
\end{itemize}

\section{Related Work}
\label{sec:related}
\subsection{Vision Language Action Models}
\textbf{Utilizing Foundation Models for Robotics.} Developing foundation models~\citep{bommasani2021opportunities} for physical artificial intelligence has emerged as a significant research frontier. One line of work involves using existing, pre-trained foundation models as ``black-box'' reasoners to handle high-level task planning. These works usually involve \textit{modular} systems, where the foundation models generate sequences of instructions, visual traces, or affordances that are subsequently executed by specialized, low-level robotic policies or controllers~\citep{saycan-2022,innermono-2022,progprompt-2022,driess2023palm, huang2023voxposer, 11361071}. While this modularity simplifies complex planning and enables stronger generalization~\citep{li2025hamster, lee2025molmoact, kaelbling1} and efficiency~\citep{dreczkowski2025learning}, it is contingent upon having a pre-existing library of low-level skills and a robust interface to bridge the gap between abstract reasoning and physical execution. Additionally, these decoupled systems face the risk of compounding errors across modules.

\textbf{VLAs.} On the other hand, end-to-end models such as Vision-Language-Action models (VLAs)~\citep{rt1-2022, rt22023arxiv, kim24openvla, zheng2025tracevla, ye2025latent, yang2025magma, black2024pi_0, bjorck2025gr00t, intelligence2025pi, team2025gemini, bu2025univla}, have gained popularity by moving away from a rigid hierarchy of planning and control, combining language-conditioned semantics and low-level robot actions within the same model. VLAs are often initialized from large vision-language (VLM) models pre-trained on web-scale datasets. While pushing the frontier on visual-semantic knowledge transfer, these models are pre-trained on \textit{static} image-text datasets, which limits their ability to inherit spatiotemporal priors required to transfer knowledge to new physical skills.

\textbf{Generalization in VLAs.} Generalization in VLAs has been mostly demonstrated on object and semantic level~\citep{rt22023arxiv, gao2025taxonomy} while generalization to completely new skills and environments has remained limited. In particular, existing work utilizing VLAs achieves environment generalization by collecting human teleoperation data across hundreds of diverse environments for specific tasks~\citep{intelligence2025pi}. Furthermore, while current VLAs attempt to achieve task generalization by covering a large library of language-conditioned motion primitives~\citep{team2025gemini}, this approach is fundamentally constrained by the impracticality of capturing the vast amount of possible physical interactions and motions with a fixed set of episode-level language-conditioned tasks. In contrast, video-based world models learn from every consecutive frame pair in the data, while also leveraging large-scale video pretraining to understand physical dynamics.

\subsection{Video Model-based Robot Policies}
\textbf{Video Generation in Robotics.} Prior works show that video generation models can be used to synthesize robot trajectories and extract executable actions at test-time through various approaches: inverse-dynamics models~\citep{du2023learning, zhou2024robodreamer}, optical flow as dense correspondence~\citep{ko2024learning}, or trajectory prediction as high-level planning~\citep{yang2024learning, du2024video}. Other works generate human videos—either with 3D tracking~\citep{liang2024dreamitate} or for novel scenes and motions~\citep{bharadhwaj2024gen2act, chen2025large}—and train policies using point tracking objectives. Most recently, \citep{jang2025dreamgen, luo2025solving} demonstrated that video generation models can produce synthetic robot data for unseen behaviors in novel environments, leveraging the strong generalization capabilities of these models.

\textbf{Joint Video and Action Generation.} Another line of work couples video and action generation for end-to-end learning. These methods demonstrate that incorporating a world modeling objective alongside action prediction improves multi-task performance, sample efficiency, and generalization to novel scenes and objects. Previous work~\citep{wu2024unleashing, cheang2024gr, li2025unified, zhu2025unified, zhao2025cot, zheng2025flare, won2025dual} learns to do joint world modeling and action prediction from scratch or from VLAs, while more recent work~\citep{kim2026cosmos, liao2025genie, hu2024video, liang2025video, pai2025mimic} leverages pretrained video diffusion models to inherit rich visual dynamics priors. We refer to these models collectively as \textit{World Action Models (WAMs)} since they leverage world modeling capability (predicting the future state) for action prediction. We use the term World Action Models (WAMs) rather than Video Action Models (VAMs) to reflect that video is just one possible world modeling objective—future WAMs may align actions with other predictive modalities such as tactile sensing, force feedback, or learned latent representations. In contrast to prior WAMs, \ourmethod systematically explores data diversity and scale to expose the full generalization potential of WAMs, adopts an autoregressive architecture better suited for long-horizon world–action modeling, achieves state-of-the-art generalization across both novel tasks and environments, and achieves state-of-the-art cross-embodiment tranfer, both learning from different embodiments (video only) and few-shot adaptation to a new embodiment.

\textbf{Why WAMs.} WAMs built upon video diffusion backbones inherit rich spatiotemporal priors from web-scale data, capturing the best of both paradigms: the seamless gradient flow of end-to-end VLAs and dense world modeling supervision for planning. Unlike latent world models~\citep{hafner2019dream, hafner2020mastering, hafner2023mastering, assran2025vjepa2}, which learn dynamics from scratch in compact latent spaces, WAMs leverage pretrained video representations that already encode physical dynamics from internet-scale data. Central to this approach is learning the joint distribution of video and action—\ourmethod simultaneously learns both modalities, with video prediction serving as an implicit visual planner that guides action generation. This formulation not only means that improving robotic capabilities reduces to improving video generation, but also enables three capabilities that elude current VLAs: zero-shot generalization to novel tasks, effective learning from heterogeneous robot data, and extremely efficient cross-embodiment transfer from videos. We provide further discussion about the differences between WAMs and alternative world model architectures (e.g., latent-space, 3D point cloud) in Appendix~\ref{appen:world_model_comparison}.
\begin{figure}[ht!]
    \centering
    \includegraphics[width=1\textwidth]{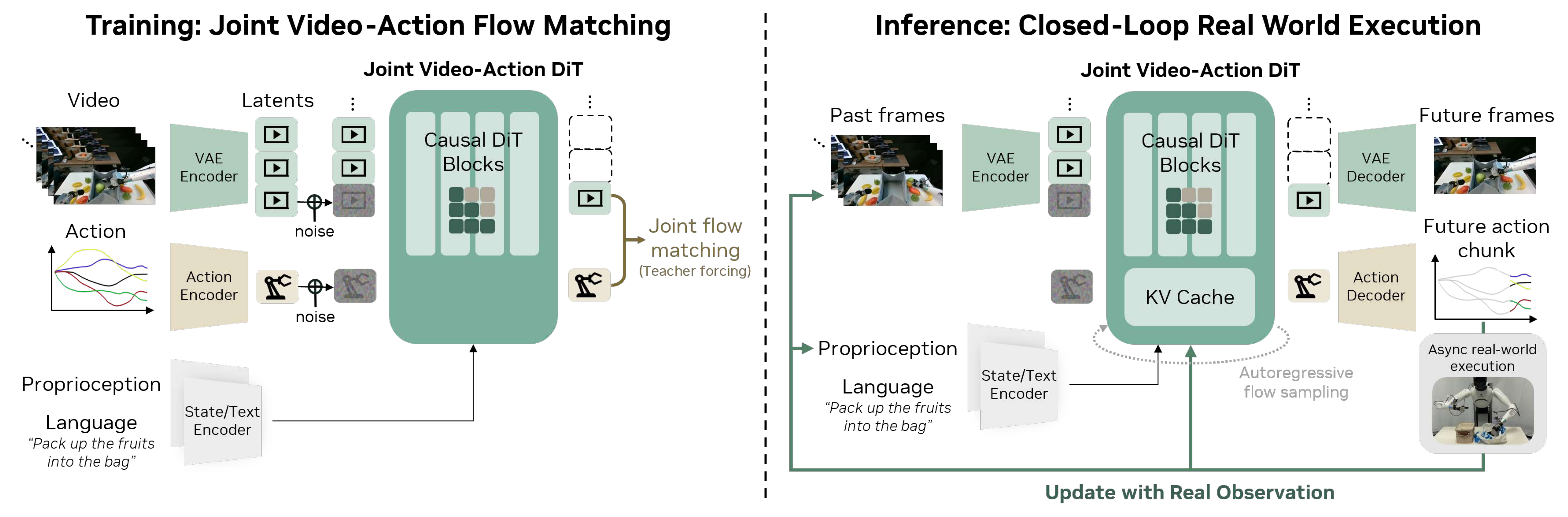}
    \caption{\textbf{Model Architecture of \ourmethod.} The model takes three inputs: visual context (encoded via a VAE), language instructions (via a text encoder), and proprioceptive state (via a state encoder). These are processed by an autoregressive DiT backbone using flow matching, which jointly predicts future video frames and actions through separate decoders. During training (left), for each chunk, the model denoises noisy video and action latents conditioned on clean video context. During inference (right), predictions are executed asynchronously in the real world, and ground-truth observations are fed back into the KV cache to prevent error accumulation.}
    \label{fig:main_arch}
\end{figure}
\section{\ourmethod}
\label{sec:method}
Pretrained video diffusion models offer rich spatiotemporal priors from web-scale data, making them attractive backbones for robot policies. However, converting these models into effective World Action Models (WAMs) presents three key challenges: (1) \textbf{Video-action alignment}: jointly predicting video and actions requires tight coupling between visual futures and motor commands, yet naively combining separate video and action heads can lead to misalignment; (2) \textbf{Architectural design}: it remains unclear whether bidirectional or autoregressive architectures are better suited for WAMs, with implications in modality alignment, error accumulation, and inference efficiency; and (3) \textbf{Real-time inference}: video diffusion models require iterative denoising across high-dimensional latent spaces, making them prohibitively slow for closed-loop control.

\ourmethod addresses these challenges through three design choices. First, we train a single end-to-end model that jointly denoises video and action with a shared objective, ensuring deep integration between modalities. Second, we adopt an autoregressive architecture and exploit the closed-loop setting: after each action chunk is executed, we replace predicted frames with ground-truth observations in the KV cache, eliminating compounding errors while enabling efficient inference via KV caching and preserving native frame rates for precise modality alignment (See right side of Figure \ref{fig:main_arch}). Third, we introduce a suite of system-, implementation-, and model-level optimizations that achieve a 38$\times$ inference speedup, enabling real-time control at 7Hz. We detail the model architecture in Section~\ref{subsec:arch} and real-time execution in Section~\ref{subsec:realtime}.

\subsection{Model Architecture}
\label{subsec:arch}
\textbf{Problem Formulation.} \ourmethod jointly predicts video $\mathbf{o}_{l:l+H}$ and actions $\mathbf{a}_{l:l+H}$ conditioned on language instruction $\mathbf{c}$, proprioceptive state $\mathbf{q}_l$ and visual observation including the current and the past history $\mathbf{o}_{0:l}$ where $H>0$ is a fixed horizon and $l$ is a random index sampled from a trajectory. Note that joint prediction of video and action is a decomposition of (1) autoregressive video prediction and (2) action prediction from an inverse-dynamics model (IDM): 
\begin{equation}
    \underbrace{\pi_0(\mathbf{o}_{l:l+H}, \mathbf{a}_{l:l+H} \mid \mathbf{o}_{0:l}, \mathbf{c}, \mathbf{q}_l)}_{\ourmethod} = \underbrace{\pi_0(\mathbf{o}_{l:l+H} \mid \mathbf{o}_{0:l}, \mathbf{c}, \mathbf{q}_l)}_{\text{video prediction}} \underbrace{\pi_0(\mathbf{a}_{l:l+H} \mid \mathbf{o}_{0:l+H}, \mathbf{q}_l)}_{\text{IDM}}
\label{eq:joint_pred}
\end{equation}
Instead of using two separate models (video prediction model and inverse dynamics model) to model the decomposed objective \citep{lingbot-va2026, pai2025mimic}, we train a single model end-to-end with joint prediction objective. We believe that this end-to-end design enables better video-action alignment through a deep integration between the two modalities. Since pretrained video models are already optimized on the video prediction objective on diverse web-scale video data, \ourmethod only needs to additionally learn to predict videos for the robot embodiment videos and extract corresponding actions from the generated videos. We further hypothesize that this encourages better generalization than the conventional practice of training VLA from VLM 
, as our approach explicitly learns temporal dynamics from video frames used both as conditioning inputs and prediction targets.

\textbf{Model Architecture.} The model architecture is shown in Figure~\ref{fig:main_arch}. To retain the generalization capability of video models, we introduce minimal additional parameters: state encoders, action encoders, and decoders. For robot training data that contains multiple views, we concatenate all views into a single frame instead of making architectural changes to the backbone model.

In particular, \ourmethod is trained to predict video frames and corresponding actions autoregressively. Autoregressive generation possesses the following advantages: (1) it enables faster inference speed by utilizing KV-cache, (2) the policy model can leverage the visual observation history as guidance for the next generation, and (3) it avoids the modality alignment challenges (video, action, and language alignment) inherent to bidirectional models. Concretely, bidirectional diffusion typically requires processing fixed-length sequences, which often necessitates video subsampling that distorts native FPS, potentially harming video-action alignment. On the other hand, autoregressive generation leverages KV caching to support arbitrarily long contexts within a single forward pass. This preserves the native frame rate, ensuring precise alignment between video frames and robot actions.
 Some details illustration of this difference is provided in Appendix \ref{appen:bidir_ar}.

 We introduce autoregressive modeling only for the video modality to avoid error propagation coming from closed-loop action prediction. \ourmethod is trained to predict video frames in a \textit{chunk} manner; each chunk has a fixed number of latent frames $K$ to match the action horizon. Chunk-wise generation enables training on variable length of videos, similar to how LLMs are trained on variable length of language tokens. We provide more details on the QKV attention masking strategy for the different modalities in Appendix \ref{appen:model_detail}.

\textbf{Training Objective.} Similar to recent video diffusion models and VLAs, we employ flow-matching \citep{liu2022flow, lipman2022flow, albergo2023stochastic} as the training objective \citep{wan2025wan, ali2025world, teng2025magi}. Unlike recent WAMs \citep{zhu2025unified, kim2026cosmos, li2025unified, liao2025genie}, \ourmethod shares the denoising timestep between video and action modality for faster convergence at the beginning of training. 
Also, we apply teacher forcing \citep{jin2024pyramidal, gao2024ca2} as a training objective; the model is trained to denoise the noisy current chunk conditioned on the clean previous chunks. 

Formally, given a chunk index $k>0$ and the denoising timestep $t_{k} \in [0,1]$, we denote the corresponding noisy video latent vector for original video $\mathbf{o}^{k}$ as $\mathbf{z}_{t_{k}}^{k}$ and noisy normalized actions as $\mathbf{a}_{t_{k}}^{k}$. 
All frames within the same chunk share the same timestep $t_{k}$, while different chunks are assigned independent timesteps.
Our model denoises 
$\mathbf{z}_{t_{k}}^{k}$ and $\mathbf{a}_{t_{k}}^{k}$, defined as linear interpolations between clean vectors and random Gaussian noises:
\begin{equation}
        \mathbf{z}_{t_{k}}^{k} = t_k \mathbf{z}_{1}^{k} + (1 - t_k) \mathbf{z}_{0}^{k},\quad
        \mathbf{a}_{t_{k}}^{k} = t_k \mathbf{a}_{1}^{k} + (1 - t_k) \mathbf{a}_{0}^{k},
\end{equation}
where $\mathbf{z}_{0}^{k} \sim \mathcal{N}(\mathbf{0},\mathbf{I}), \mathbf{a}_{0}^{k} \sim \mathcal{N}(\mathbf{0},\mathbf{I})$, and $\mathbf{z}_{1}^{k}$ and $\mathbf{a}_{1}^{k}$ are a clean video latent vector and a normalized action, respectively.
Thus, the clean context from the previous chunks can be denoted as $\mathcal{C}_{k} = \{(\mathbf{z}_{1}^{j}, \mathbf{a}_{1}^{j})\}_{j=1}^{k-1}$. 

We train the model $\mathbf{u}_{\theta}$ to predict the joint velocity for both modalities using the following flow-matching objective:
\begin{equation}
    \mathcal{L}(\theta) = \mathbb{E}_{\mathbf{z}, \mathbf{a}, \{t_k\}} \Bigg[\frac{1}{K} \sum_{k=1}^{K} w(t_k) \big\lVert \mathbf{u}_{\theta}([\mathbf{z}_{t_k}^{k}, \mathbf{a}_{t_k}^{k}];\mathcal{C}_{k}, \mathbf{c}, \mathbf{q}_{k}, t_{k}) - \mathbf{v}^{k} \big\rVert^2\Bigg],
\end{equation}
where $w(t_k)>0$ is a predefined weight function for $t_k$, $\mathbf{c}$ is the text condition, $\mathbf{q}_{k}$ is the proprioceptive states of $k$-th chunk, and the velocity $\mathbf{v}^{k} \coloneqq [\mathbf{z}_{1}^{k}, \mathbf{a}_{1}^{k}] - [\mathbf{z}_{0}^{k}, \mathbf{a}_{0}^{k}]$. To enable efficient training, we perform trajectory-level updates and apply attention masking (e.g., see Figure~\ref{fig:attn_mask} for details) so that the current noisy chunk can attend to clean context of previous chunks. We provide the pseudo-code in Algorithm \ref{alg:dreamzero_train}.

\textbf{Model Inference.} As shown in Figure~\ref{fig:main_arch}, during inference, \ourmethod jointly denoises video and action chunks, leveraging KV caching for efficiency~\citep{huang2025self, teng2025magi, yin2025slow}. Unlike pure video generation, our closed-loop setting allows ground-truth observations to replace generated frames in the KV cache after each action execution (see Figure~\ref{fig:attn_mask}). This eliminates the compounding error problem inherent to autoregressive video generation—a key advantage unique to WAMs. Moreover, as a stateful policy, \ourmethod can leverage visual history for tasks requiring memory.\footnote{In this work, we do not explicitly evaluate or post-train \ourmethod on tasks that can only succeed with memory. We leave this for future work.}. We provide the pseudo-code of inference in Algorithm \ref{alg:dreamzero_inference}
\subsection{Real-time Execution of \ourmethod}
\label{subsec:realtime}

Diffusion-based WAMs inherit powerful generalization from video foundation models, but their iterative denoising process creates a fundamental tension with reactive robotic control. We address two questions: (1) What prevents WAMs from being reactive policies? (2) How do we resolve this for real-time control?

\subsubsection{The Reactivity Gap}

Reactive policies must respond to environmental changes within tens of milliseconds. A naive implementation of \ourmethod on a single GPU requires approximately 5.7 seconds per action chunk due to three bottlenecks: (1) iterative denoising across 16 diffusion steps required for smooth actions, (2) the computational cost of a 14B parameter DiT backbone, and (3) sequential execution that blocks robot motion during inference. This latency makes closed-loop control infeasible.\footnote{One might expect that generating only actions (not video) would accelerate inference, but at 14B scale we empirically found out that the speed gain is minimal—the number of diffusion steps and the number of DiT blocks dominate latency. Moreover, because video and action are jointly trained for strong cross-modal alignment, naively reducing action denoising steps degrades quality. This motivates \ourmethod-Flash.}

\subsubsection{Asynchronous Closed-Loop Execution}

Our first step towards resolving this is through asynchronous execution that decouples inference from action execution. Rather than waiting for each inference to complete, the motion controller continuously executes the most recent action chunk while inference runs concurrently on the latest observation. This structure transforms the latency constraint from ``inference must complete before the robot moves'' to ``inference must complete before the current action chunk expires.'' In our experiments, we deploy policies at an action horizon of 48 steps at 30Hz control frequency (1.6 seconds per chunk) for bimanual manipulation robots. Hence, we target inference latency below approximately 200ms to ensure sufficient overlap for smooth, reactive control.

\subsubsection{System-level Optimizations}

Given the asynchronous execution structure, we optimize inference throughput through parallelism and caching.

\begin{itemize}
    \item \textbf{CFG Parallelism.} Classifier-free guidance \citep{ho2022classifier} requires two forward passes (conditional and unconditional). We distribute these across two GPUs, reducing per-step latency by 47\%.
    \item \textbf{DiT Caching.} We exploit the directional consistency of velocity predictions during flow matching. When cosine similarity between successive velocities exceeds a threshold, we reuse cached velocities, reducing effective DiT steps from 16 to 4 with minimal quality loss on action prediction.
\end{itemize}

\subsubsection{Implementation-level Optimizations}

We further reduce latency through compiler and kernel enhancements.
\begin{itemize}
    \item \textbf{Torch Compile and CUDA Graphs.} We apply \texttt{torch.compile} with CUDA Graphs to eliminate CPU overhead and fuse operators. Static shapes cause recompilations only during the first trajectory.
    \item \textbf{Post-Training Quantization.} On Blackwell architecture, we quantize weights and activations to NVFP4 while keeping sensitive operations (QKV, Softmax) in FP8 and non-linear operations in FP16.
    \item \textbf{Kernel and Scheduler Enhancements.} We use the cuDNN backend for attention and migrate scheduler operations to GPU to eliminate CPU-GPU synchronization stalls.
\end{itemize}
\begin{figure}[h!]
    \centering
    \includegraphics[width=0.8\textwidth]{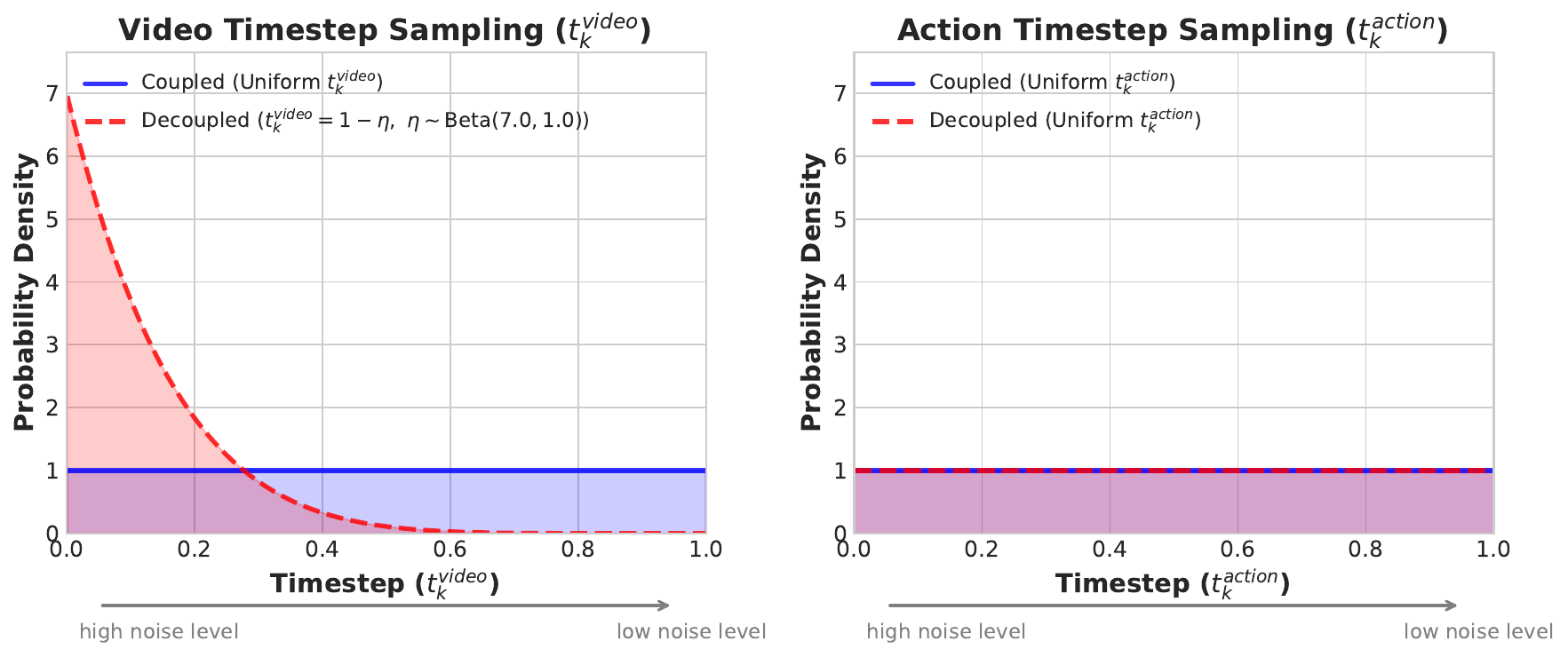}
    \caption{\textbf{Decoupled Noise Schedules.} \ourmethod (blue) uses coupled noise for video and action (both uniform). \ourmethod-Flash (red) biases video toward high-noise states via a Beta distribution while keeping action noise uniform, training the model to predict clean actions from noisy visual context.}
    \label{fig:dreamzeroflash}
\end{figure}
\subsubsection{Model-level Optimizations: \ourmethod-Flash}
\label{subsec:dreamzeroflash}

Even with system optimizations, the number of diffusion steps remains the primary latency bottleneck. However, naively reducing steps degrades action quality because residual visual noise propagates into action predictions.

\ourmethod-Flash addresses this by decoupling video and action noise schedules during training. 
The key insight is that, at inference time, actions should denoise to their final values while being conditioned on a still-noisy video representation within the current chunk, since with very few denoising steps (e.g., fewer than 4), the generated video tokens may remain inaccurate and thus provide a noisy conditioning signal.
Standard \ourmethod samples a shared timestep $t_k \sim \mathcal{U}(0,1)$ for both modalities. This creates a train-test mismatch: during training, the model learns to predict actions when video and action are at the \textit{same} noise level, but few-step or single-step inference requires predicting clean actions while video remains partially noisy.

\ourmethod-Flash closes this gap by biasing video timesteps toward high-noise states via $t_k^{\text{video}} = 1 - \eta$, where $\eta \sim \text{Beta}(\alpha, \beta)$ with $\alpha > \beta$.
In practice, we use $\text{Beta}(7, 1)$ as an example configuration, yielding $\mathbb{E}[t_k^{\text{video}}] = 0.125$ (predominantly noisy), while action timesteps remain uniform (Figure~\ref{fig:dreamzeroflash}). During training, this exposes the model to configurations where it must predict clean actions from noisy visual context, directly matching the few-step or single-step inference regime. As a result, we reduce the diffusion steps from four to one, cutting inference from ${\sim}350$ms to ${\sim}150$ms with minimal performance loss (Table~\ref{tab:tidying_one_step}). Moreover, the Flash formulation enables flexible training configurations—such as varying the noise sampling ratios of video and action—to better align training with different few-step or single-step inference regimes. In practice, we mainly apply Flash training as the final stage following the main \ourmethod model training.

\textbf{Action Chunk Smoothing.} To suppress high-frequency noise in generated actions, we upsample chunks to $2\times$ resolution, apply a Savitzky-Golay filter, and downsample to original resolution.

\subsubsection{Summary}

Table~\ref{tab:inference_speedup} summarizes cumulative speedups. System and implementation optimizations yield ${\sim}9\times$ speedup on H100 and ${\sim}16\times$ on GB200; adding \ourmethod-Flash achieves 38$\times$ on GB200, reducing latency from 5.7s to 150ms. With the exception of DiT caching and quantization, all system and implementation-level optimizations are mathematically equivalent to baseline and show no measurable performance degradation.

\begin{table}[h]
\centering
\begin{threeparttable}
\begin{tabular}{@{}lcc@{}}
\toprule
\textbf{Optimization} & \textbf{H100} & \textbf{GB200}\\ 
\midrule
Baseline & 1$\times$ & 1.1$\times$ \\
\midrule
\multicolumn{3}{l}{\textit{System-level}} \\
\quad + CFG Parallelism  & 1.9$\times$ & 1.8$\times$ \\
\quad + DiT Caching & 5.5$\times$  & 5.4$\times$ \\
\midrule
\multicolumn{3}{l}{\textit{Implementation-level}} \\
\quad + Torch Compile + CUDA Graphs & 8.9$\times$ & 10.9$\times$ \\
\quad + Kernel \& Scheduler Opts. & 9.6$\times$ & 14.8$\times$ \\
\quad + Quantization (NVFP4) & — & 16.6$\times$ \\ 
\midrule
\multicolumn{3}{l}{\textit{Model-level}} \\
\quad + \ourmethod-Flash & — & 38$\times$ \\
\bottomrule
\end{tabular}
\caption{\textbf{Cumulative inference speedups.} Each row includes all optimizations above it. Entries marked ``—'' indicate features not applicable to that hardware.}
\label{tab:inference_speedup}
\end{threeparttable}
\vspace{-0.1in}
\end{table}

\section{Experimental Setup}
We validate our main hypotheses about learning from diverse data on two robot embodiments: the AgiBot G1 mobile bimanual manipulator and the Franka single-arm robot. We pretrain separately for each embodiment, leaving multi-embodiment training for future work. For cross-embodiment experiments, we utilize both the YAM robot and human egocentric data. The experimental setup for AgiBot G1 is illustrated in Figure~\ref{fig:main_eval}.

We compare against two state-of-the-art Vision-Language-Action models (VLAs): GR00T N1.6~\citep{bjorck2025gr00t} and $\pi_{0.5}$~\citep{intelligence2025pi}. For each baseline, we evaluate two initialization strategies: (1) \textit{from-scratch}, using pretrained VLM weights without prior robot data training for a fair apple-to-apple comparison with \ourmethod, and (2) \textit{from-pretrained}, using official checkpoints pretrained on thousands of hours of cross-embodiment robot data. Both variants are then trained on identical data as \ourmethod: ${\sim}500$ hours of teleoperation data we collected for AgiBot G1, and DROID~\citep{khazatsky2024droid} for Franka. We keep the compute budget comparable across all methods by matching total batch size and gradient steps.\footnote{For from-pretrained baselines, this constitutes continual training on top of the official weights.}

\begin{figure}[ht!]
    \centering
    \begin{subfigure}[b]{0.48\textwidth}
        \centering
        \includegraphics[width=\textwidth]{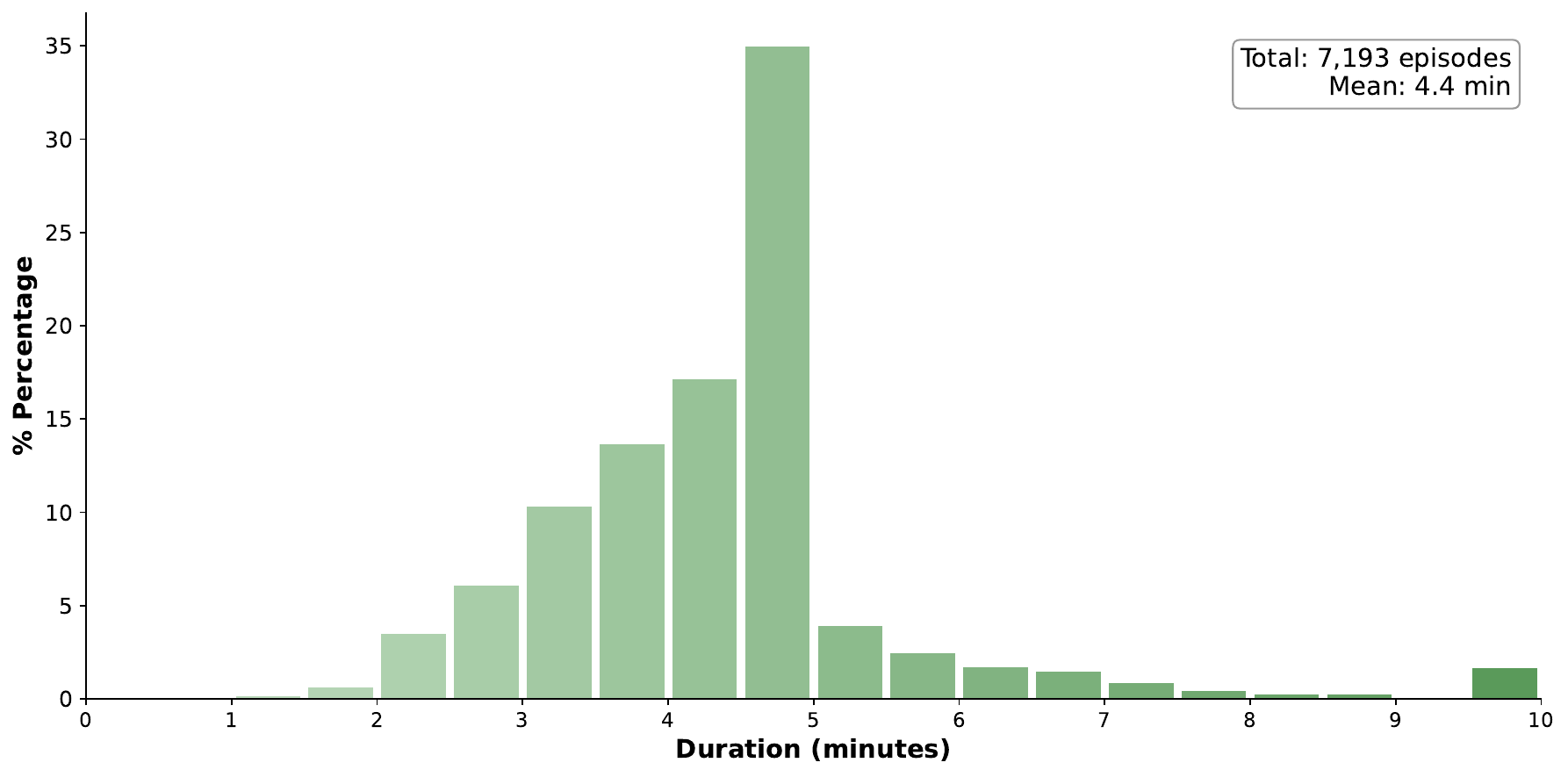}
        \caption{Episode Duration Distribution}
        \label{fig:subfig1}
    \end{subfigure}
    \hspace{1mm}
    \begin{subfigure}[b]{0.48\textwidth}
        \centering
        \includegraphics[width=\textwidth]{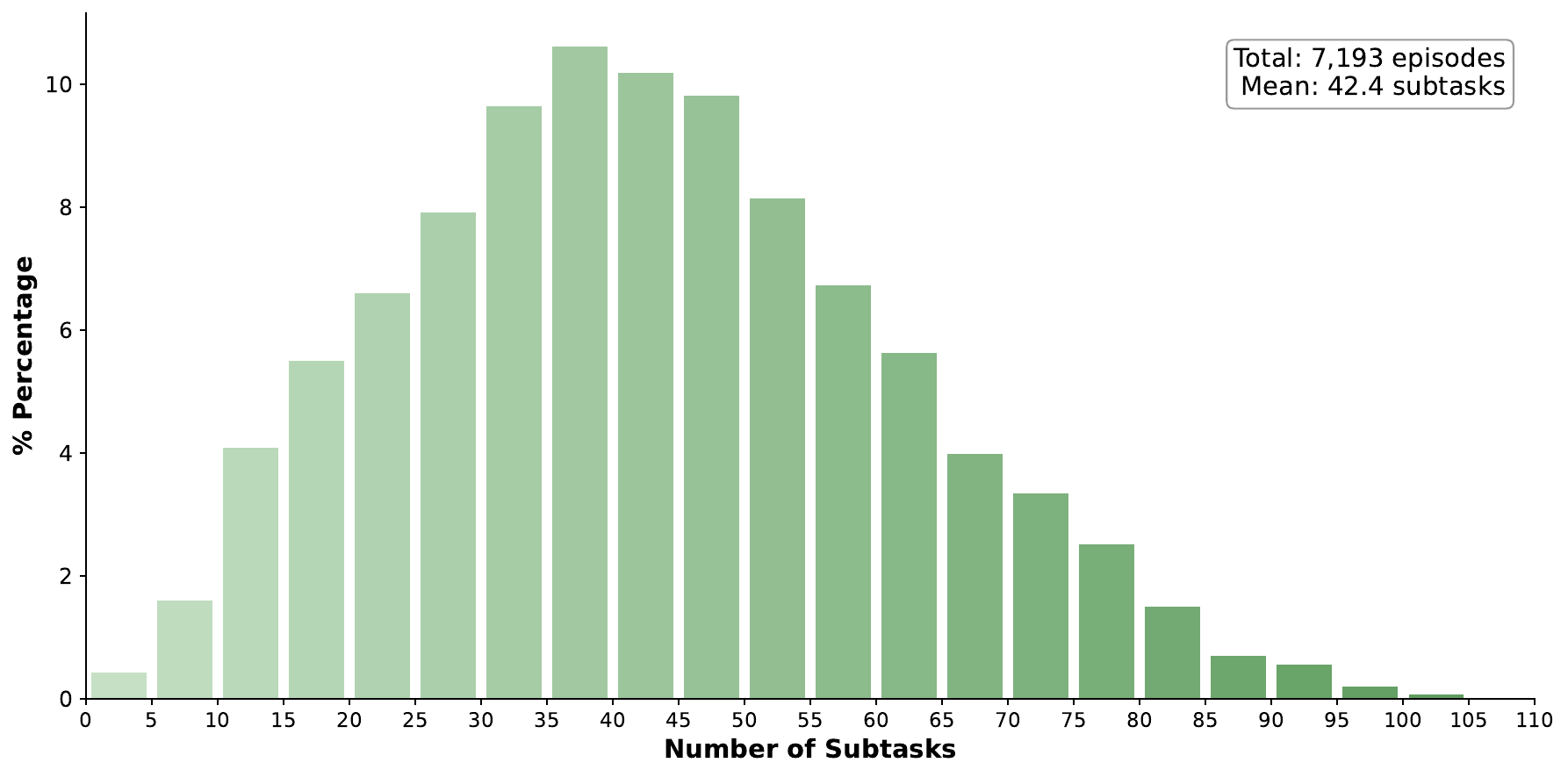}
        \caption{Subtask Count Distribution per Episode}
        \label{fig:subfig2}
    \end{subfigure}
    
    \vspace{1mm} 
    
    \begin{subfigure}[b]{\textwidth}
        \centering
        \includegraphics[width=\textwidth]{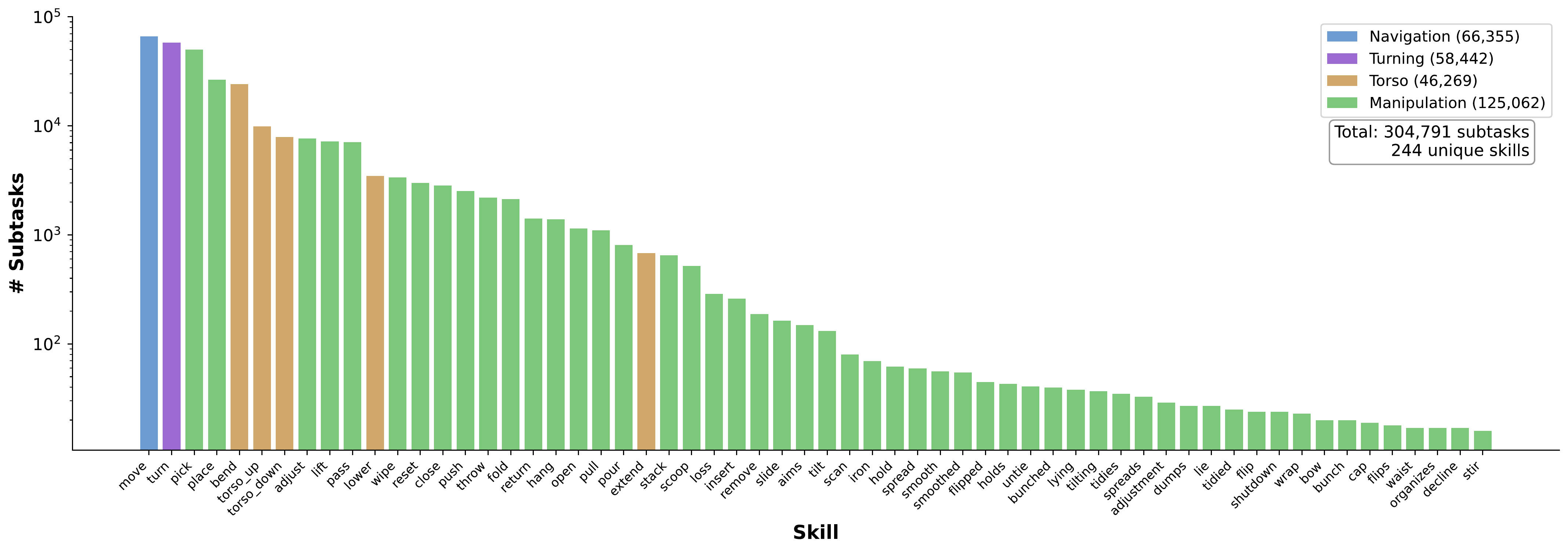}
        \caption{Skill Distribution}
        \label{fig:subfig3}
    \end{subfigure}
    
    \caption{\textbf{Distribution statistics} for the AgiBot pretraining corpus: episode durations, subtask density, and skill coverage across 7.2K episodes ($\sim$500 hours).}
    \label{fig:data_stats}
    \vspace{-0.07in}
\end{figure}

\subsection{Pretraining}
\textbf{Data.} Our data collection philosophy differs from that of existing VLAs. While recent works have shown that VLAs can learn effective policies from moderate-sized datasets, these approaches typically rely on structured, task-focused demonstrations to ensure consistent behavior. We hypothesize that learning to only predict actions without encoding the knowledge about future world states makes it challenging to leverage highly heterogeneous, non-repetitive data effectively, as the model must implicitly infer dynamics from noisy state-action pairs. In contrast, we hypothesize \ourmethod's world modeling objective enables effective learning from diverse demonstrations, allowing us to prioritize breadth and utility over repetition during data collection.

Using AgiBot G1, we collect approximately 500 hours of teleoperation data across 22 unique environments (see Figure \ref{fig:data_collection_envs}), including homes, restaurants, supermarkets, coffee shops, and offices—prioritizing task diversity and real-world utility over task-specific repetition. As shown in Figure~\ref{fig:data_stats}, each episode averages around 4.4 minutes and encompasses approximately 42 subtasks—significantly longer-horizon than typical robotic manipulation datasets \citep{khazatsky2024droid, walke2023bridgedata}. The skill distribution reflects real-world deployment requirements: navigation enables movement between workspaces, while torso adjustments allow interaction with objects at varying heights (shelves, cabinets). Additional details on the data collection pipeline are provided in Appendix \ref{appen:agibot_dc}.\footnote{We plan to open-source this dataset in upcoming releases. Some samples can be found at \url{https://dreamzero0.github.io/training_data_gallery}}

We also validate \ourmethod on the Franka single-arm robot using DROID~\citep{khazatsky2024droid}, one of the most heterogeneous publicly available robotic datasets to demonstrate the effectiveness of WAMs on diverse, open-source data and enables reproducibility prior to the release of our in-house AgiBot dataset. We open-source the checkpoint and inference code to run some DROID-sim evals in PolaRiS~\citep{jain2025polaris}.\footnote{Available at \url{https://github.com/dreamzero0/dreamzero}.}

\textbf{Training.} We use Wan2.1-I2V-14B-480P~\citep{wan2025wan}, a 14B image-to-video diffusion model, as the backbone for \ourmethod. We train for 100K steps with a global batch size of 128 for AgiBot and 100K steps with a global batch size of 128 for DROID datasets. We update all DiT blocks, the state encoder, action encoder, and action decoder, while freezing the text encoder, image encoder, and VAE.\footnote{We experimented with LoRA~\citep{hu2022lora} but found it led to suboptimal results.} For both datasets, we filter out idle actions and use relative joint positions as the default action representation. We also conduct some ablations (Section \ref{subsec:ablations}) where we initialize from Wan2.1-I2V-5B-480P to see the effect of model size (5B vs. 14B).

\begin{figure}[t!]
    \centering
    \includegraphics[width=1\textwidth]{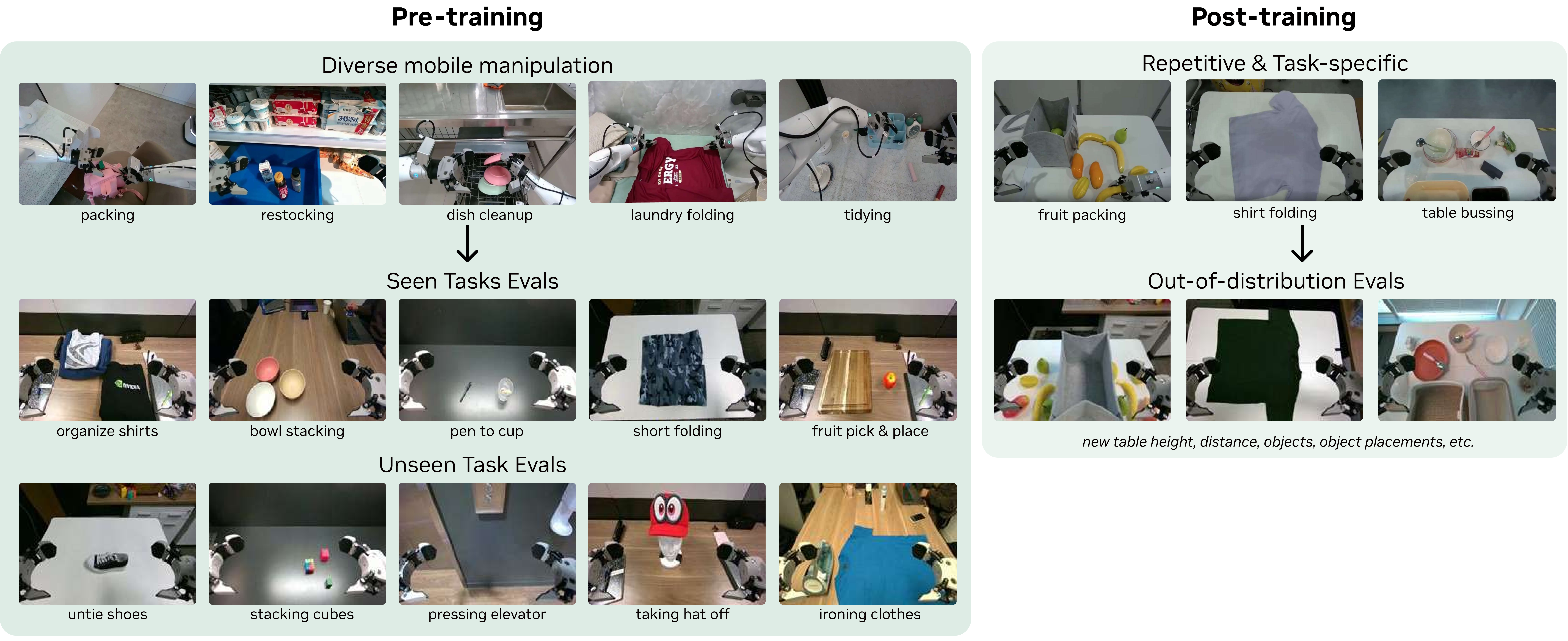}
    \caption{\textbf{AgiBot Evaluation Set-up.} We are first-citizens of generalization evals, where the default setting is \textit{unseen} environment and \textit{unseen} objects.}
    \label{fig:main_eval}
    \vspace{-0.05in}
\end{figure}

\textbf{Evaluation Protocol.} We evaluate models out of the box after pretraining. Our default evaluation setting is \textit{unseen environments, unseen objects}—because our pretraining and post-training data were collected in a different geographic location from our evaluation sites, every benchmark inherently tests out-of-distribution generalization rather than interpolation within the training distribution.
We evaluate on two categories: \textit{seen} and \textit{unseen} tasks. We define the granularity of a task as a combination of the motion required for the task and the object type. For example, if the training data contains folding a red-colored shirt and evaluate the model to fold a black-colored shirt with a different size, it is considered as a \textit{seen} task. On the other hand, if we evaluate the model to fold socks, it is considered as an unseen task because the motion required to fold socks is different from folding a shirt (See samples in Figure \ref{fig:main_eval}).

\textbf{AgiBot Evaluation Protocol.} For \textit{seen} tasks, we select 10 tasks from the pretraining distribution, including pick-and-place variants, stacking, wiping, and folding; we run 8 rollouts per task across 4 robots, each in different environments and different objects (80 rollouts total per checkpoint). We divide 10 seen tasks into three categories: PnP-Easy (Pick and place fruit, Wipe the mess, Take out fruit from bag), PnP-Hard (Pick and place fork/spoon, put the pen in pen holder, put the cup on the coaster, stack bowls/cups in a row), and Contact-Rich Manipulation (fold shirts, fold shorts, stack clothes). For \textit{unseen} tasks, we evaluate 10 tasks absent from training—such as ironing, painting, pulling carts, cube stacking, removing a hat from a mannequin, and untying shoe laces—with 8 rollouts per task across 4 robots (80 rollouts total per checkpoint). The full list of each evaluation rollout initial frame and prompt is provided in Appendix \ref{appen:agibot_evals}, and some evaluation rollouts can be found here.\footnote{\url{https://dreamzero0.github.io} for main evaluation rollouts and \url{https://dreamzero0.github.io/evals_gallery} for accumulation of unique evaluation rollouts.}

\textbf{DROID Evaluation Protocol.} We evaluate on 20 \textit{seen} tasks and 20 \textit{unseen tasks} (verbs absent from DROID), performing 2 rollouts per task, for a total of 80 evaluation rollouts across 40 tasks for each checkpoint. We compare \ourmethod against the publicly released $\pi_{0.5}$-DROID and an internally trained GR00T N1.6-DROID checkpoint. Object positions are fixed across checkpoints to ensure fairness. Each rollout is scored from 0 to 1.0 based on partial task completion; full details are provided in Appendix~\ref{appen:droid_evals}.

\subsection{Post-training}

Beyond pre-training, we evaluate whether WAMs improve fine-tuning performance on task-specific data using the AgiBot robot.

\textbf{Data.} We collect post-training data on three downstream tasks:
\begin{itemize}[leftmargin=*, itemsep=2pt, topsep=2pt]
    \item \textit{Shirt folding} (33 hrs): Fold a flattened t-shirt through 5 sequential stages. We randomize initial shirt position across 2 shirt types. 
    \item \textit{Fruit packing} (12 hrs): Pack 10 fruits from a table into a bag. We randomize fruit combinations and positions of fruits and bag. 
    \item \textit{Table bussing} (40 hrs): Clear 5 pieces of trash into a trash bin and 5 pieces of dishware (dish, bowl, fork, and spoon) into a dish bin. We randomize object types, combinations, and positions.
\end{itemize}

\textbf{Training.} We post-train for 50K steps per task. As in pretraining, we update all parameters except the text encoder, image encoder, and VAE.

\textbf{Evaluation Protocol.} We measure average task progress across 10 rollouts per task.
Task progress is defined as: (1) folding stages completed out of 5 for shirt folding, (2) fruits successfully packed out of 10 for fruit packing and (3) items cleared for table bussing. Following \citet{barreiros2025careful}, we apply an image overlay to the initial scene to reduce variance.

\section{Experimental Results}
\label{sec:result}
\subsection{Main Results}
\begin{figure}[htbp]
    \centering
    \includegraphics[width=1\textwidth]{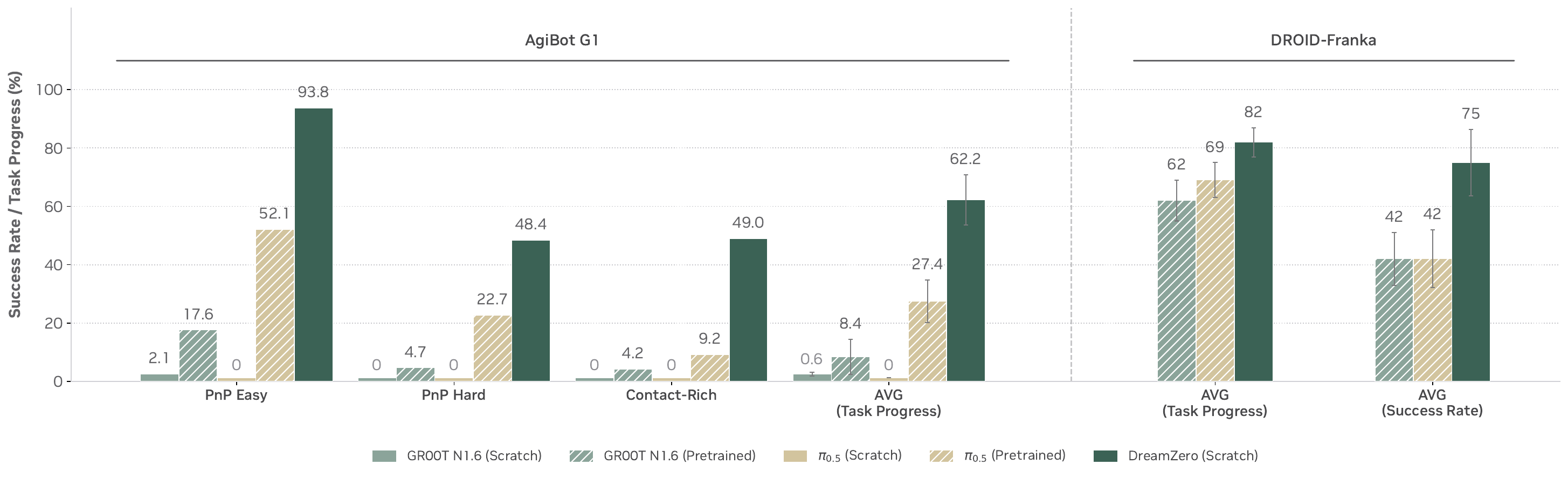}
    \caption{\textbf{Seen Task Evaluation.} \ourmethod effectively learns from diverse data and generalizes to new environments, outperforming VLAs across all task categories. VLAs trained \textit{from scratch} achieve near-zero success, while \textit{pretrained} VLAs show modest performance—likely benefiting from embodiment-specific knowledge acquired through repetitive demonstrations during pretraining.}
    \label{fig:seen}
\end{figure}
We evaluate the zero-shot generalization performance of \ourmethod compared to baseline models and investigate the following research questions:

\paragraph{Q1. Do WAMs learn better from diverse, non-repetitive data?}

We evaluate pretrained models out-of-the-box on tasks present in the pretraining data, but in zero-shot environments with unseen objects. Results are shown in Figure~\ref{fig:seen}.

On AgiBot G1, \textit{from-scratch} VLAs achieve near-zero task progress score across all categories. Even on simple pick-and-place tasks (PnP Easy), VLAs occasionally reach toward the correct object but fail to interact accurately with unseen objects in novel environments. In contrast, \ourmethod successfully learns from heterogeneous data, achieving 62.2\% average task progress—over 2$\times$ higher than the best \textit{pretrained} VLA baseline (27.4\%), despite those baselines being pretrained on thousands of hours of cross-embodiment robot data before continued training on our data mix. On DROID-Franka, we show a similar result as well; \ourmethod which is only trained on the DROID dataset outperforms pre-trained baseline models trained on multiple robot embodiment data. 

We attribute this gap to the joint video-action formulation: while VLAs require massive robot data to learn direct observation-to-a ction mappings, WAMs leverage video generation as a strong prior for action prediction, enabling effective learning of diverse data and generalization to unseen environments. Notably, we observe tight alignment between generated videos and real-world execution, even for suboptimal behaviors (Figure~\ref{fig:video_alignment}). Most \ourmethod failures stem from video generation errors rather than action prediction—the policy faithfully executes whatever trajectory the video predicts. This suggests that improvements to the video backbone would directly translate to better WAM performance. 

\begin{figure}[t!]
    \centering
    \includegraphics[width=1\textwidth]{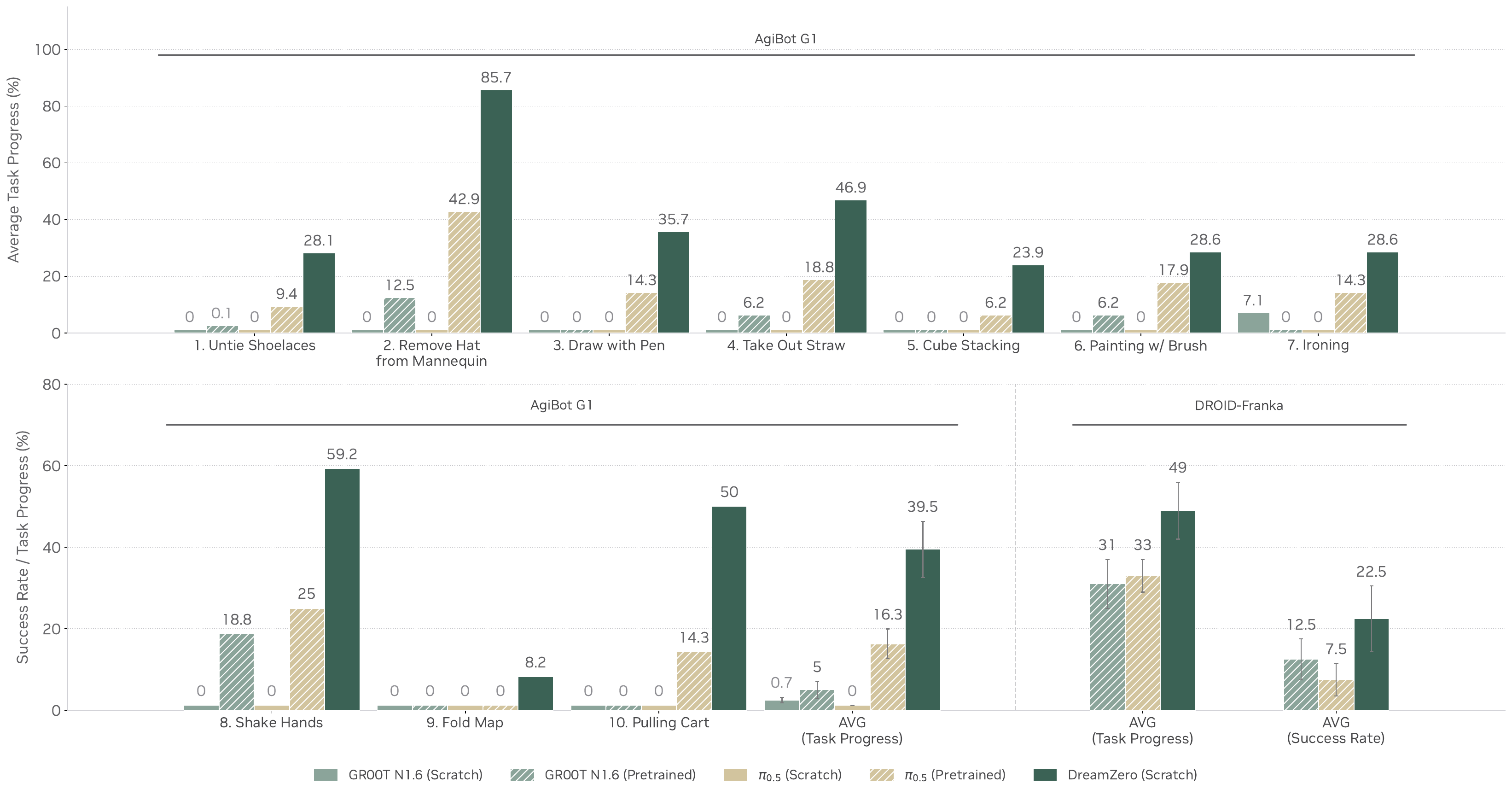}
    \caption{\textbf{Zero-shot Generalization to Unseen Tasks.} \ourmethod achieves non-trivial task progress on 10 tasks absent from training, while VLAs struggle across both embodiments.}
    \label{fig:unseen}
\end{figure}

\paragraph{Q2. Do WAMs generalize to unseen tasks?}

Figure~\ref{fig:unseen} evaluates generalization to 10 tasks entirely absent from the pretraining distribution, including untying shoelaces, ironing, painting with a brush, and shaking hands.

On AgiBot G1, \textit{from-scratch} VLAs achieve near-zero task progress ($<1\%$), while \ourmethod reaches 39.5\% on average—with strong performance on tasks like ``Remove Hat from Mannequin'' (85.7\%) and ``Shake Hands'' (59.2\%). \ourmethod also significantly outperforms pretrained VLA baselines (39.5\% vs.\ 16.3\%), even though those baselines may have encountered some of these tasks during cross-embodiment pretraining. Also on the DROID-Franka setup, \ourmethod significantly outperforms (49\% task progress, 22.5\% success rate) other pretrained baselines (31\% task progress, 12.5\% success rate for GR00T N1.6 and 33\% task progress, 7.5\% success rate for $\pi_{0.5}$).

Qualitatively, we observe that pretrained VLAs often reach toward objects and attempt grasping regardless of the instruction, suggesting they overfit to dominant training behaviors (e.g., pick-and-place) rather than understanding novel task semantics, accounting for their partial task progress despite failing to complete the intended tasks. In contrast, \ourmethod performs visual planning for unseen tasks and executes them successfully, with strong alignment between generated videos and real-world actions.

Beyond structured evaluation, we conduct free-form testing on over 100+ additional tasks, including ``Pop the ballon'' and ``Press elevator button'', by doing free-form prompting with verbal instructions.\footnote{Rollouts of these tasks are provided in \url{https://dreamzero0.github.io/evals_gallery/}.}

\begin{figure}[htbp]
    \centering
    \includegraphics[width=1\textwidth]{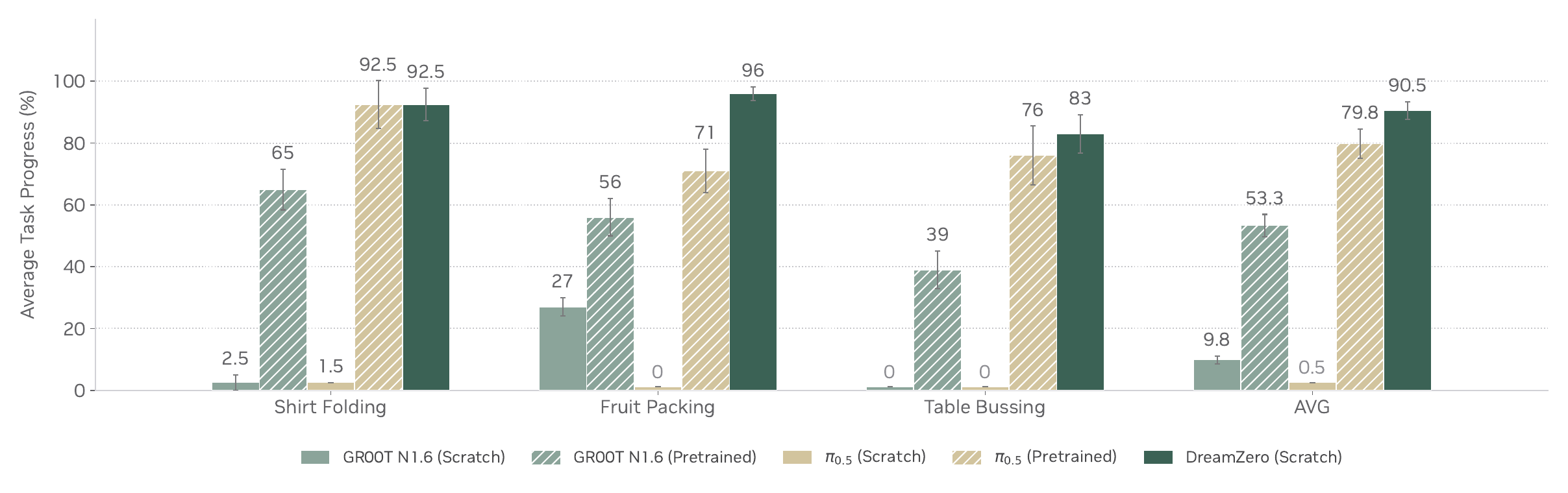}
    \caption{\textbf{Posttraining Results}. WAMs enable stronger post-training results across three tasks, indicating that environment generalization of \ourmethod is retained after post-training.}
    \label{fig:posttrain}
\end{figure}


\paragraph{Q3. Do WAMs improve post-training performance?}

We investigate whether WAMs retain their generalization even after fine-tuning on task-specific data. Figure~\ref{fig:posttrain} shows results on three tasks with varying distribution diversity.

\ourmethod matches or outperforms VLA baselines across all tasks: comparable performance on shirt folding and table bussing while significantly outperforming on fruit packing. Similar to the findings from Figure~\ref{fig:seen} and Figure~\ref{fig:unseen}, \textit{from-scratch} baselines fail to learn accurate motions to grasp the target objects; this means that \textit{from-scratch} VLAs tend to overfit to the training data and fail to generalize to scenarios where we vary the table height, table distance, objects, and object placements, mostly due to the evaluation site being in a different geographic location (see Figure \ref{fig:main_eval} for samples). Although pretraining on multiple robot embodiments with repetitive data largely boosts the post-training genearlization performance for pretrained baselines, \ourmethod still matches or outperforms pretrained VLA baselines without cross embodiment pretraining. Since we still evaluate on unseen environments for post-training, this implies that the environment generalization of \ourmethod is retained after post-training.

\paragraph{Q4. Do WAMs enable strong cross-embodiment transfer to unseen tasks?}

Having shown that WAMs generalize to unseen tasks (Figure~\ref{fig:unseen}), we now investigate whether this generalization can be further improved by leveraging video data from different embodiments performing the same tasks. Crucially, we use only the video prediction objective for the cross-embodiment data (no actions), while maintaining the joint video-action objective for the AgiBot pretraining data; the cross-embodiment data thus serves as additional visual experience to strengthen the world model's understanding of task dynamics and expected behavior.

\begin{figure}[ht!]
    \centering
    \includegraphics[width=1\textwidth]{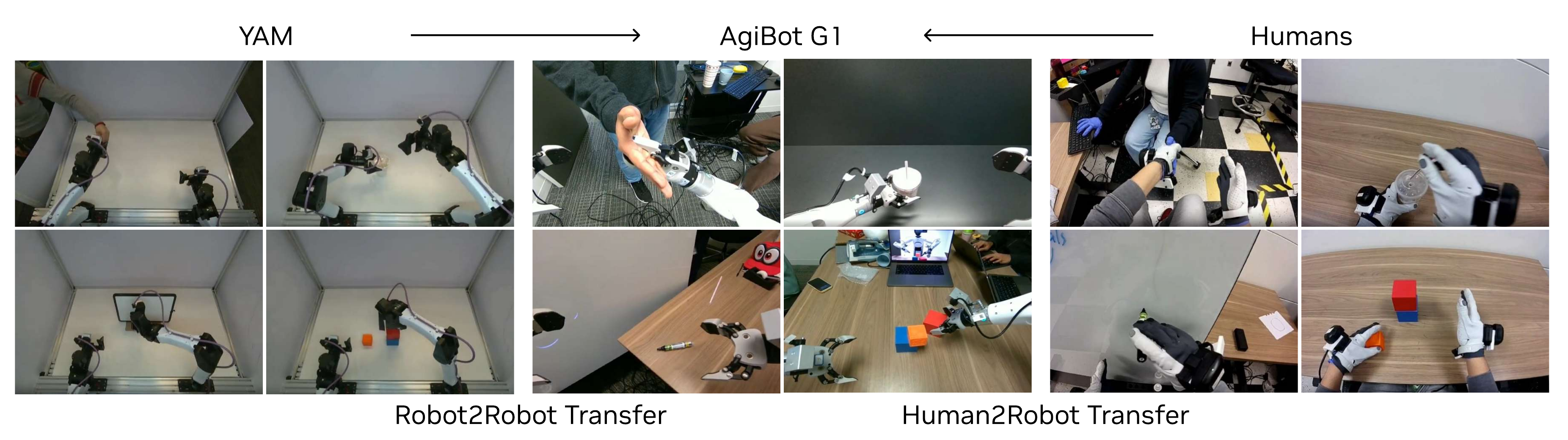}
    \caption{\textbf{Cross-Embodiment Transfer.} We explore robot-to-robot (YAM $\rightarrow$ AgiBot) and human-to-robot embodiment transfer to unseen tasks.}
    \label{fig:embodiment_transfer}
\end{figure}

We explore two settings (Figure~\ref{fig:embodiment_transfer}): (1) \textit{Robot-to-robot transfer} using the bimanual YAM robot, and (2) \textit{Human-to-robot transfer} using egocentric human demonstrations. For each setting, we collect 72 multi-view trajectories of the 9 unseen tasks (8 demonstrations per task, 20 minutes for YAM, 12 minutes for human).\footnote{We exclude \textit{Pulling Cart} task since data collection through teleoperation was infeasible with our bimanual YAM robot setup.} We then co-train from the \ourmethod-AgiBot checkpoint on a 1:1 mix with pretraining data for 10K steps. 

Results on the 9 unseen tasks (Table~\ref{tab:embodiment_transfer}) show that both transfer settings improve performance over the baseline \ourmethod. Robot-to-robot transfer yields the largest gain (38.3\% $\rightarrow$ 55.4\%), likely due to the narrower embodiment gap; both YAM and AgiBot are bimanual parallel grippers. Human-to-robot transfer also improves performance (38.3\% $\rightarrow$ 54.3\%), despite the larger morphological gap and dynamic egocentric viewpoints.

\begin{table}[H]
\small
\centering
\begin{tabular}{@{}ll@{}}
\arrayrulecolor{nvidiagreen} 
\toprule
\textbf{Method}  & \textbf{Task Progress}\\
\midrule
\ourmethod        & 38.3\% \scriptsize{$\pm$ 7.6\%} \\ 
\ourmethod + Human2Robot Transfer    & 54.3\% \scriptsize{$\pm$ 10.4\%} \\
\ourmethod + Robot2Robot Transfer & 55.4\% \scriptsize{$\pm$ 9.5\%} \\
\bottomrule
\end{tabular}
\caption{\textbf{Cross-Embodiment Transfer Results.} Average task progress on unseen tasks ($\pm$ standard error). Both transfer settings improve over baseline (result from Table \ref{fig:unseen}) using only 10--20 minutes of video-only demonstration data.}
\label{tab:embodiment_transfer}
\end{table}

These results point to a promising property of WAMs: unlike recent VLA approaches to embodiment transfer~\citep{kareer2025emergence, generalist2025gen0}, our method relies solely on visual information without action labels. While current success rates remain moderate, the consistent improvement from just 10–20 minutes of video-only data provides an early signal that cross-embodiment visual experience transfers meaningfully. This opens a potential scaling pathway: abundant human video data—orders of magnitude larger than robot datasets—could enable WAMs to acquire diverse skills without action annotation, pending further research into strengthening the transfer mechanism.

\paragraph{Q5. Do WAMs enable few-shot new embodiment adaptation?}

\begin{figure}[ht!]
    \centering
    \includegraphics[width=1\textwidth]{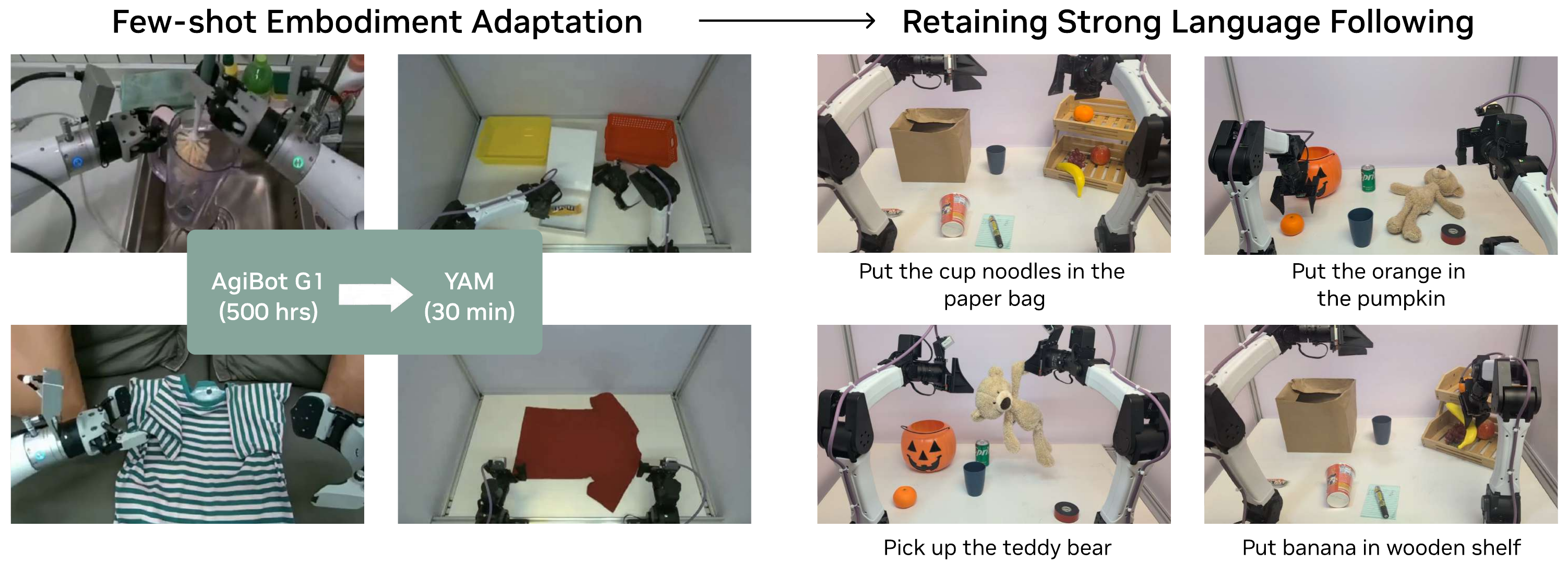}
    \caption{\textbf{Few-shot Embodiment Adaptation.} We explore few-shot embodiment adaptation by post-training on 30 minutes of new embodiment \textit{play} data and evaluating on pick-and-place variants requiring strong language following.}
    \label{fig:few_shot_transfer}
\end{figure}

We post-trained the \ourmethod-AgiBot checkpoint on a new bimanual manipulator (YAM robot) using only 55 trajectories across 11 unique tasks ($\sim$30 minutes of data).\footnote{We visualize the entire 30 minutes of play data \url{https://dreamzero0.github.io/yam_gallery/}} As illustrated in Figure~\ref{fig:few_shot_transfer}, despite limited data and diversity, the post-trained policy retains strong language following ability, even generalizing to novel objects unseen during training, including pumpkins, teddy bears, pens, cup noodles, and paper bags. Even with minimal data, we observe tight video-action alignment, demonstrating very efficient cross-embodiment transfer. 

We hypothesize that two factors enable this efficiency: (1) the visual similarity of AgiBot G1 and YAM embodiment (both equipped with bi-manual parallel grippers), and (2) more fundamentally, learning an implicit IDM from predicted videos may be inherently more sample-efficient than direct policy learning—the model only needs to learn the mapping from visual futures to actions, while leveraging the pretrained video model's existing understanding of physical dynamics. Consistent with our AgiBot findings, failures primarily stem from video prediction errors rather than action extraction, suggesting that increasing task diversity during post-training could further improve performance.\footnote{In this specific post-training experiment, we only utilized 11 short, global language annotations unique for each task. We hypothesize diversifying the language can also enable stronger tranfer, but leave further investigation to future work.}

\paragraph{Q6. Does \ourmethod-Flash maintain performance with fewer denoising steps?}

We evaluate whether \ourmethod-Flash can maintain task performance under aggressive single-step denoising. As shown in Table~\ref{tab:tidying_one_step}, reducing \ourmethod from 4 denoising steps to 1 step drops task progress substantially (83\% $\rightarrow$ 52\%) on the table bussing task. In contrast, \ourmethod-Flash achieves a higher average success rate (74\%) at single-step inference, sitting only 9\% below the 4-step baseline while being ${\sim}2\times$ faster. This suggests that decoupled noise scheduling offers a more effective speed–accuracy trade-off for real-time deployment.

\begin{table}[H]
\small
\centering
\begin{tabular}{lcccc}
\arrayrulecolor{nvidiagreen} 
\toprule
\textbf{Method} & \textbf{Denoising steps}  & \textbf{Task Progress} & \textbf{Inference speed} & \textbf{$\times$ Speed up}\\
\midrule
\ourmethod       & 4 & 83\% \,\scriptsize{$\pm$ 6.1\%} & 350ms & 1\\ 
\ourmethod       & 1 & 52\% \,\scriptsize{$\pm$ 10.2\%} & 150ms & 2.33$\times$\\
\ourmethod-Flash & 1 & 74\% \,\scriptsize{$\pm$ 10.1\%} & 150ms & 2.33$\times$\\
\bottomrule
\end{tabular}
\caption{\textbf{\ourmethod-Flash Evaluation.} Task progress on table bussing with varying denoising steps ($\pm$ standard error). \ourmethod-Flash recovers most of the 4-step performance using only 1 denoising step.}
\label{tab:tidying_one_step}
\end{table}

\subsection{Model and Data Ablations}
\label{subsec:ablations}

We conduct ablations to isolate the contributions of data diversity, model scale, and architecture. Due to computational constraints, all ablation models are trained with 50K steps and batch size 32, and evaluated on \textit{PnP Easy} tasks for consistent comparison.

\begin{table*}[ht!]
\centering
\small
\arrayrulecolor{nvidiagreen}
\setlength{\tabcolsep}{10pt}
\begin{tabular}{@{}llccc@{}} 
\toprule
& \textbf{Architecture} & \textbf{Model Size} & \textbf{Data} & \textbf{Task Progress} \\
\midrule
\multicolumn{5}{l}{\textit{Q1. Data Diversity}} \\
& \ourmethod (AR) & 14B & Repetitive & 33\%\scriptsize{\,$\pm$\,4.2\%} \\
& \ourmethod (AR) & 14B & Diverse    & 50\%\scriptsize{\,$\pm$\,6.3\%} \\
\midrule
\multicolumn{5}{l}{\textit{Q2. Model Scale}} \\
& \ourmethod (AR) & 5B  & Diverse & 21\%\scriptsize{\,$\pm$\,4.2\%} \\
& \ourmethod (AR) & 14B & Diverse & 50\%\scriptsize{\,$\pm$\,6.3\%} \\
& VLA             & 5B  & Diverse & \phantom{5}0\%\scriptsize{\,$\pm$\,0.0\%} \\
& VLA             & 14B & Diverse & \phantom{5}0\%\scriptsize{\,$\pm$\,0.0\%} \\
\midrule
\multicolumn{5}{l}{\textit{Q3. Architecture (Bidirectional vs. AR)}} \\
& \ourmethod (BD) & 14B & Diverse & 50\%\scriptsize{\,$\pm$\,14.4\%} \\
& \ourmethod (AR) & 14B & Diverse & 50\%\scriptsize{\,$\pm$\,6.3\%} \\
\bottomrule
\end{tabular}
\caption{\textbf{Model and Data Ablations.} Task progress on PnP Easy tasks ($\pm$ standard error). AR = autoregressive, BD = bidirectional. All models trained with 50K steps and batch size 32.}
\label{tab:ablation}
\end{table*}

\paragraph{Q1. Does data diversity improve generalization?}

We compare \ourmethod trained on 500 hours of \textit{diverse} data versus 500 hours of \textit{repetitive} data, where the latter contains 70 tasks with many repeated demonstrations per task using similar object positions and configurations. As shown in Table~\ref{tab:ablation}, diverse data substantially improves generalization (33\% → 50\%), even on simple pick-and-place tasks. We hypothesize this reflects WAMs' learning dynamics: since video prediction is largely inherited from pretraining, the key challenge is learning inverse dynamics. A robust IDM requires diverse state-action correspondences across varied contexts, which repetitive data inherently lacks.

\paragraph{Q2. Does WAM performance scale with model size?}

For VLAs, scaling model size improves semantic reasoning but not necessarily action prediction. We find that WAMs exhibit clearer scaling behavior: the 14B model significantly outperforms the 5B model (50\% vs.\ 21\%), with the smaller model prone to visual hallucinations that propagate to erroneous actions.

To ensure fair comparison, we also scale VLA baselines to match \ourmethod's size by initializing from 8B and 32B pretrained VLMs~\citep{yang2025qwen3}, truncating to the first half of the transformer blocks, and attaching DiT-based action modules following \citet{bjorck2025gr00t}. As shown in Table~\ref{tab:ablation}, larger VLAs still fail to learn from diverse data (0\% task progress), often hovering near objects without making contact. This suggests that scaling model capacity alone does not address VLAs' difficulty with diverse data distributions.

\paragraph{Q3. Does autoregressive architecture outperform bidirectional?}

We compare \ourmethod's autoregressive (AR) architecture against a bidirectional (BD) variant. While task progress is similar (Table~\ref{tab:ablation}), the AR model produces substantially smoother motions—backpropagating through entire action sequences enables better temporal consistency. Additionally, AR inference is 3--4$\times$ faster due to KV caching.

\section{Discussion and Future Work}
\textbf{Scaling Laws of WAMs.} We have identified that leveraging larger video backbone model and training on diverse data boosts downstream performance in Table \ref{tab:ablation}. However, we still have lacking evidence for scaling laws for robot foundation models, specifically for WAMs. Similar to scaling laws for language models \citep{kaplan2020scaling}, scaling laws for WAMs depending on the model size, dataset size, and the training compute need to be explored to determine the optimal configuration to extract the maximal capability of WAMs. We expect that the tendency of scaling WAMs to be different from VLAs, showing a more direct scaling law for actions. We leave deep investigation on scaling laws for WAMs as future work. 

\textbf{Learning from In-the-wild Human Data.} Although we have investigated leveraging egocentric human data to boost the performance on unseen tasks (Section \ref{sec:result}), our experiments are still constrained to small scale in-lab data (only 12 minutes). Recently, a large amount of human video data has been released that has more diverse distribution compared to robot data \citep{grauman2022ego4d, hoque2025egodex, chen2026action100m}. Since WAMs are pretrained on diverse internet video data, we hypothesize that leveraging large-scale egocentric human video data that are related to robot manipulation tasks would lead to stronger transfer to downstream robot tasks compared to current VLAs. We leave this direction as future work.

\textbf{Faster Inference.} Through model and system optimizations, we enable \ourmethod to run at 7Hz using 2 GB200s. However, compared to current VLAs which runs up to over 20Hz on consumer GPUs, \ourmethod is still computationally expensive due the large parameter size and the iterative denoising nature of video models. In the future, if smaller video backbone models also have strong generalization capability, WAMs could potentially be utilized as a real-time System 1 model on a lightweight edge device.

\textbf{Long-horizon Reasoning.} Current \ourmethod architecture functions primarily as a System 1 model. Although \ourmethod has a concept of visual memory, it is currently short-horizon (6 seconds). Robust long-horizon execution will require either a System 2 planner or WAMs with significantly extended context windows. For the former, both modular dual-system architectures \citep{shi2025hi} and unified approaches \citep{deng2025emerging} offer promising directions. For the latter, techniques from video-based world models that maintain coherent generation over extended horizons \citep{genie3, hyworld2025} could be adapted to expand WAM context length.

\textbf{High-Precision Tasks.} While \ourmethod generalizes broadly across tasks and environments, it inherits limitations common to behavior cloning on tasks requiring sub-centimeter precision, such as key insertion or fine assembly. Our diverse pretraining strategy prioritizes breadth, which may underrepresent the dense demonstrations needed for these high-precision manipulation. That said, recent work~\citep{kim2026cosmos} showed promising results that WAMs may actually hold an advantage for high-precision manipulation tasks with millimeter tolerance, an encouraging signal that the trade-off between broad generalization and fine-grained dexterity may be reconcilable with further investigation.

\textbf{Embodiment Design for WAMs.} We hypothesize two key factors will shape the best robot embodiments for future WAM development: (1) \textit{Degrees of freedom}: Higher-DOF robots will require more play data to learn an accurate implicit IDM, as the mapping from visual futures to motor commands grows combinatorially with kinematic complexity. Quantifying the accuracy of implicit IDMs remains a challenge. (2) \textit{Human similarity}: Embodiments that more closely resemble humans—particularly humanoids with dexterous manipulation capabilities—may transfer more efficiently despite higher DOF, as they can leverage both the motion priors from video pretraining and the massive scale of human egocentric videos. These factors pull in opposite directions—yet human-like embodiments may win out by trading mechanical simplicity for access to web-scale human data—the fuel for next-generation robot foundation models.

\section{Acknowledgment}
This work would not have been possible without our incredible robot operators: Alec Nagal, Inho Rha, Ava Yazdanfar, Sally Huynh, Rachit Deo, Alaa Eltayeb, Andres Rocha, Brian Dang, Cher Choi, Cristaldo Campos, Ethan Sushil Dhilpe, Jeremy Chimienti, Leilee Naderi, Manish Shah, Manoj Hallegere, Nick Aguilar, Paul Truong, Rahul Sampagaon, Shreya Raj, Nadia Laswi, Amitoj Sandhu, Omkaar Buddhikot, and Wesley Durbano. We also thank Pranav Atreya for their support on integrating DreamZero-Droid into RoboArena Benchmark. We also thank Danyi Chen, Ming-Yu Liu, and Spencer Huang for their continuous support, and Nishanth Kumar, KR Zentner, Fernando Castaneda Garcia-Rozas, Zhengyi Luo, Yunfan Jiang, and Max Li for fruitful discussions.

\clearpage
\appendix
\section{Comparison with Alternative World Model Architectures}
\label{appen:world_model_comparison}

The term ``world model'' encompasses a broad family of approaches beyond video-based prediction. Here we discuss how WAMs differ from these alternatives.

\textbf{Latent-Space World Models.} Joint Embedding Predictive Architectures (JEPAs)~\citep{lecun2022path, assran2023self, bardes2024vjepa, assran2025vjepa2} predict future states in abstract latent spaces rather than pixel space, offering computational efficiency and the ability to discard unpredictable details. V-JEPA 2~\citep{assran2025vjepa2} demonstrates this approach on robotics, achieving zero-shot planning on manipulation tasks after post-training on 62 hours of robot data. Similarly, the Dreamer series~\citep{hafner2019dream, hafner2020mastering, hafner2023mastering, hafner2025trainingagentsinsidescalable} learns compact latent dynamics models for model-based reinforcement learning. However, these approaches model dynamics as $p(s_{t+1} | s_t, a_t)$—predicting next state conditioned on current state and action—and thus require goal-conditioned planning or search at test time to yield trajectories.

\textbf{3D Point Cloud World Models.} PointWorld~\citep{huang2025pointworld} unifies state and action in a shared 3D spatial domain, predicting scene dynamics as 3D point flows conditioned on robot actions. This formulation enables embodiment-agnostic learning and real-time integration with model-predictive control (MPC). However, like latent world models, it requires explicit optimization (e.g., MPPI sampling) at inference to generate action trajectories.

\textbf{Key Distinction.} These alternative approaches share a common characteristic: they model forward dynamics and require separate inverse dynamics models or explicit planning/search procedures at deployment. In contrast, WAMs jointly model $p(\mathbf{o}_{t:t+H}, \mathbf{a}_{t:t+H} | \mathbf{o}_{0:t}, \mathbf{c})$, directly producing action trajectories aligned with predicted visual futures without test-time optimization. This enables real-time closed-loop control at 7Hz—a frequency that would be challenging for search-based methods—while inheriting rich spatiotemporal priors from video pretraining.

\section{Bidirectional vs. Autoregressive WAMs}
\label{appen:bidir_ar}
\begin{figure}[ht!]
    \centering
    \includegraphics[width=1\textwidth]{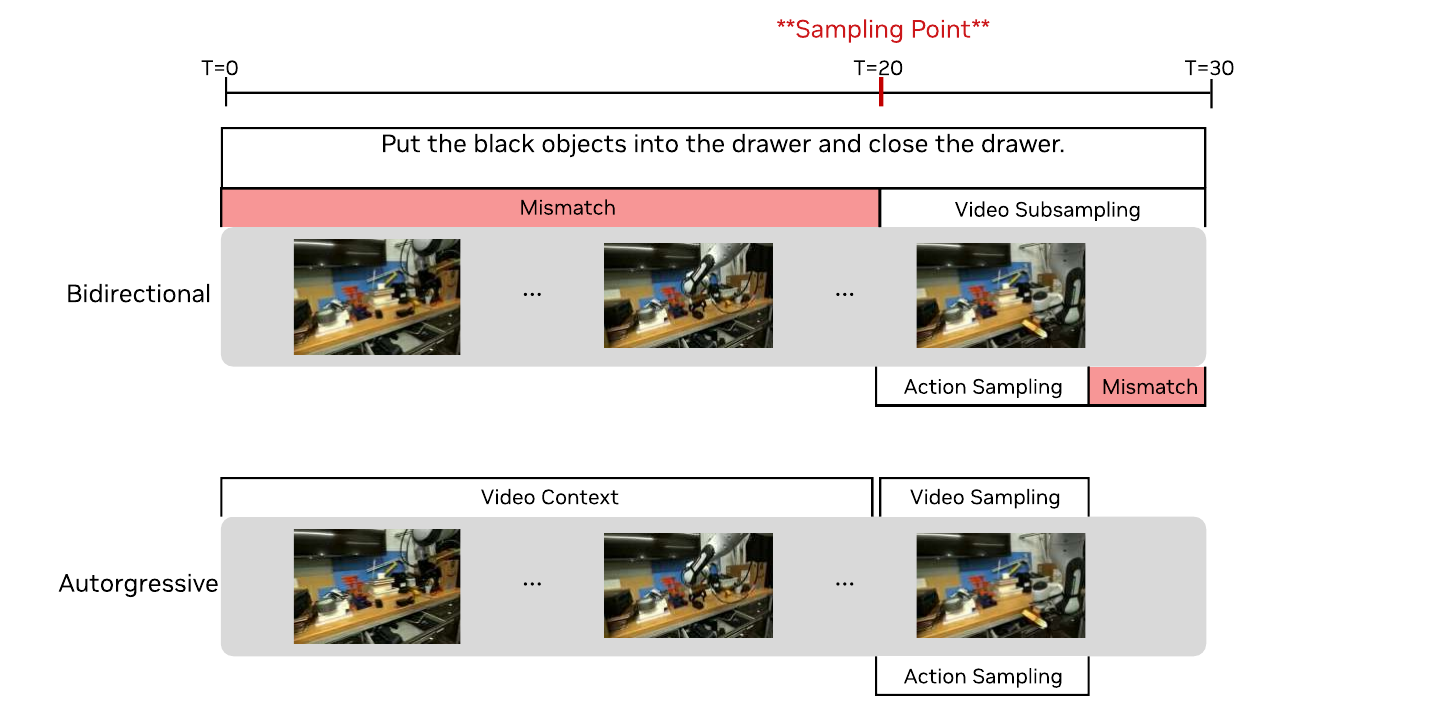}
    \caption{\textbf{Bidirectional vs. Autoregressive WAMs.} When the sampling point falls mid-task (T=20), bidirectional WAMs must subsample video to align with the language caption, distorting native FPS and degrading video-action alignment. Autoregressive WAMs avoid this trade-off by conditioning on video context, preserving both language-video correspondence and native frame rate.}
    \label{fig:bidiraug}
\end{figure}
Figure~\ref{fig:bidiraug} illustrates a key challenge for bidirectional WAMs when training closed-loop systems. Given a language annotation in a given long-horizon demonstration, the model must learn that the instruction corresponds to a specific video interval. In bidirectional architectures, without subsampling, the model receives a language instruction (e.g., 'put the black objects into the drawer') but usually generates video covering only a fraction of the task interval, causing the language to describe actions not yet visible in the predicted frames. This mismatch leads to significant language-following degradation, as measured via an internal video prediction benchmark.

A natural solution is to subsample the video to match the task caption interval. However, this creates a new problem for closed-loop training: we need the model to receive observations from arbitrary points within a task—not just the beginning. When the sampling point falls mid-task (e.g., T=20 in Figure~\ref{fig:bidiraug}), subsampling distorts the native FPS, making video-action alignment substantially harder to learn.

Autoregressive WAMs can sidestep this dilemma. By conditioning on video context rather than subsampling, they preserve both the language-video correspondence and the native frame rate, enabling tight alignment across all three modalities: language, video, and action.

\section{Model and Training details}
\label{appen:model_detail}
\begin{figure}[htbp]
    \centering
    \includegraphics[width=0.8\textwidth]{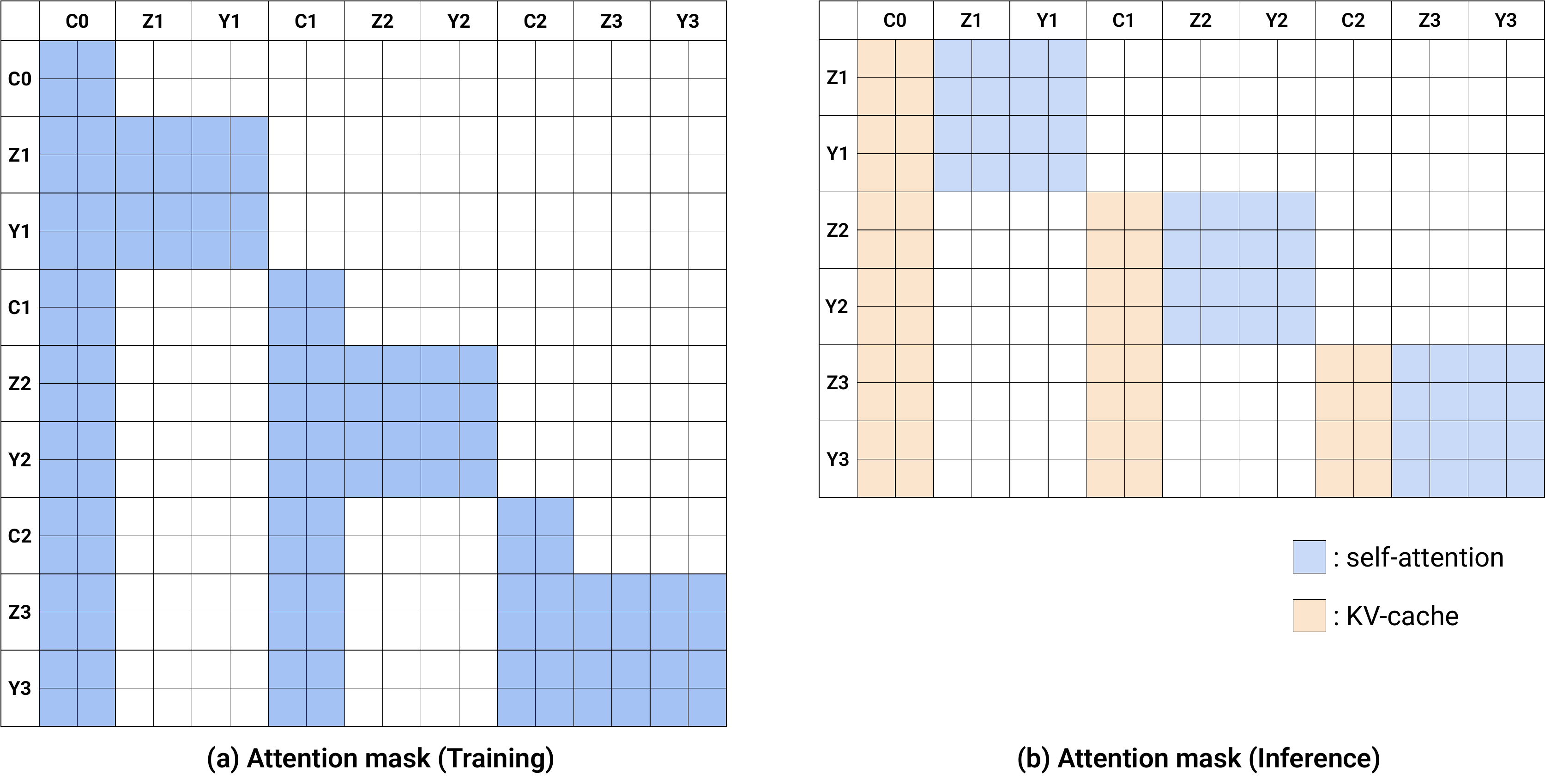}
    \caption{\textbf{Attention startegy of \ourmethod.} (a) QKV Self-Attention mask for DreamZero training. Y-axis shows the Query (Q) and X-axis shows the Key/Value (KV). Given conditioning frames (C0, C1, C2), we train the model to predict the velocities of next frames (Z1, Z2, Z3) and actions (Y1, Y2, Y3). (b) During inference, we compute the KV-cache of conditional frames and concatenate them to predict the action and frames. For example, Y3 (action) is able to attend to C0, C1, and C2, taking into account previous visual observations as history to predict the current actions during both training and inference. Note that the C0, C1, and C2 during inference is replaced with the GT observations.}
    \label{fig:attn_mask}
\end{figure}

We visualize the attention mask for training and inference in Figure~\ref{fig:attn_mask}. For \ourmethod, we set each chunk as $K=2$ latent frames. From preliminary results, we have observed that $K=2$ outperforms $K=1$ empirically. We set the number of chunks $M=4$ by default. If the trajectory length is shorter than $M=4$, $M$ can be smaller than 4.
For Agibot training data, the video is sampled at 5FPS ratio and action is sampled at 30Hz. We use the action horizon of $H=48$. Therefore, the video and action span 1.6 seconds per chunk. For DROID training, the video is sampled at 5FPS ratio and action is sampled at 15Hz. We use the action horizon of $H=24$. Therefore, same as Agibot, the video and action span 1.6 seconds per chunk. The maximum context length is 8 latent frames (4x2), which is equivalent to 33 raw frames, spanning 6.6 seconds. We leave increasing the visual context for WAMs as future work.

\begin{figure*}[t!]
    \centering
    \begin{minipage}[t]{0.48\textwidth}
        \begin{algorithm}[H] 
            \caption{\ourmethod Training (Flow Matching)}
            \label{alg:dreamzero_train}
            \begin{algorithmic}[1]
                \State \textbf{Input:} Dataset $\mathcal{D}$, Text condition $\mathbf{c}$
                \State \textbf{Hyperparams:} Number of chunks $M$
                \State \textbf{Model:} $\mathbf{u}_{\theta}$ (Joint Video-Action DiT)
                \While{not converged}
                    \State Sample trajectory $\tau \sim \mathcal{D}$
                    \State Encode video to clean latents $\mathbf{z}_{1}^{1:M}$, normalize actions $\mathbf{a}_{1}^{1:M}$
                    \State Split $\tau$ into $M$ chunks
                    
                    \For{$k = 1, \dots, M$} \Comment{Chunk-wise Training}
                        \State Define clean context $\mathcal{C}_{k} \leftarrow \{(\mathbf{z}_1^j, \mathbf{a}_1^j)\}_{j=1}^{k-1}$ \Comment{TF History}
                        \State Sample timestep $t_k \sim \mathcal{U}(0,1)$
                        
                        \State \textit{// Optional: DreamZero-Flash Decoupling}
                        \If{Flash Mode}
                            \State $t_{vid} \sim \text{Beta}(7, 1), \ t_{act} \sim \mathcal{U}(0,1)$
                        \Else
                            \State $t_{vid} \leftarrow t_k, \ t_{act} \leftarrow t_k$
                        \EndIf
                        
                        \State Sample noise $\mathbf{z}_{0}^{k}, \mathbf{a}_{0}^{k} \sim \mathcal{N}(\mathbf{0}, \mathbf{I})$
                        \State Interpolate (Eq. 2):
                        \State \quad $\mathbf{z}_{t}^{k} \leftarrow t_{vid} \mathbf{z}_{1}^{k} + (1-t_{vid}) \mathbf{z}_{0}^{k}$
                        \State \quad $\mathbf{a}_{t}^{k} \leftarrow t_{act} \mathbf{a}_{1}^{k} + (1-t_{act}) \mathbf{a}_{0}^{k}$
                        
                        \State Predict velocity:
                        \State \quad $\mathbf{v}_{pred} \leftarrow \mathbf{u}_{\theta}([\mathbf{z}_{t}^{k}, \mathbf{a}_{t}^{k}]; \mathcal{C}_{k}, \mathbf{c}, \mathbf{q}_{k}, t_k)$
                        
                        \State Target vel. $\mathbf{v}^{k} \coloneqq [\mathbf{z}_{1}^{k}, \mathbf{a}_{1}^{k}] - [\mathbf{z}_{0}^{k}, \mathbf{a}_{0}^{k}]$
                        \State Loss $\mathcal{L} \leftarrow ||\mathbf{v}_{pred} - \mathbf{v}^{k}||^2$ \Comment{Eq. 3}
                        \State Update $\theta \leftarrow \theta - \eta \nabla \mathcal{L}$
                    \EndFor
                \EndWhile
            \end{algorithmic}
        \end{algorithm}
    \end{minipage}
    \hfill 
    \begin{minipage}[t]{0.48\textwidth}
        \begin{algorithm}[H]
            \caption{\ourmethod Inference (Closed-Loop Control)}
            \label{alg:dreamzero_inference}
            \begin{algorithmic}[1]
                \State \textbf{Input:} Instruction $\mathbf{c}$, Initial Image $\mathbf{o}_{\text{init}}$, State $\mathbf{q}_{\text{init}}$
                \State \textbf{Hyperparams:} Steps $N$, Cache Thresh $\epsilon$
                \State \textbf{Init:} $\mathcal{KV} \leftarrow \emptyset$, $\mathbf{v}_{\text{prev}} \leftarrow \emptyset$, $\mathbf{q}_{\text{curr}} \leftarrow \mathbf{q}_{\text{init}}$
                
                \State \textcolor{gray}{\textit{// 1. Prefill Cache (Context Phase, $t=0$)}}
                \State $\mathbf{z}_{\text{init}} \leftarrow \text{VAE}(\mathbf{o}_{\text{init}})$
                \State \textcolor{gray}{\textit{// Pass clean video, no action/state}}
                \State $(\cdot, \cdot, \mathcal{KV}) \leftarrow \mathbf{u}_{\theta}([\mathbf{z}_{\text{init}}, \emptyset]; \mathcal{KV}, \mathbf{c}, \emptyset,$
                \Statex \hskip 8.5em $t=0, \text{update=True})$
                
                \State \textcolor{gray}{\textit{// 2. Autoregressive Loop}}
                \While{task not done}
                    \State Sample $\mathbf{x}_0 = [\mathbf{z}_0, \mathbf{a}_0] \sim \mathcal{N}(\mathbf{0}, \mathbf{I})$ \Comment{Noise at $t=0$}
                    
                    \State \textit{// Joint Denoising (Flow Matching $t: 0 \to 1$)}
                    \For{$i = 0 \dots N-1$} 
                        \State $t_{i}, t_{i+1} \leftarrow \text{Scheduler}(i, N)$
                        
                        \State \textit{// Optimization: DiT Caching}
                        \If{$\mathbf{v}_{\text{prev}} \neq \emptyset \textbf{ and } \text{CosSim}(\mathbf{v}_{\text{prev}}, \mathbf{v}_{\text{last}}) > \epsilon$}
                            \State $\mathbf{v}_i \leftarrow \mathbf{v}_{\text{prev}}$ 
                        \Else
                            \State $(\mathbf{v}^{\text{vid}}_i, \mathbf{v}^{\text{act}}_i, \cdot) \leftarrow \mathbf{u}_{\theta}(\mathbf{x}_{t_i}; \mathcal{KV}, \mathbf{c}, \mathbf{q}_{\text{curr}},$
                            \Statex \hskip 8.5em $t_i, \text{update=False})$
                            \State $\mathbf{v}_i \leftarrow [\mathbf{v}^{\text{vid}}_i, \mathbf{v}^{\text{act}}_i]$
                            \State $\mathbf{v}_{\text{prev}} \leftarrow \mathbf{v}_i$
                        \EndIf
                        
                        \State \textit{// Solver Step}
                        \State $\mathbf{x}_{t_{i+1}} \leftarrow \mathbf{x}_{t_i} + \text{Step}(\mathbf{v}_i, t_i, t_{i+1})$
                    \EndFor
                    
                    \State \textcolor{gray}{\textit{// 3. Execution \& Cache Update}}
                    \State $\mathbf{\hat{a}} \leftarrow \text{Filter}(\mathbf{x}_1^{\text{action}})$ \Comment{Clean Action $t=1$}
                    \State \textbf{Async Execute} $\mathbf{\hat{a}}$ on Robot
                    
                    \State \textit{// Critical: Inject Ground Truth ($t=0$)}
                    \State Real observation $\mathbf{o}_{\text{real}}, \mathbf{q}_{\text{real}}$
                    \State $\mathbf{q}_{\text{curr}} \leftarrow \mathbf{q}_{\text{real}}, \quad \mathbf{z}_{\text{real}} \leftarrow \text{VAE}(\mathbf{o}_{\text{real}})$
                    \State $(\cdot, \cdot, \mathcal{KV}) \leftarrow \mathbf{u}_{\theta}([\mathbf{z}_{\text{real}}, \emptyset]; \mathcal{KV}, \mathbf{c}, \emptyset,$
                    \Statex \hskip 8.5em $t=0, \text{update=True})$
                    \State \textit{*Discard predicted video latent from $\mathbf{x}_1$}
                \EndWhile
            \end{algorithmic}
        \end{algorithm}
    \end{minipage}
\end{figure*}
\label{algo:main}

\section{Real-time Execution Details}
\label{appen:realtime}
This appendix provides additional details on the system-, implementation-, and model-level optimizations introduced in Section~\ref{subsec:realtime}.

\subsection{System-level Optimizations}
\textbf{CFG Parallelism.} To address the computation bottleneck inherent to DiT, we employed Classifier-Free Guidance (CFG) parallelism. Standard CFG requires two distinct model evaluations—the conditional and unconditional forward passes—which are typically executed sequentially. We parallelized these operations by distributing the conditioned and null-conditioned score estimations across two independent GPUs. This effectively reduced the latency per diffusion step by nearly half, with no impact on overall model quality.

\textbf{DiT Caching.} The iterative nature of DiT inference imposes a significant computational bottleneck for real-time applications. Previous efforts have focused on training-free caching techniques that exploit temporal redundancies across diffusion steps. TeaCache \citep{liu2024timestep} leverages the relative $L_1$ difference between timestep-embedding modulated inputs as a heuristic to estimate output variance and skip redundant computations, while TaylorSeer \citep{TaylorSeer2025} employs higher-order Taylor expansions to extrapolate future latent states from historical derivatives. In \ourmethod, we implement a caching mechanism that exploits the directional consistency of velocity vectors learned during flow matching.

During inference, the model tracks cosine similarity between successive velocity predictions. When this metric exceeds a predefined threshold ($\tau$), the model bypasses the DiT forward pass for a window of a few steps by reusing the cached velocity vector. This adaptive scheduling concentrates computational resources on critical trajectory updates, reducing the average number of DiT steps from 16 to 4 with minimal degradation to predicted video and action fidelity.

\textbf{Asynchronous Execution.} Sequential execution of action blocks introduces stalls while waiting for upstream models to produce the next chunk, making the robot more open-loop and less reactive to real-time state changes, especially with large chunk sizes. We use an asynchronous execution mechanism that decouples model inference from action execution, allowing both stages to run concurrently. The motion controller always executes the most recent action scheduled for the current timestamp, while the inference module always uses the latest observation.

\subsection{Implementation-level Optimizations}
\textbf{Torch Compile and CUDA Graphs.} Inference is predominantly CPU-bound due to kernel launch overheads and Python execution. We employ torch.compile with CUDA Graphs (\texttt{mode="reduce-overhead"}) and enforce full graph capture (\texttt{fullgraph=True}) to eliminate graph breaks. In addition to reducing CPU overhead, \texttt{torch.compile} decreases memory bandwidth requirements through operator fusion. We apply compilation to five model components: the diffusion transformer, scheduler, text encoder, image encoder, and VAE.
We enforce static shapes (\texttt{dynamic=False}), which results in multiple recompilations during the first inference trajectory due to the evolving KV cache shape. From the second trajectory onward, inference proceeds without recompilation. To enable error-free compilation, we refactor the model to follow a functional programming paradigm: the KV cache is passed explicitly as input and returned as output of the compiled function.

\textbf{Post-Training Quantization.} We implement a mixed-precision strategy using the NVIDIA Model Optimizer \citep{nvidia2024modelopt} on Blackwell (SM100) architecture. We quantize model weights and activations to NVFP4 (E2M1) while maintaining sensitive QKV projections and Softmax operations in FP8 (E4M3). To preserve numerical stability, we employ FP16 accumulation for non-linear operations including LayerNorm and RoPE. This configuration improves latency with negligible impact on generated video and action quality.

\textbf{Kernel-Level Enhancements.} We use the cuDNN backend for dot-product attention via PyTorch Scaled Dot-Product Attention (SDPA), requiring PyTorch version $\ge$2.9. Earlier versions of the Transformer Engine library may also access these efficient cuDNN kernels.

\textbf{Scheduler Optimizations.} The initial Flow UniPC scheduler \citep{zhao2023unipc} implementation required CPU execution for several operations, causing frequent CPU–GPU synchronization and GPU stalls. We migrated these operations to GPU, eliminating unnecessary CPU overhead.

\subsection{Model-level Optimizations — \ourmethod-Flash}

In the standard \ourmethod formulation, video and action modalities share the same denoising timestep $t_k$:
\begin{equation}
t_k^{\text{video}} = t_k^{\text{action}} = t_k, \quad t_k \sim \mathcal{U}(0, 1)
\end{equation}

\ourmethod-Flash decouples these schedules by biasing video timesteps toward lower values (higher noise) while keeping action timesteps uniform:
\begin{equation}
t_k^{\text{video}} = 1 - \eta, \quad \eta \sim \text{Beta}(\alpha, \beta), \quad t_k^{\text{action}} \sim \mathcal{U}(0, 1)
\end{equation}
where $\alpha > \beta$ (e.g., $\alpha=7, \beta=1$). Since $\text{Beta}(\alpha, \beta)$ with $\alpha > \beta$ concentrates mass near $\eta \approx 1$, the transformed variable $t_k^{\text{video}} = 1 - \eta$ is biased toward 0, corresponding to high-noise video states. The noisy samples become:
\begin{equation}
\mathbf{z}_{t_k^{\text{video}}}^{k} = t_k^{\text{video}} \mathbf{z}_{1}^{k} + (1 - t_k^{\text{video}}) \mathbf{z}_{0}^{k}, \quad
\mathbf{a}_{t_k^{\text{action}}}^{k} = t_k^{\text{action}} \mathbf{a}_{1}^{k} + (1 - t_k^{\text{action}}) \mathbf{a}_{0}^{k}
\end{equation}

For $\text{Beta}(7,1)$, $\mathbb{E}[\eta] = 0.875$, yielding $\mathbb{E}[t_k^{\text{video}}] = 0.125$ compared to $0.5$ in the coupled setting. This exposes the model during training to configurations where actions must be predicted from predominantly noisy visual context (Figure~\ref{fig:dreamzeroflash}), aligning training with rapid-action-denoising inference where actions denoise from noise level $1 \rightarrow 0$ in one step while video remains partially noisy.

\textbf{Action Chunk Smoothing.} Generated action chunks may contain high-frequency noise from denoising. We apply filtering to ensure stable real-world behavior: first upsampling the action chunk to $2\times$ resolution via cubic interpolation, then applying a Savitzky-Golay filter (window size 21, polynomial order 3) to suppress noise while preserving trajectory shape, and finally downsampling to original resolution.

\section{AgiBot Diverse Data Collection Strategy}
\label{appen:agibot_dc}

\begin{figure}[ht!]
    \centering
    \includegraphics[width=1\textwidth]{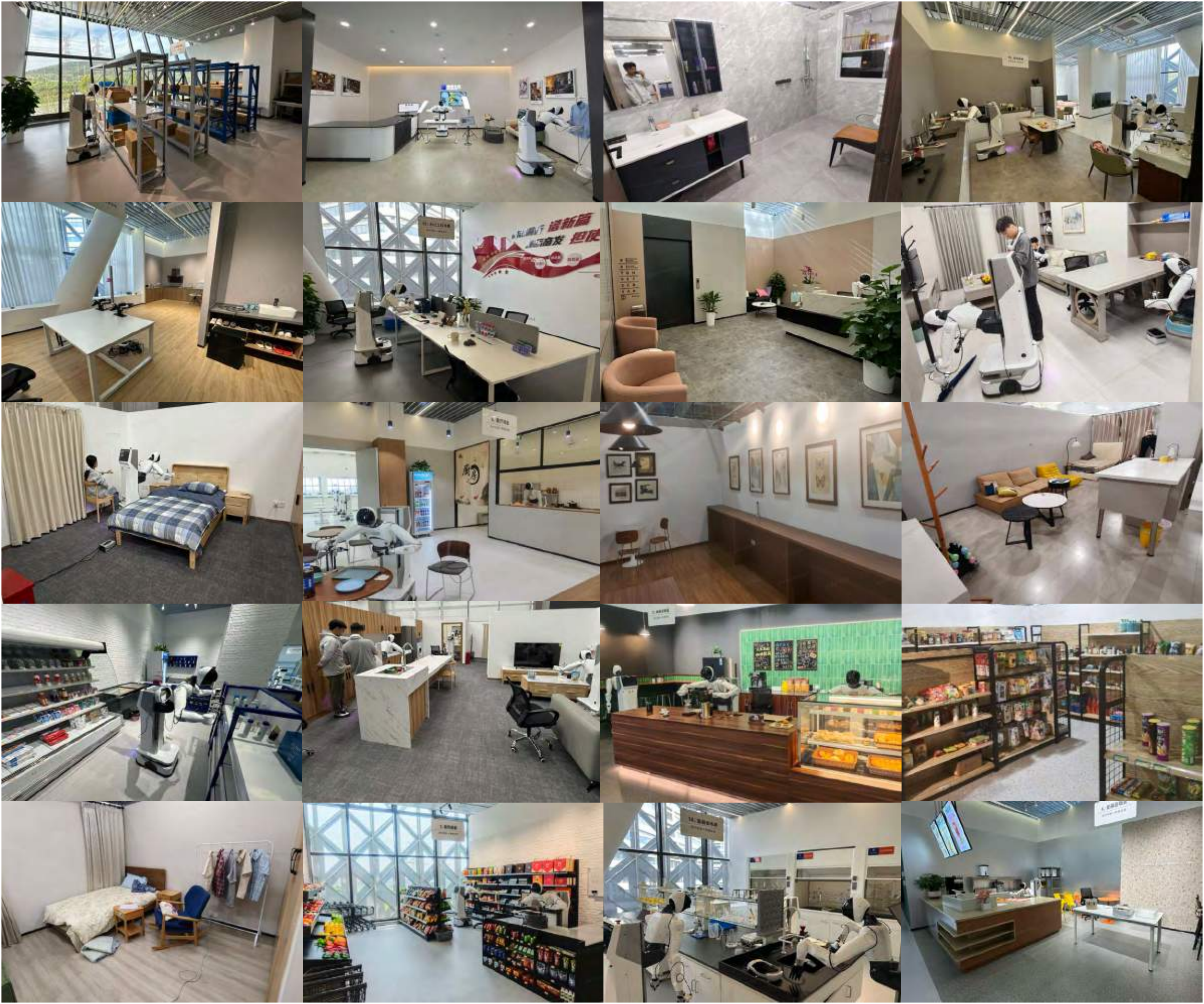}
    \caption{\textbf{Data Collection Environments.} We collect teleoperation data across 22 diverse real-world environments, including offices, laboratories, restaurants, supermarkets, coffee shops, warehouses, homes, hotels, and retail stores. This diversity enables \ourmethod to generalize to unseen environments without task-specific fine-tuning.}
    \label{fig:data_collection_envs}
\end{figure}

Our data collection philosophy prioritizes \textit{diversity over repetition}. Unlike conventional approaches that collect hundreds of demonstrations per task in controlled lab settings, we collect data across 22 real-world environments spanning homes, restaurants, supermarkets, coffee shops, offices, warehouses, laboratories, and hotels (Figure~\ref{fig:data_collection_envs}).

\subsection{Daily Collection Workflow}

Each day, teleoperators receive a printed task sheet listing available tasks for their assigned area (e.g., kitchen area, checkout counter). For each episode, they select three \textit{tasks} (usually very coarse-grained like ``tidy up") from this sheet and execute them consecutively. Each task typically requires 1–2 minutes, resulting in approximately 5-minute episodes.

At the end of each day, teleoperators log the frequency count for each task. Once a task was collected in 50 episodes, it is deprecated and removed from the task sheet. Teleoperators are incentivized to propose new tasks, which they inevitably must do as existing tasks become deprecated. This mechanism continuously expands the task distribution throughout data collection, yielding a long-tail of diverse behaviors.

Because we prioritize utility over repetition, our tasks are naturally more coarse-grained than typical robot learning datasets. Examples include organizing items, ground garbage cleaning, shopping basket return, toy box tidying, table tidying, and clothes hanging—but the full set grows organically as the collection progresses.

\subsection{Multi-Task Episode Structure}

The three-task episode structure serves two purposes: it maximizes diversity within each episode and encourages the model to learn smooth task transitions. For example, a single episode might involve (1) clearing dishes from a table, (2) wiping the table surface, and (3) organizing condiments. This design yields an average of 42 subtasks per episode (Figure~\ref{fig:data_stats}), significantly more than typical single-task datasets.

This combination of environment diversity, task deprecation with forced expansion, and multi-task episodes produces a heterogeneous dataset that differs substantially from conventional robot learning corpora. Rather than learning narrow task-specific policies, \ourmethod learns generalizable skills that transfer across environments and tasks. See \url{https://dreamzero0.github.io/training_data_gallery/} for examples of these coarse-grained tasks (the videos shown in the website concatenated each coarse-grained tasks)

\section{AgiBot Evaluation Details}
\label{appen:agibot_evals}


We provide the evaluation setup for both seen tasks in Tables~\ref{tab:seen_tasks} and unseen tasks in  Table~\ref{tab:unseen_tasks} for Agibot. Each row shows the initial frame and instruction for 4 robots. We conduct 2 rollouts per robot for each task by varying the objects, locations and the robot arm to use for the task (The current table mostly shows the evaluation using the left arm).


\begin{table*}[htbp]
\centering
\caption{\textbf{Seen Tasks Evaluation Setup for AgiBot G1}}
\label{tab:seen_tasks}
\setlength{\tabcolsep}{2pt}
\resizebox{\textwidth}{!}{
\begin{tabular}{m{1.8cm}cm{1.8cm}m{3.2cm}m{3.2cm}m{3.2cm}m{3.2cm}m{3.2cm}m{3.2cm}m{3.2cm}m{3.2cm}}
\toprule
\centering\textbf{Category} & \centering\textbf{\#} & \centering\textbf{Task} & \centering\textbf{Image} & \centering\textbf{Instruction} & \centering\textbf{Image} & \centering\textbf{Instruction} & \centering\textbf{Image} & \centering\textbf{Instruction} & \centering\textbf{Image} & \centering\arraybackslash\textbf{Instruction} \\
\midrule
 & 1 & \centering\textbf{PnP Fruit} & \includegraphics[width=3.2cm]{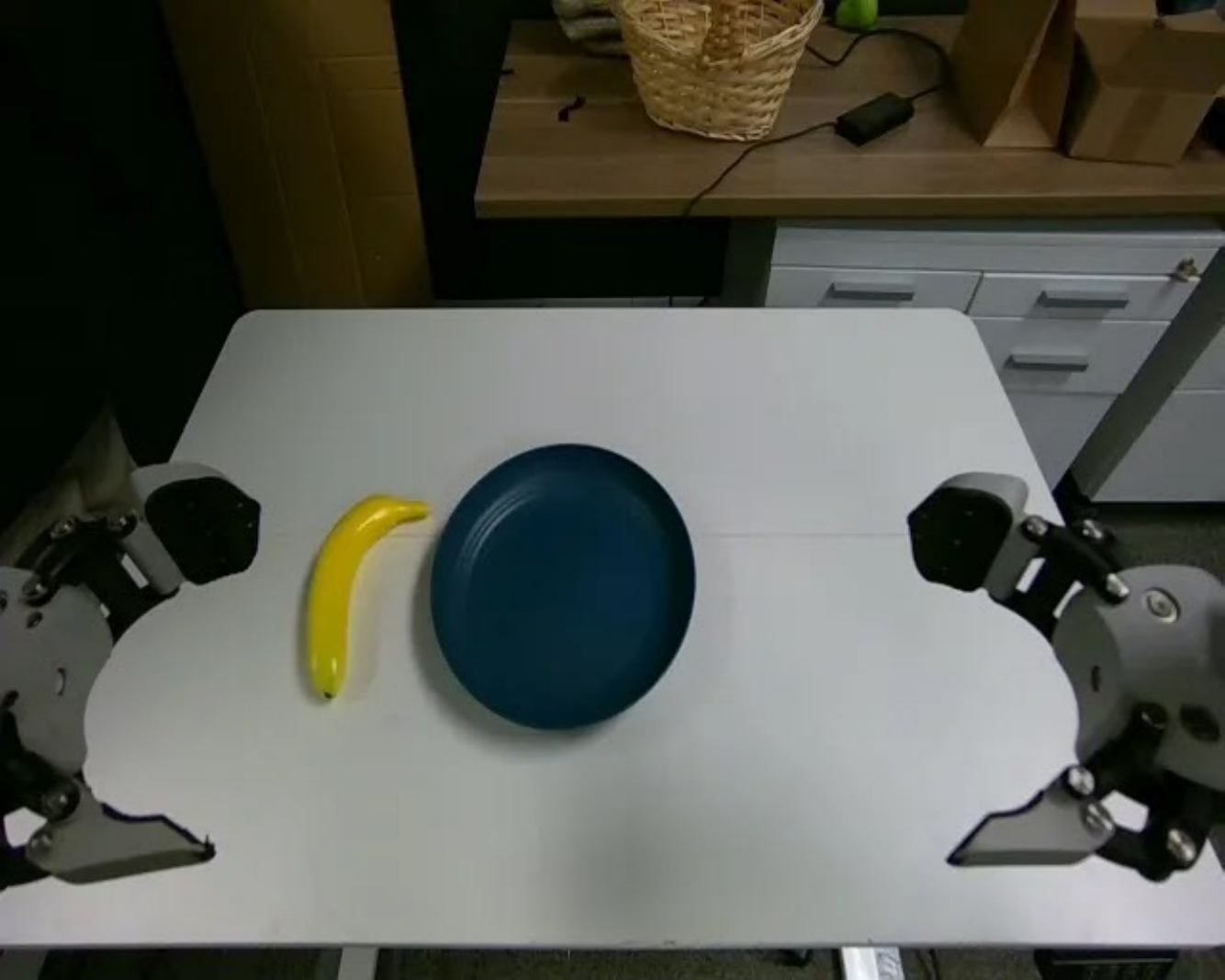} & \centering\scriptsize{The left arm picks up the Banana on the table and places it in the Blue Plate.} & \includegraphics[width=3.2cm]{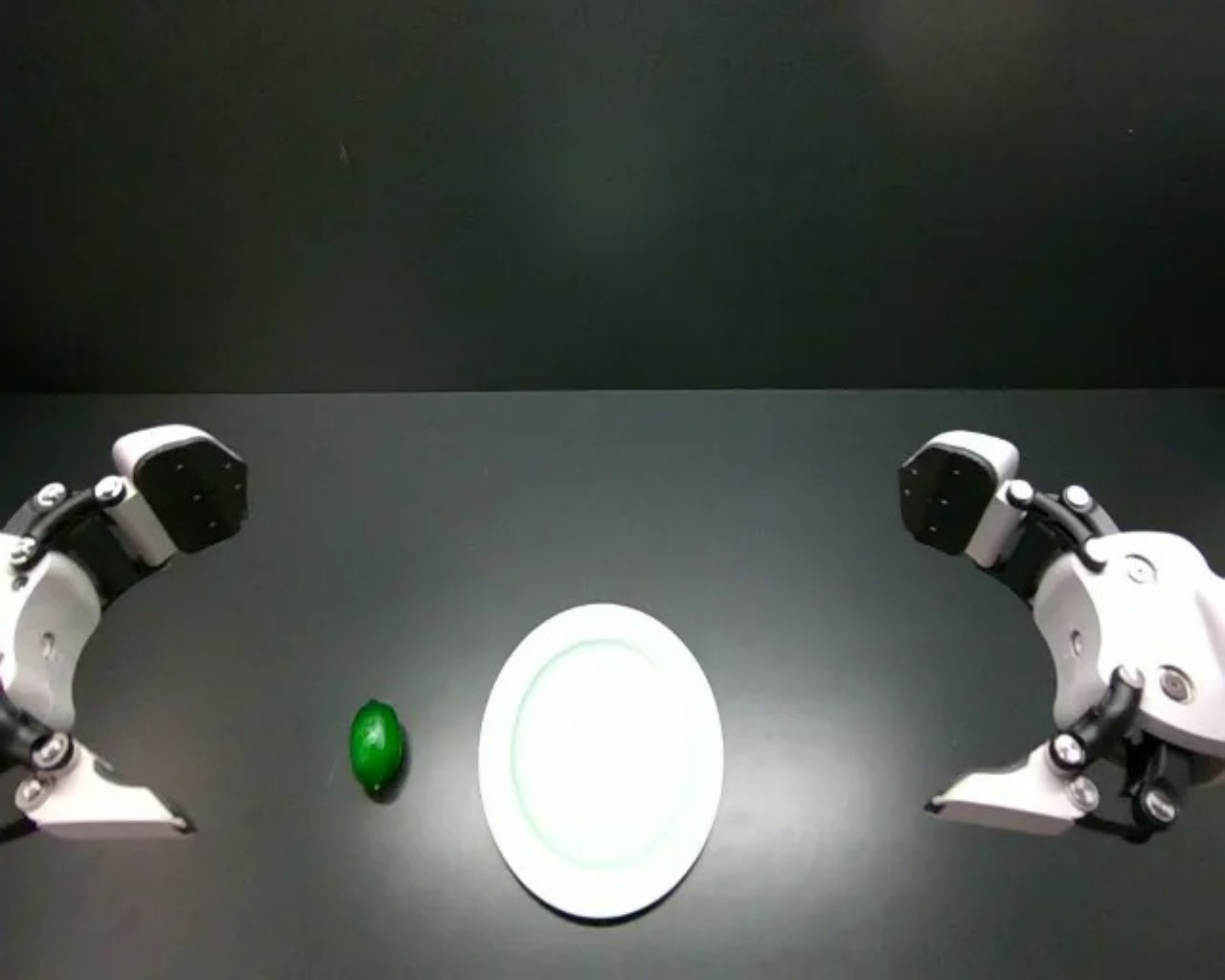} & \centering\scriptsize{The left arm picks up the lime on the table and places it on the light green plate.} & \includegraphics[width=3.2cm]{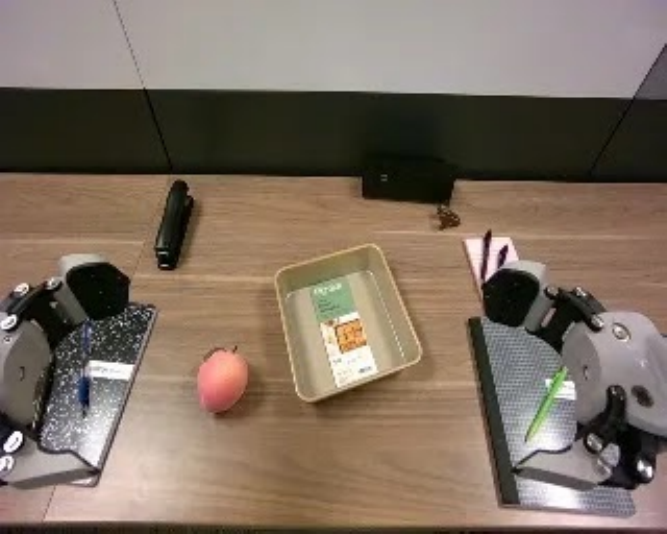} & \centering\scriptsize{The left arm picks up the peach on the table and places it on the baking pan.} & \includegraphics[width=3.2cm]{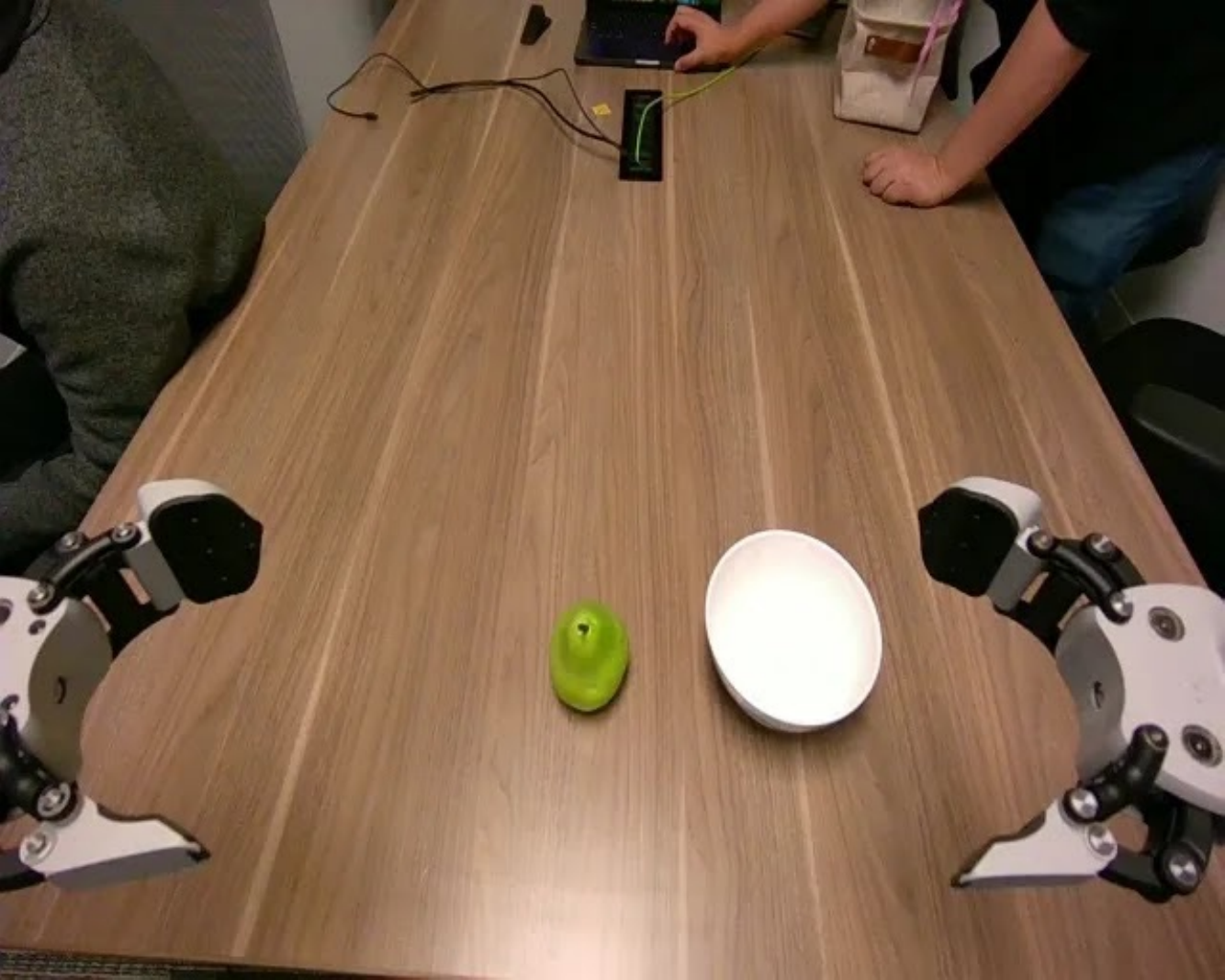} & \centering\arraybackslash\scriptsize{The left arm picks up the green pear on the table and places it on the blue checkered bowl.} \\
\cmidrule(lr){2-11}
\centering\textbf{\shortstack{PnP \\ Easy}} & 2 & \centering\textbf{Taking out Fruit} & \includegraphics[width=3.2cm]{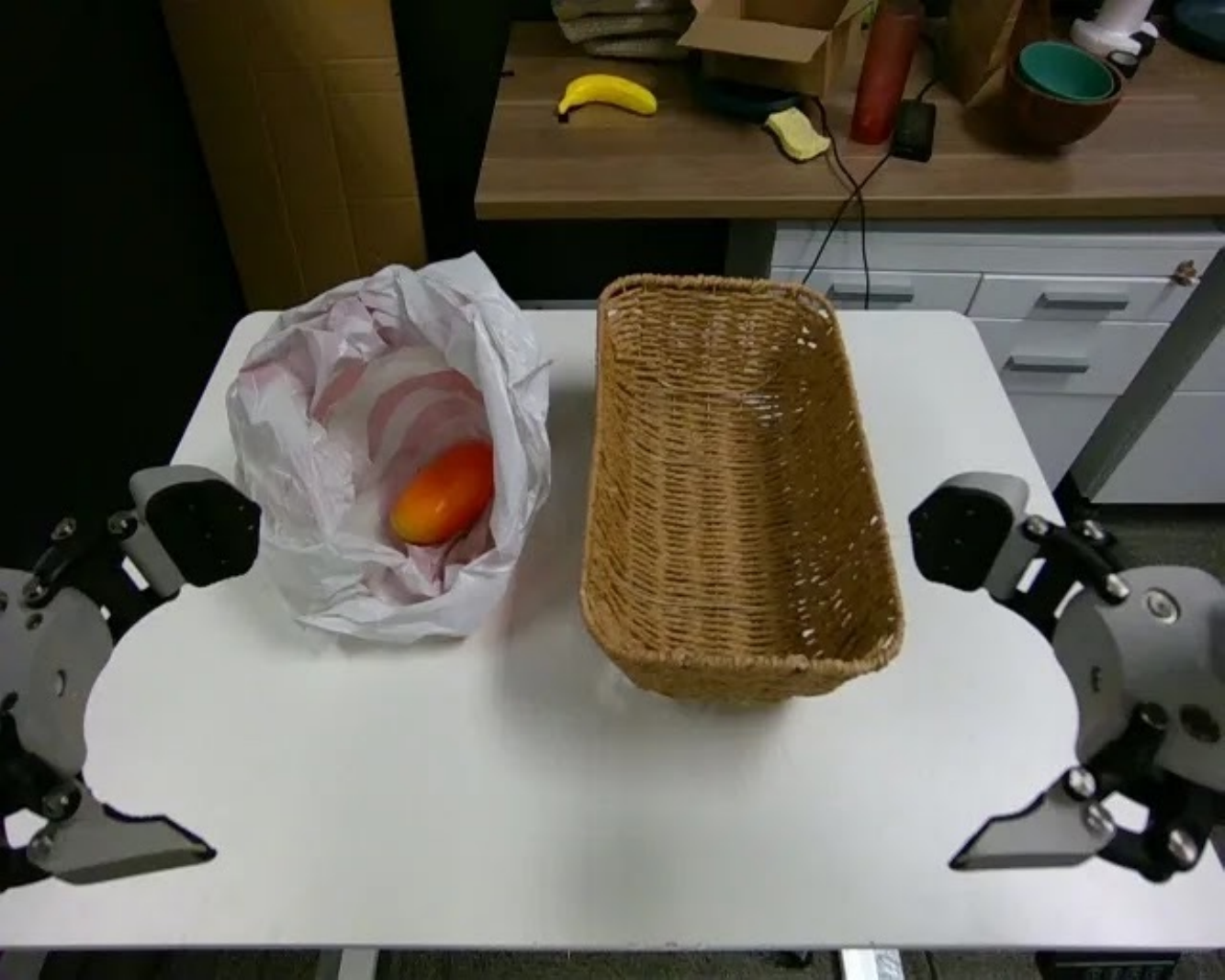} & \centering\scriptsize{The left arm picks up the Mango from the plastic bag and places it into the Wooden Basket} & \includegraphics[width=3.2cm]{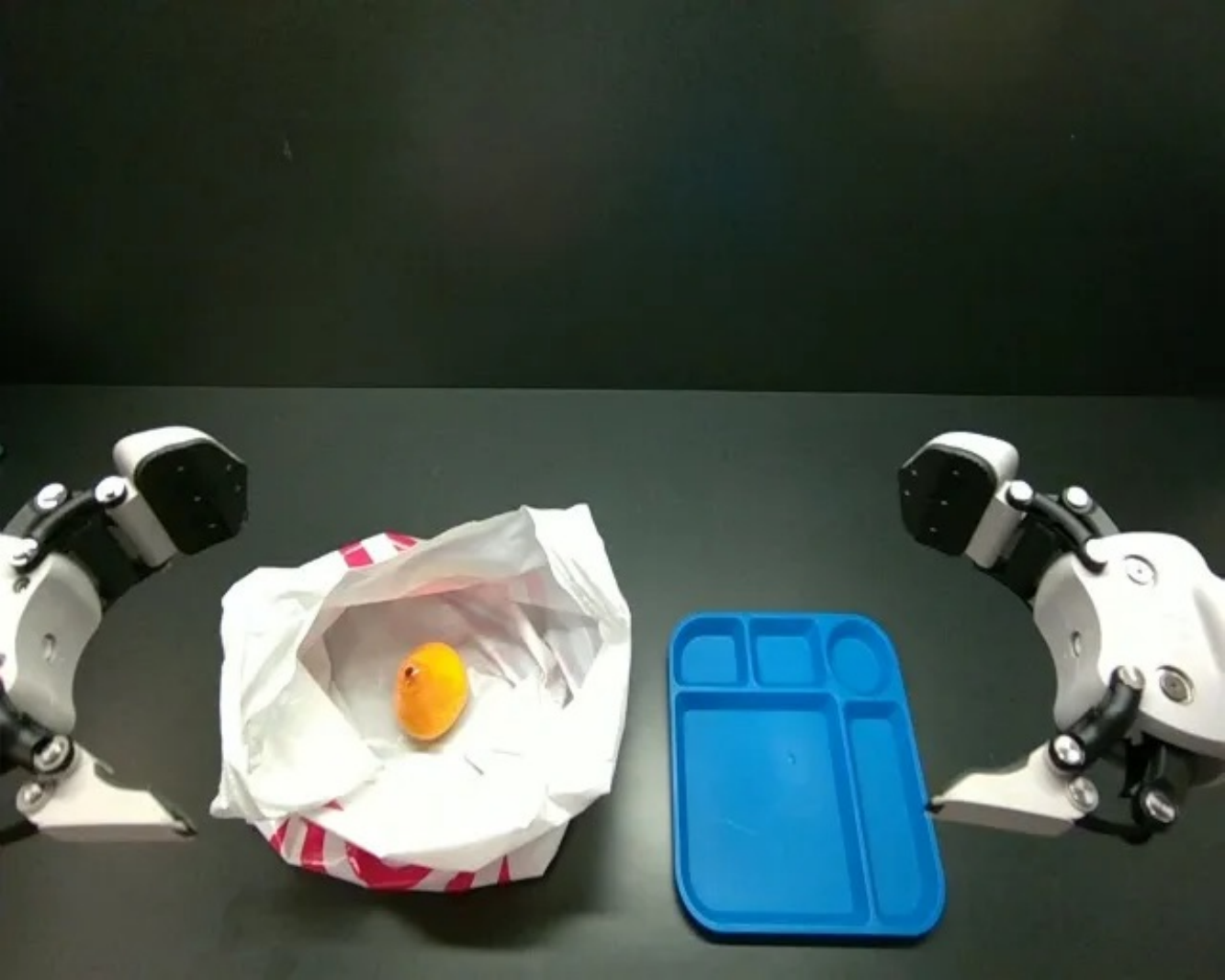} & \centering\scriptsize{The left arm picks up the yellow pear from the plastic bag and places it onto the blue tray.} & \includegraphics[width=3.2cm]{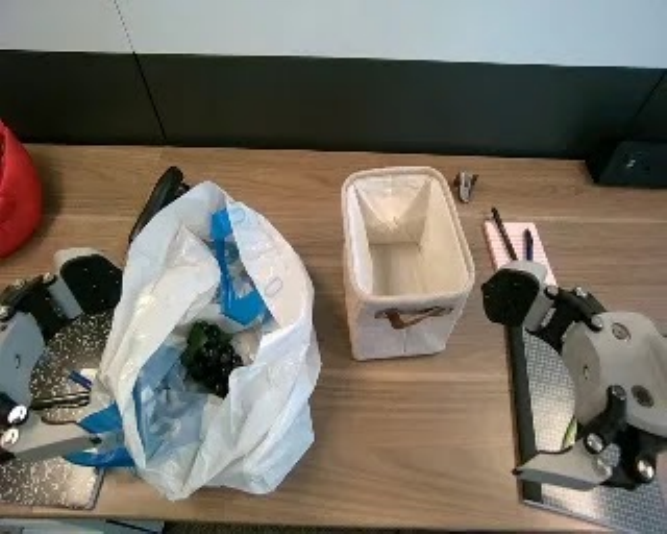} & \centering\scriptsize{The left arm picks up the purple grapes from the plastic bag and places it into the brown basket.} & \includegraphics[width=3.2cm]{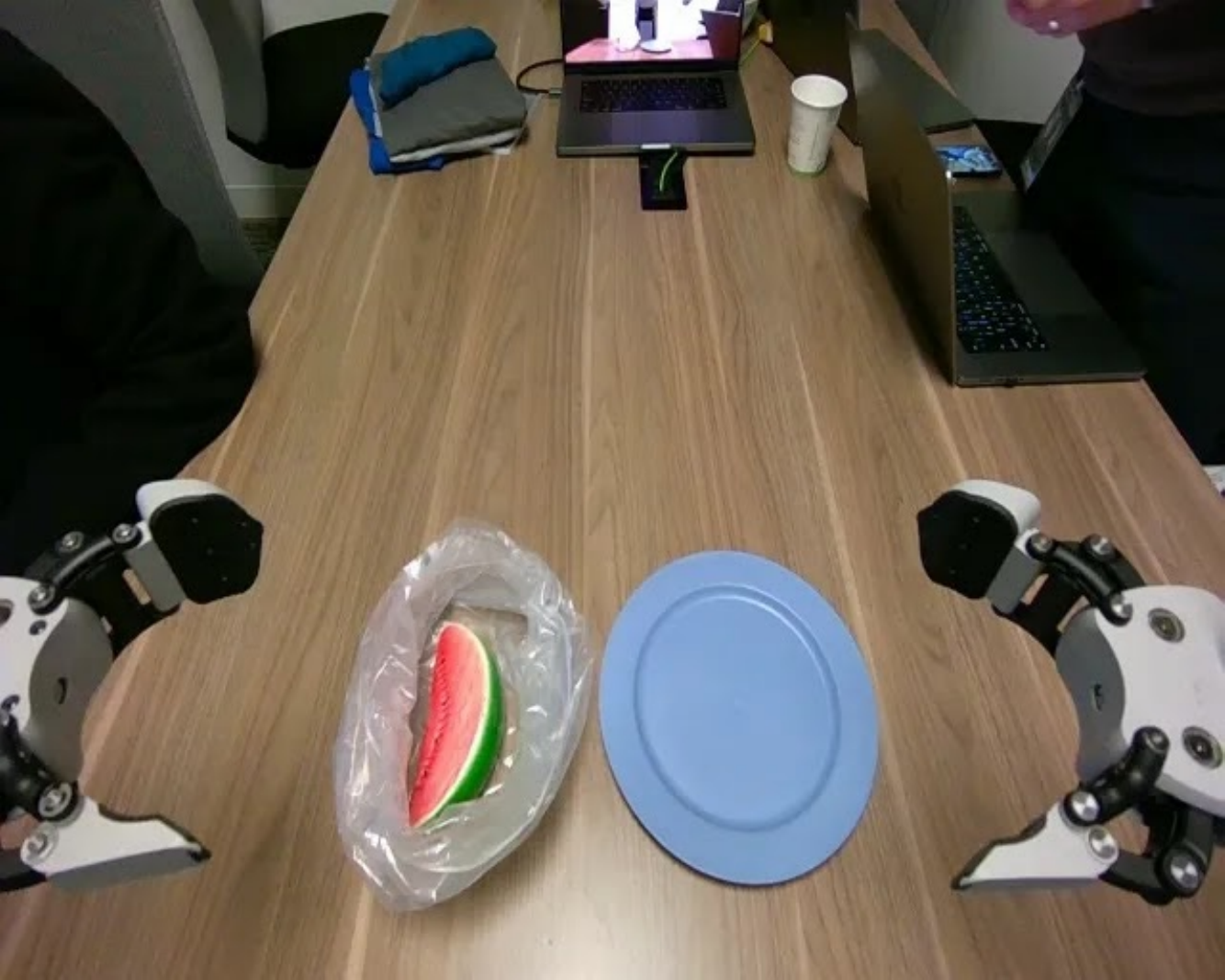} & \centering\arraybackslash\scriptsize{The left arm picks up the watermelon from the plastic bag and places it onto the Blue Plate.} \\
\cmidrule(lr){2-11}
 & 3 & \centering\textbf{Wipe the Mess} & \includegraphics[width=3.2cm]{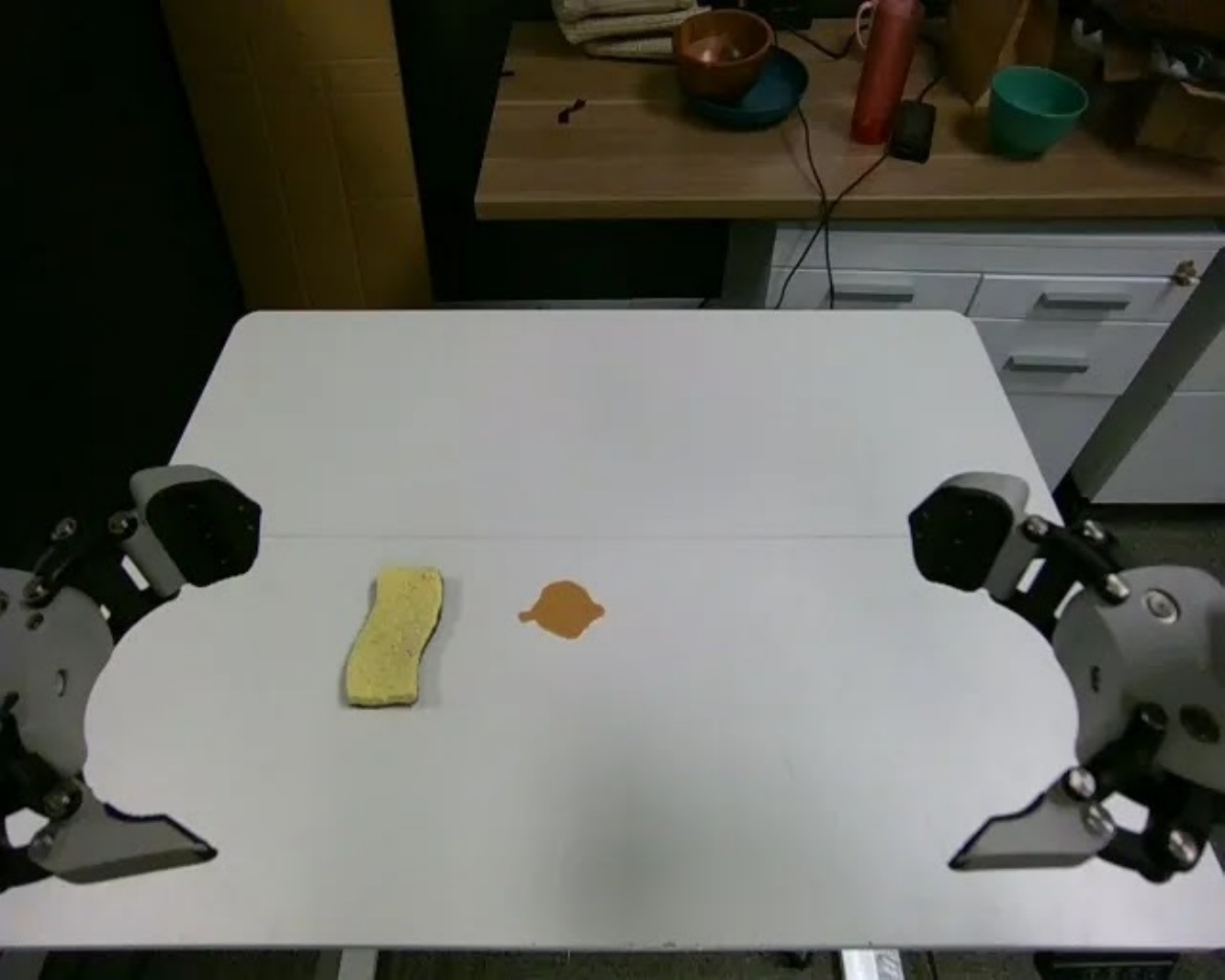} & \centering\scriptsize{The left arm uses a sponge to wipe the coffee spill off the table.} & \includegraphics[width=3.2cm]{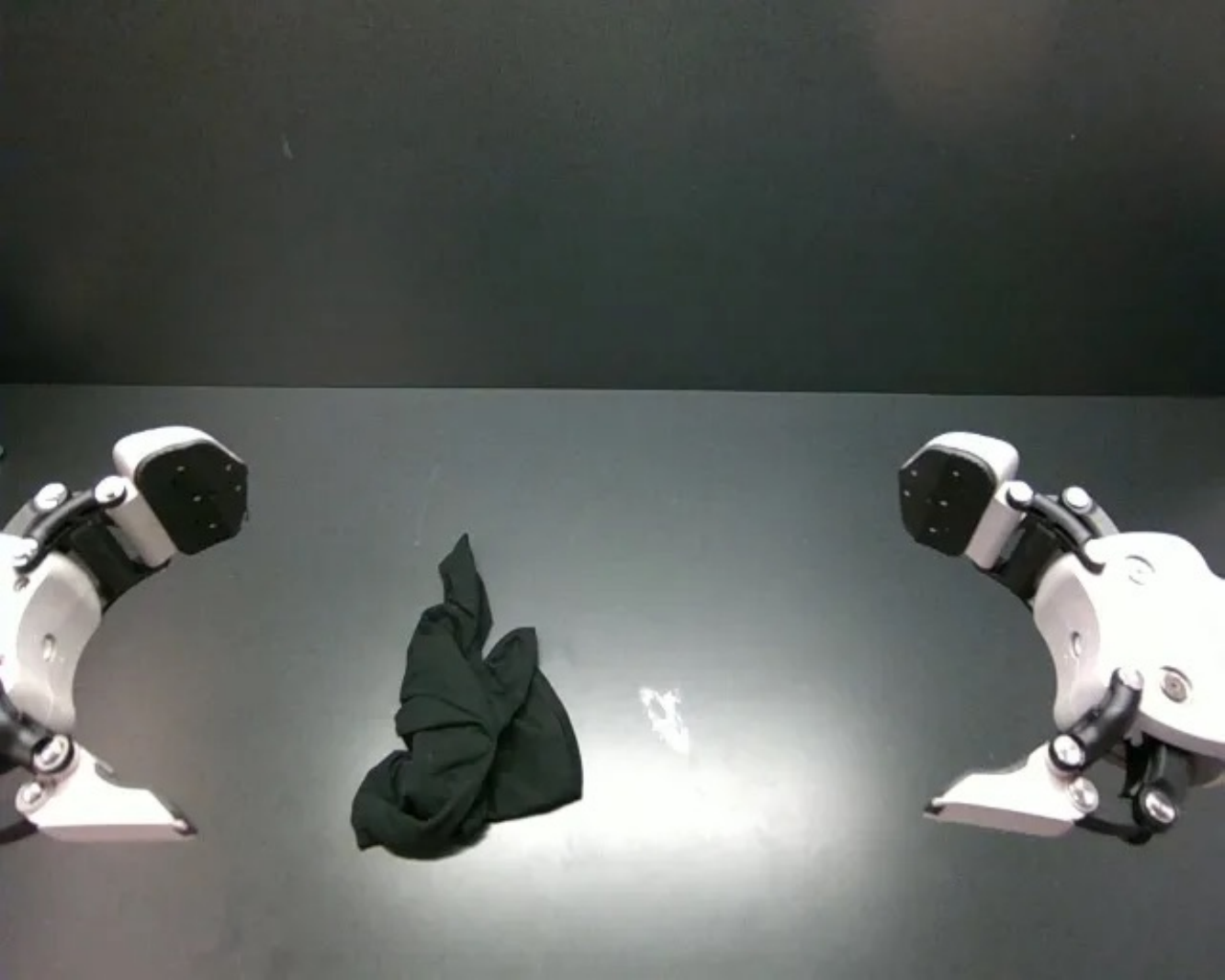} & \centering\scriptsize{The left arm used a black cloth to wipe the white powder off the table.} & \includegraphics[width=3.2cm]{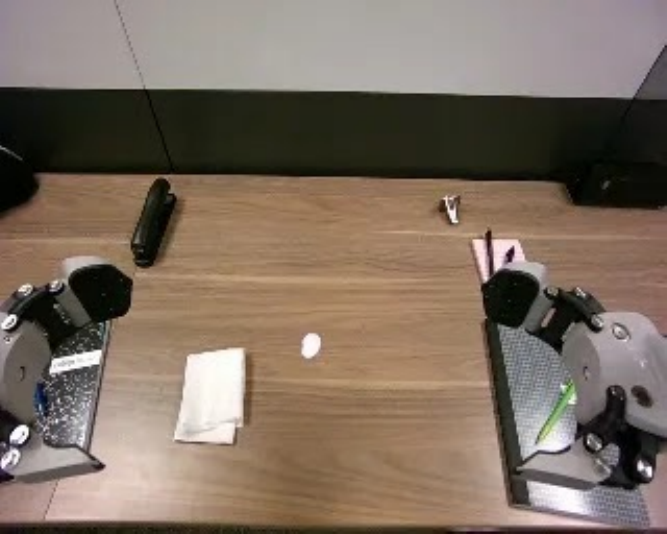} & \centering\scriptsize{The left arm uses a napkin to wipe the creamer spill off the table.} & \includegraphics[width=3.2cm]{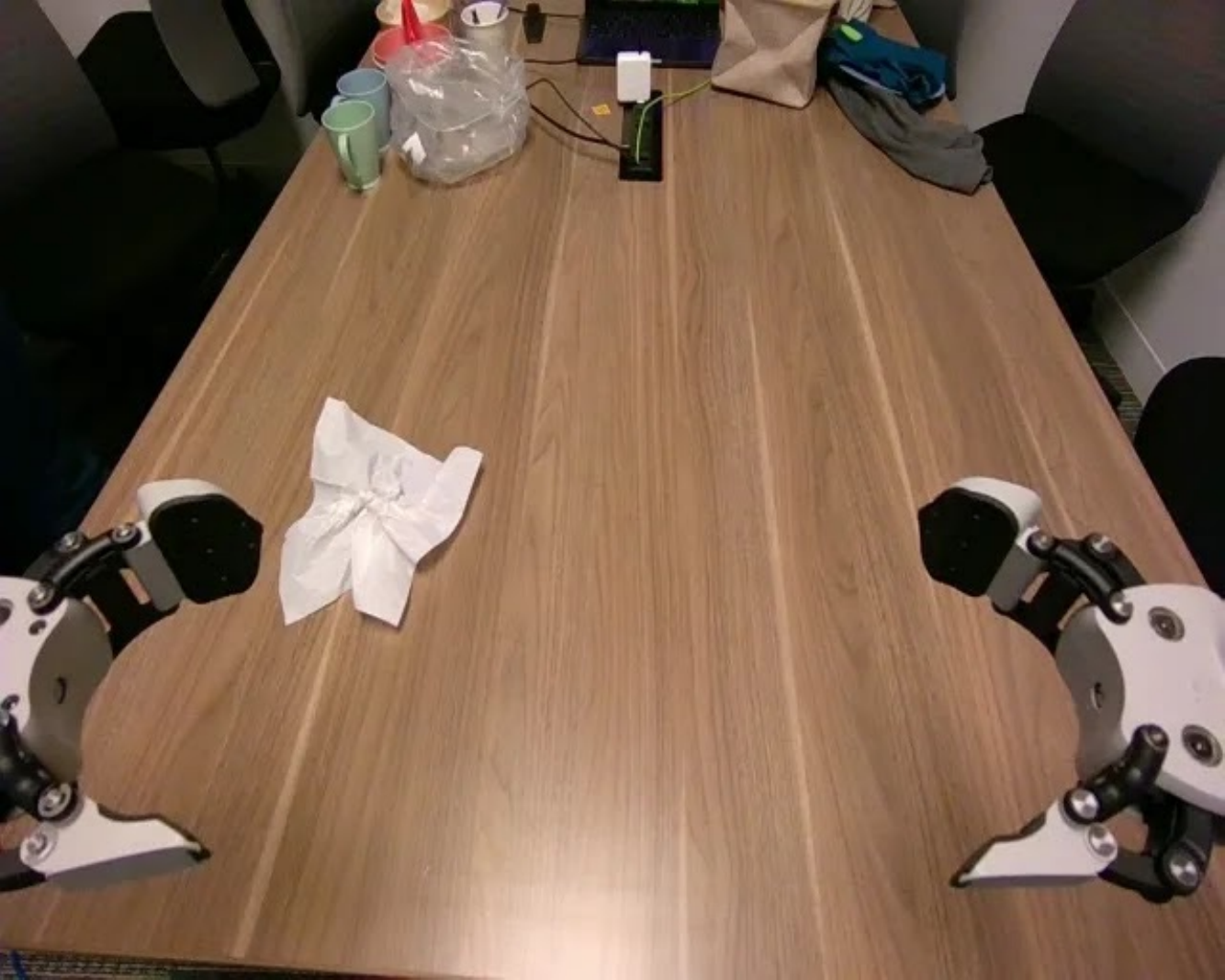} & \centering\arraybackslash\scriptsize{The left arm used a paper towel to wipe the water off the table.} \\
\midrule
 & 4 & \centering\textbf{PnP Fork/Spoon} & \includegraphics[width=3.2cm]{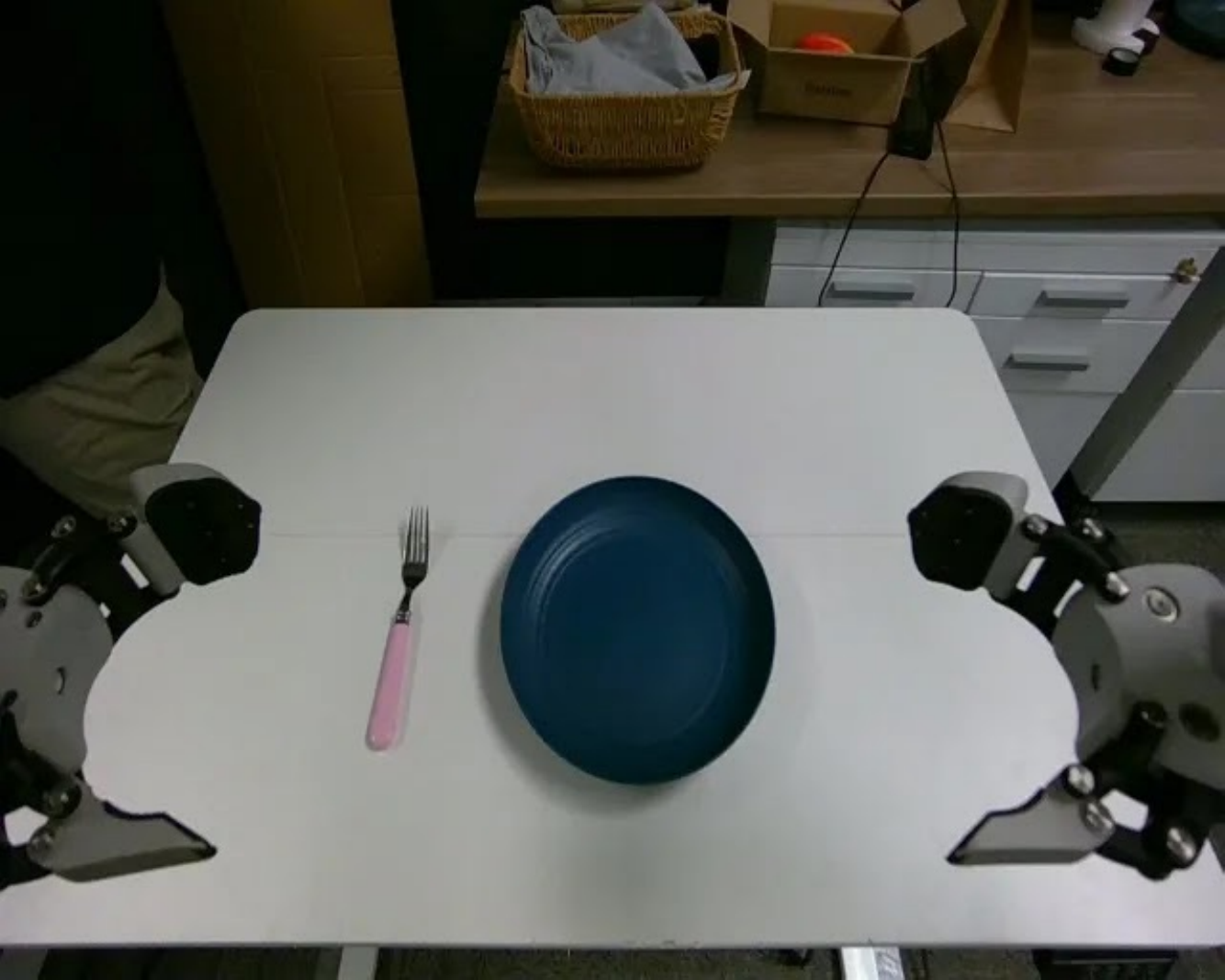} & \centering\scriptsize{The left arm picks up the Pink Fork from the table and places it onto the blue plate.} & \includegraphics[width=3.2cm]{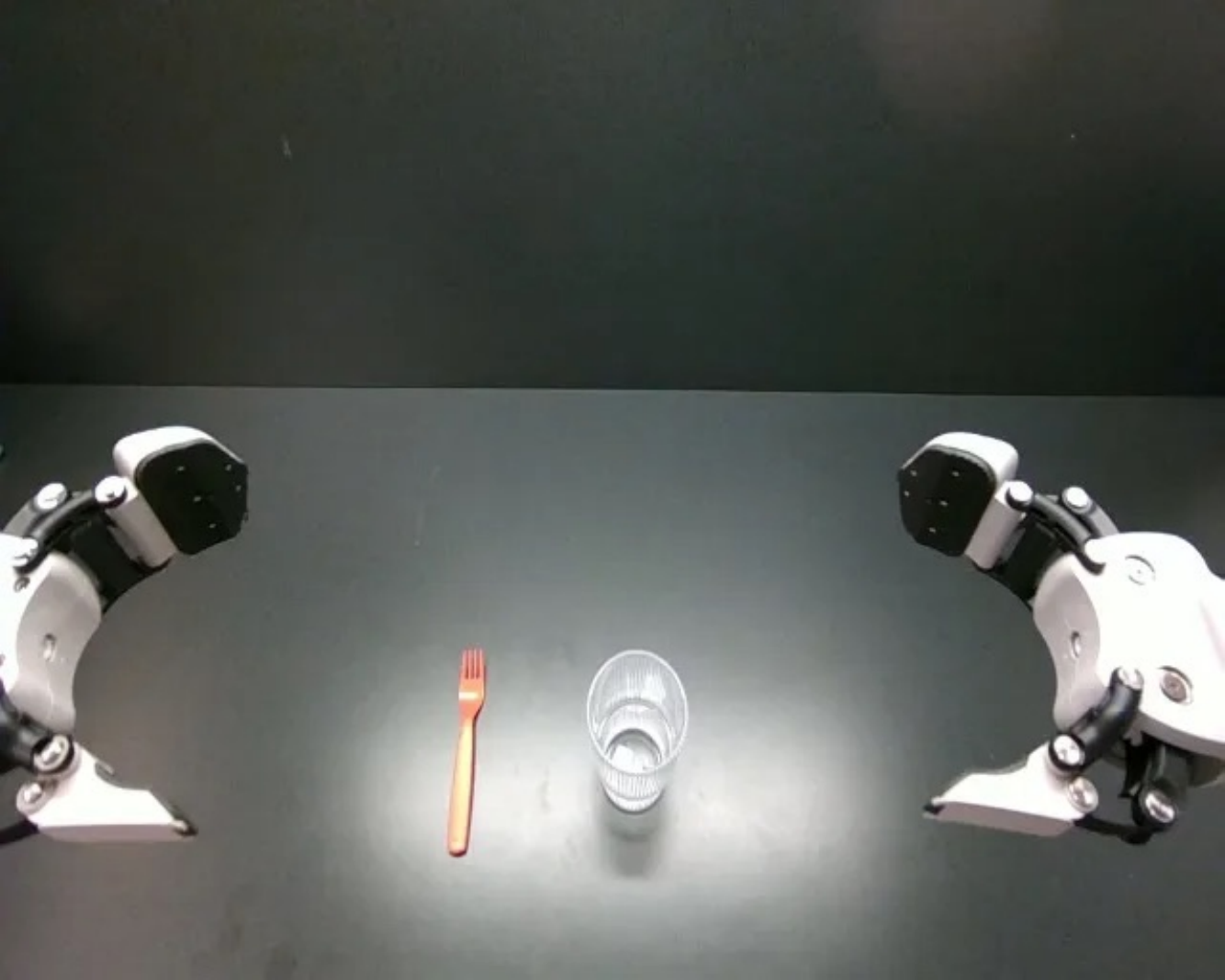} & \centering\scriptsize{The left arm picks up the orange fork from the table and places it into the glass cup} & \includegraphics[width=3.2cm]{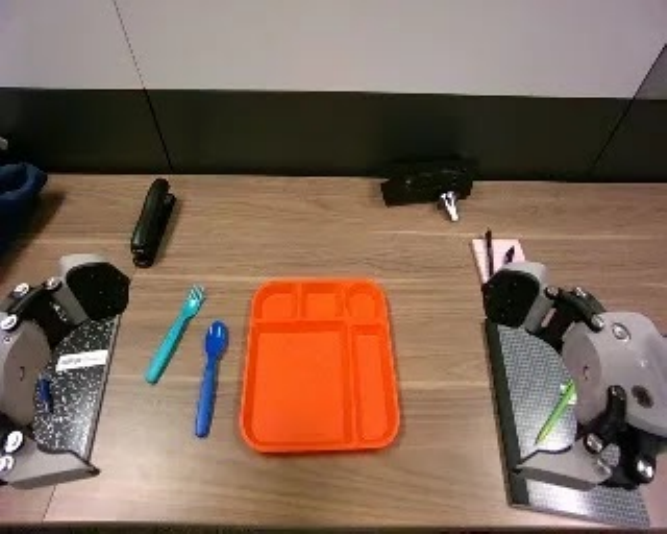} & \centering\scriptsize{The left arm picks up the light blue fork from the table and places it onto the orange plate} & \includegraphics[width=3.2cm]{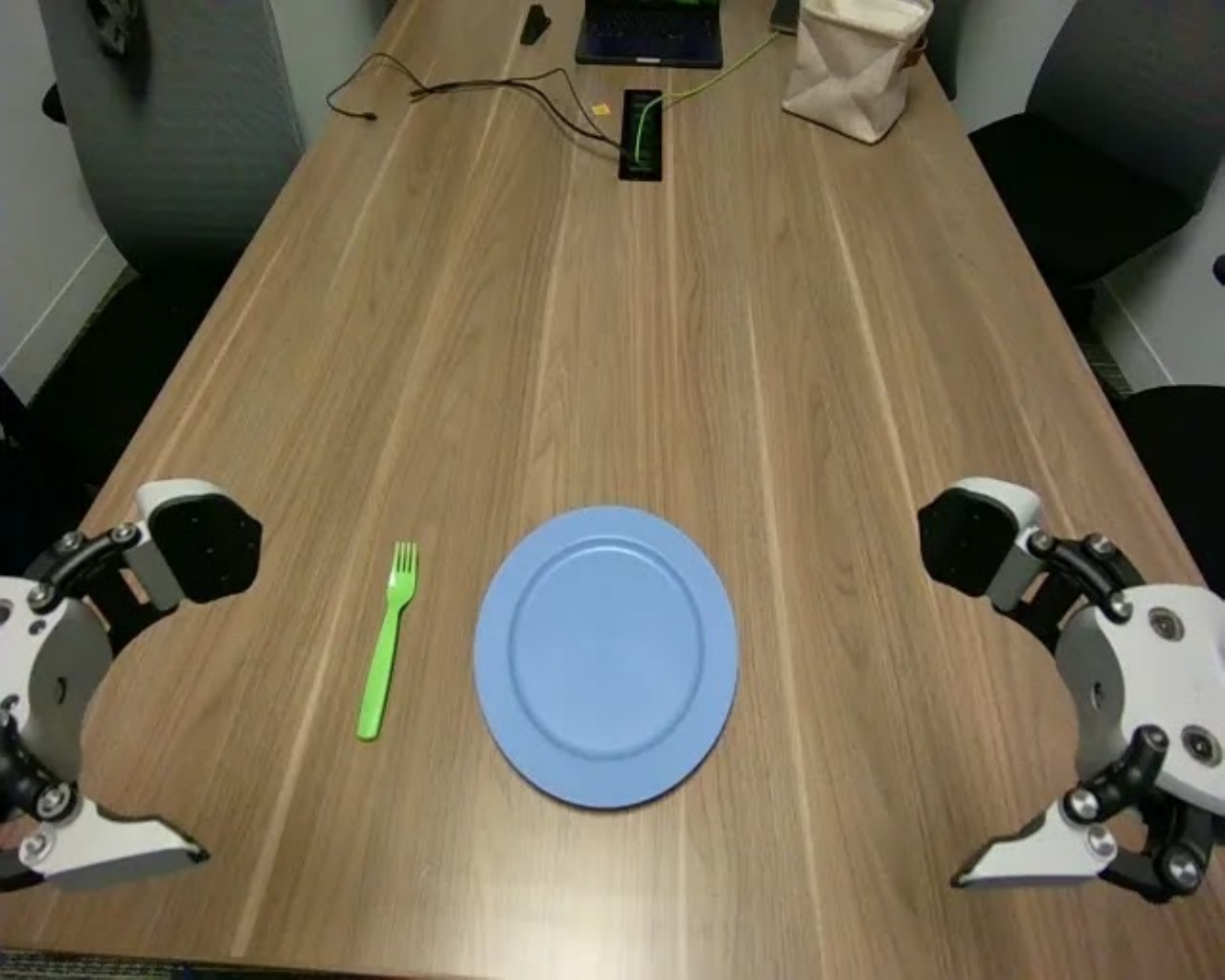} & \centering\arraybackslash\scriptsize{The left arm picks up the green fork from the table and places it into the blue plate.} \\
\cmidrule(lr){2-11}
\multirow{2.3}{*}{\centering\textbf{\shortstack{PnP \\ Hard}}} & 5 & \centering\textbf{Put Pen in Holder} & \includegraphics[width=3.2cm]{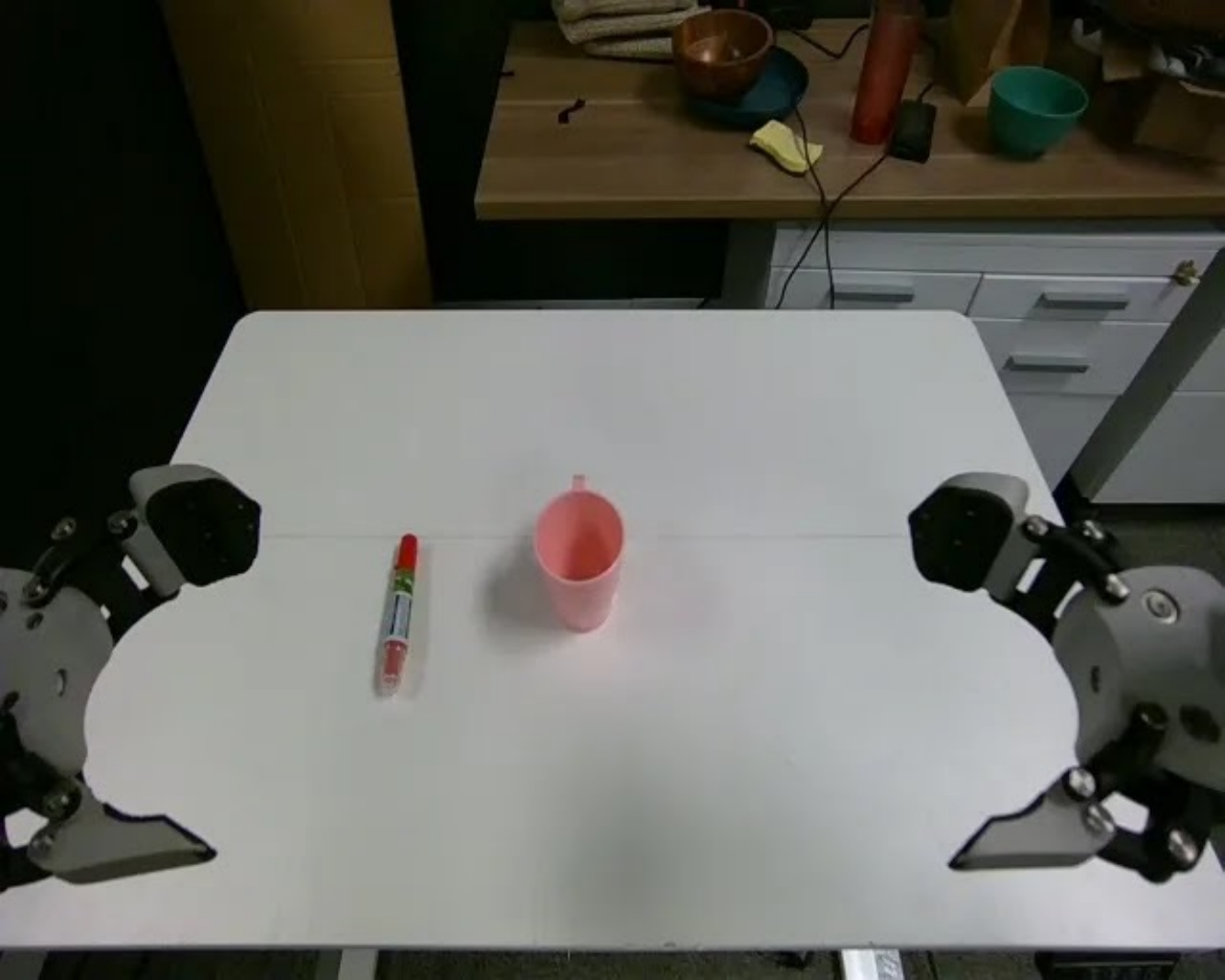} & \centering\scriptsize{The left arm picks up the Red Marker pen from the table and placed it into the pen holder.} & \includegraphics[width=3.2cm]{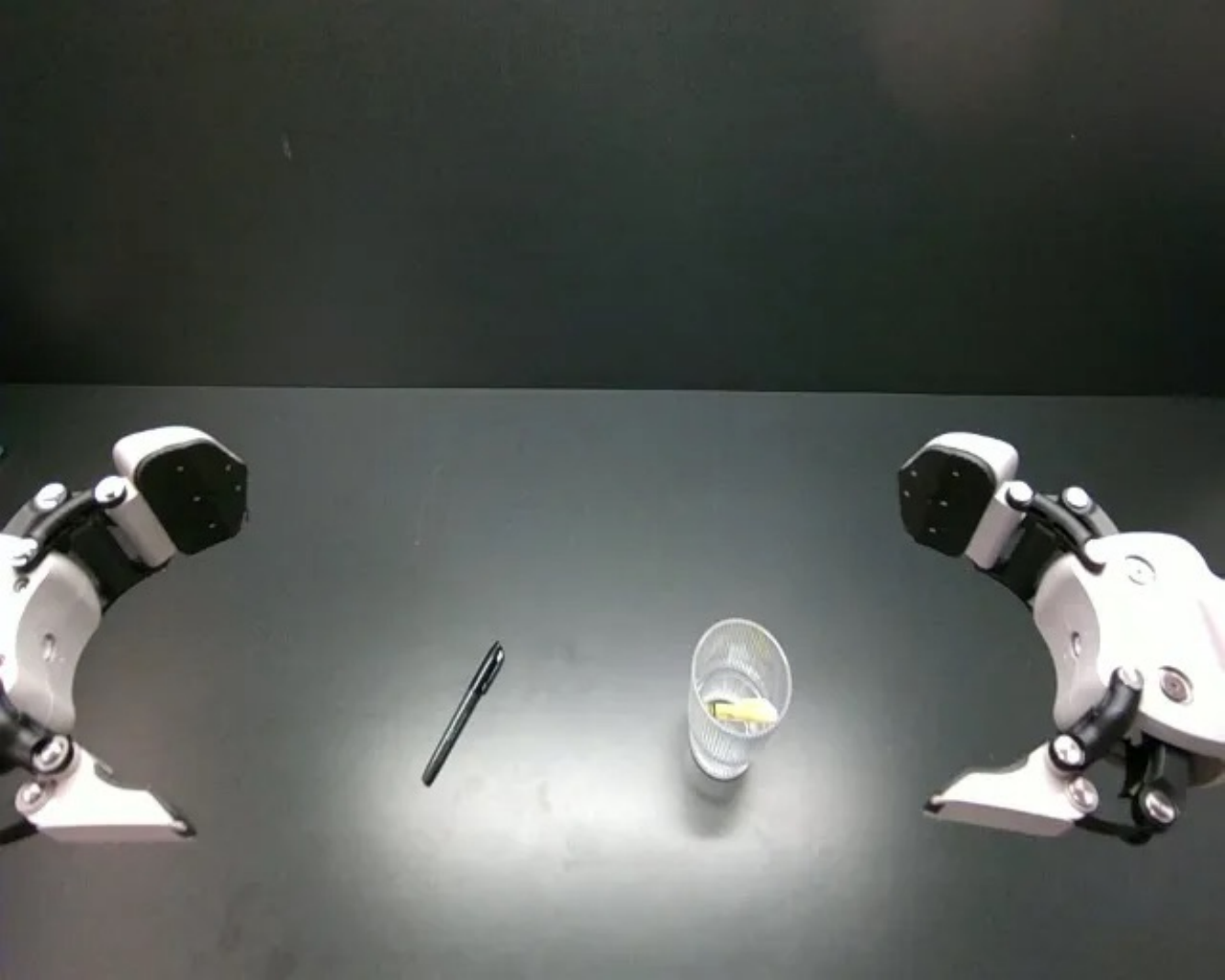} & \centering\scriptsize{The left arm picked up the black marker from the table and placed it into the pen holder.} & \includegraphics[width=3.2cm]{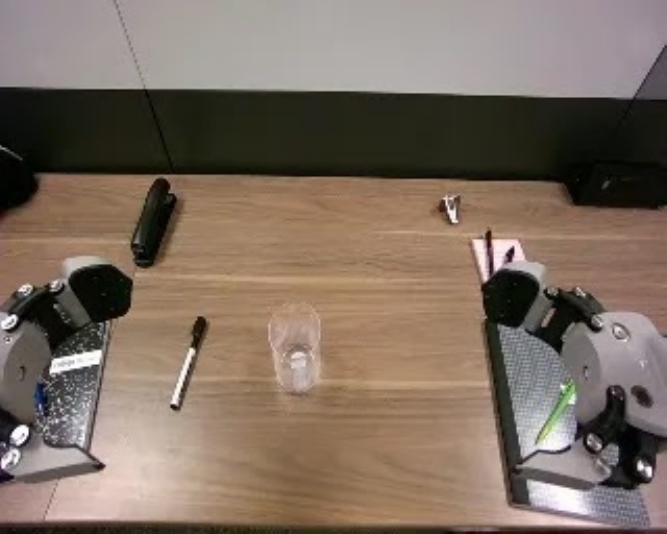} & \centering\scriptsize{The left arm picked up the white marker pen from the table and placed it into the pen holder.} & \includegraphics[width=3.2cm]{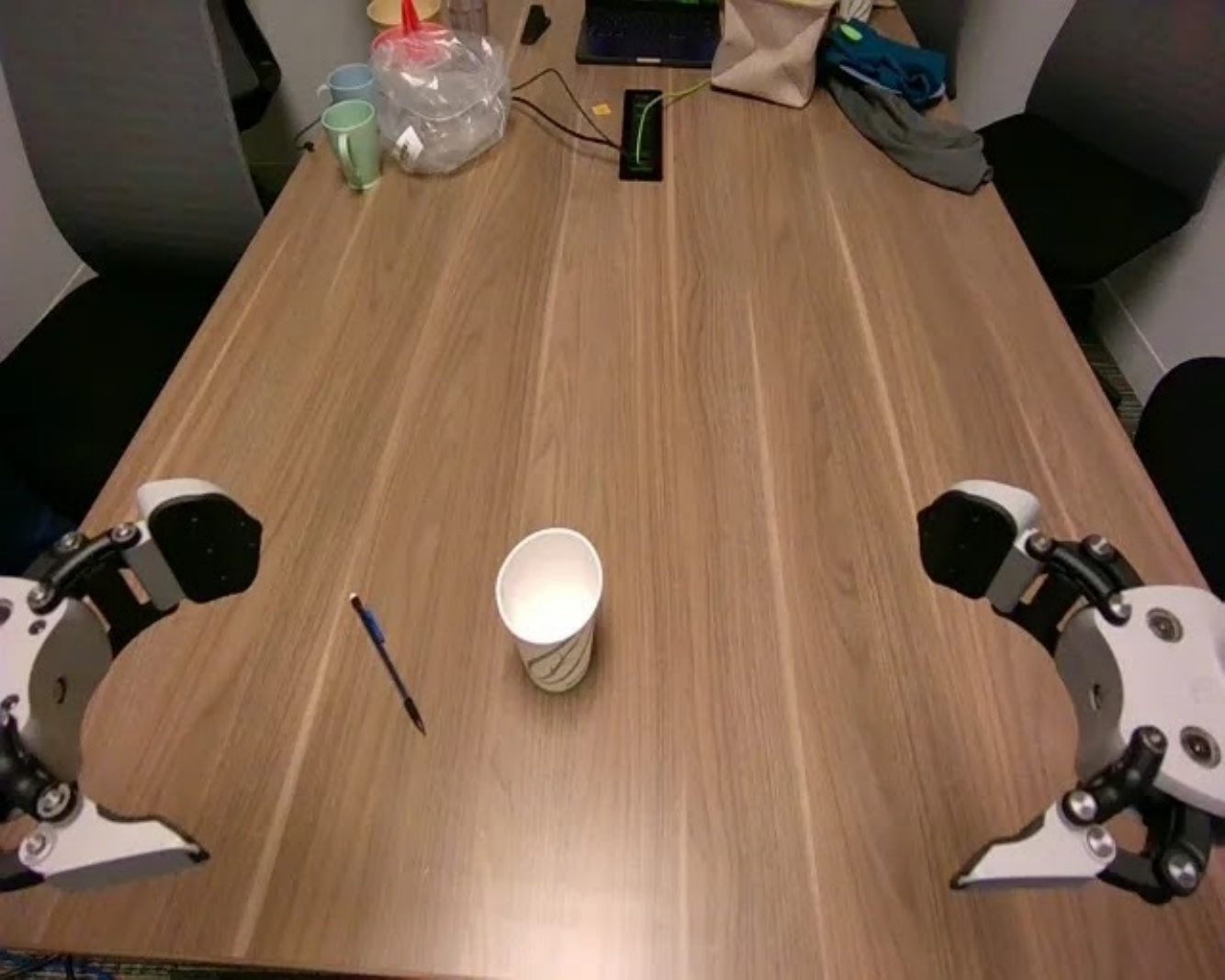} & \centering\arraybackslash\scriptsize{The left arm picked up the mechanical pencil from the table and placed it into the pen holder.} \\
\cmidrule(lr){2-11}
 & 6 & \centering\textbf{Put Cup on Coaster} & \includegraphics[width=3.2cm]{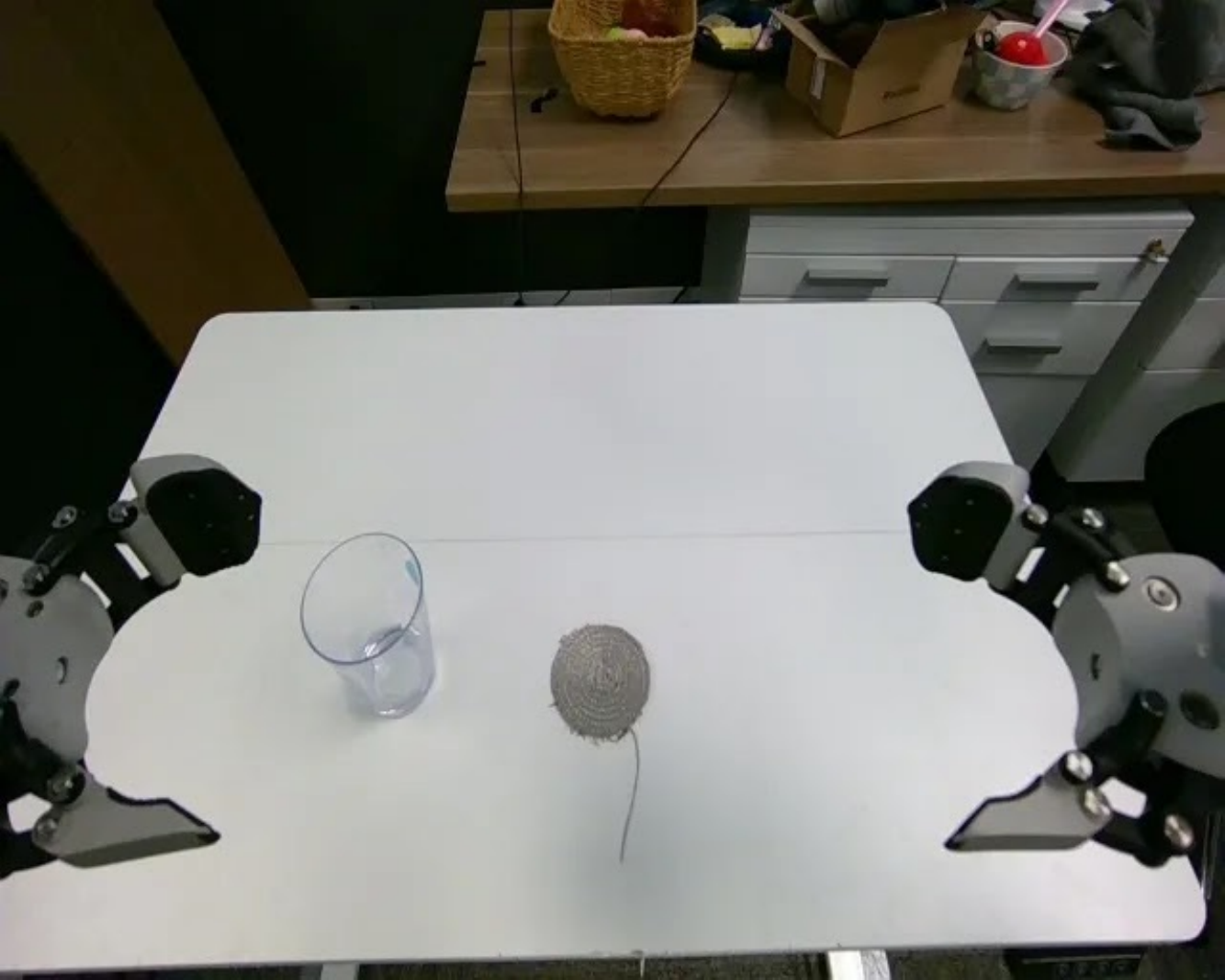} & \centering\scriptsize{The left arm picks up the clear cup from the table and places it on the grey coaster.} & \includegraphics[width=3.2cm]{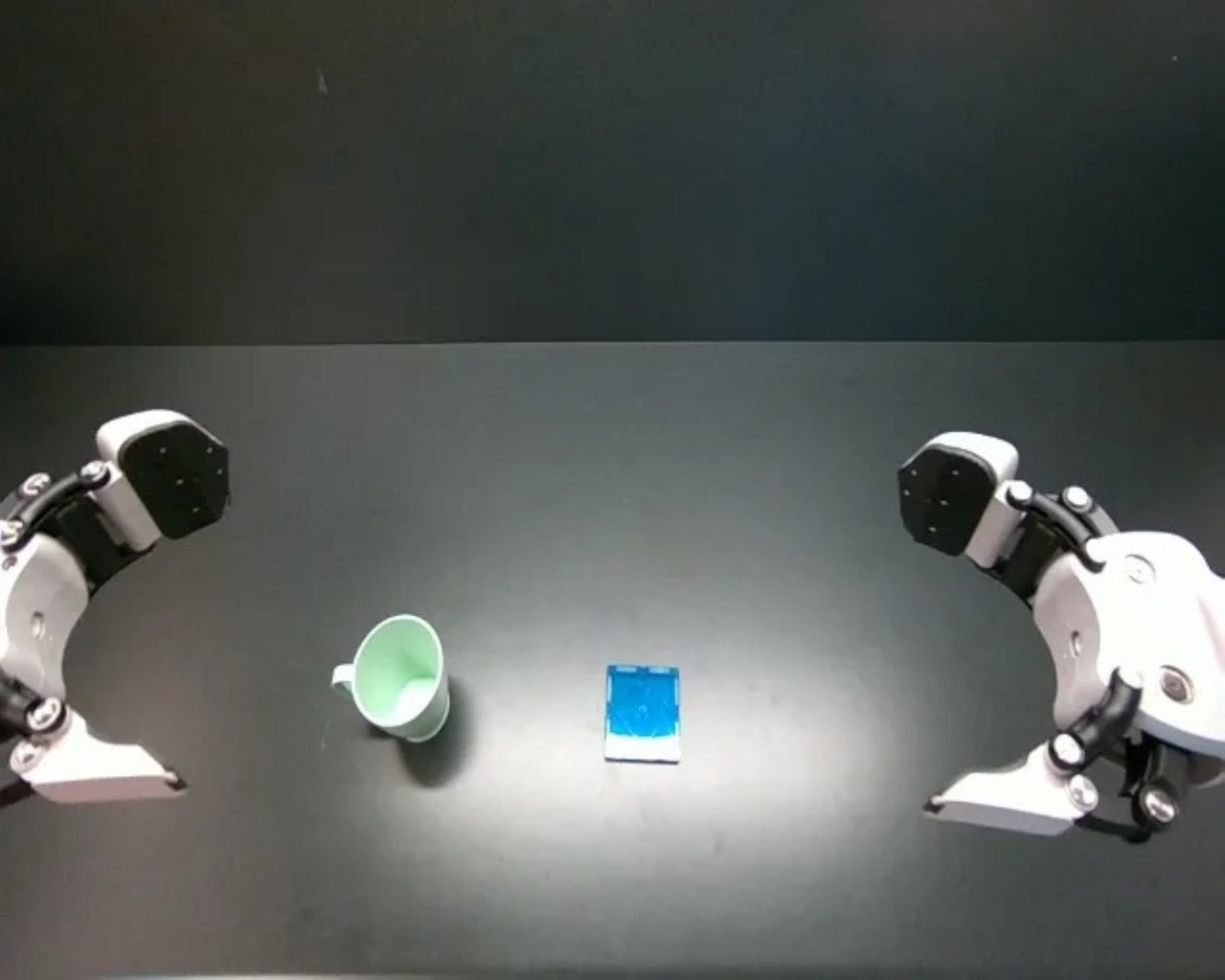} & \centering\scriptsize{The left arm picks up the plastic cup from the table and places it on the blue coaster.} & \includegraphics[width=3.2cm]{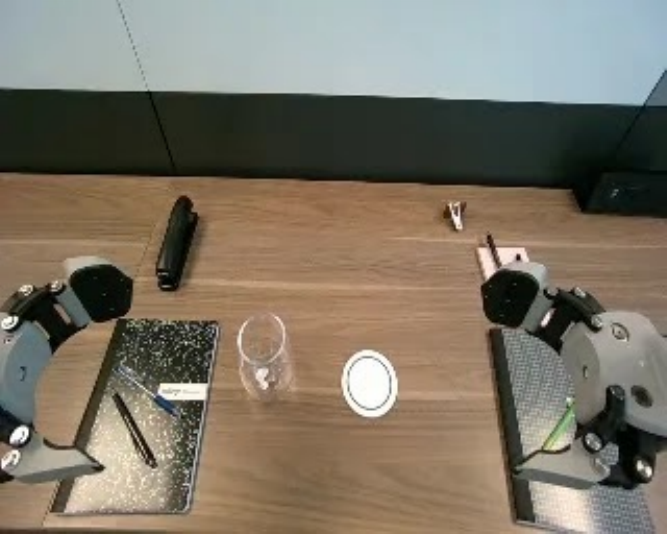} & \centering\scriptsize{The left arm picks up the plastic cup from the table and places it on the white coaster.} & \includegraphics[width=3.2cm]{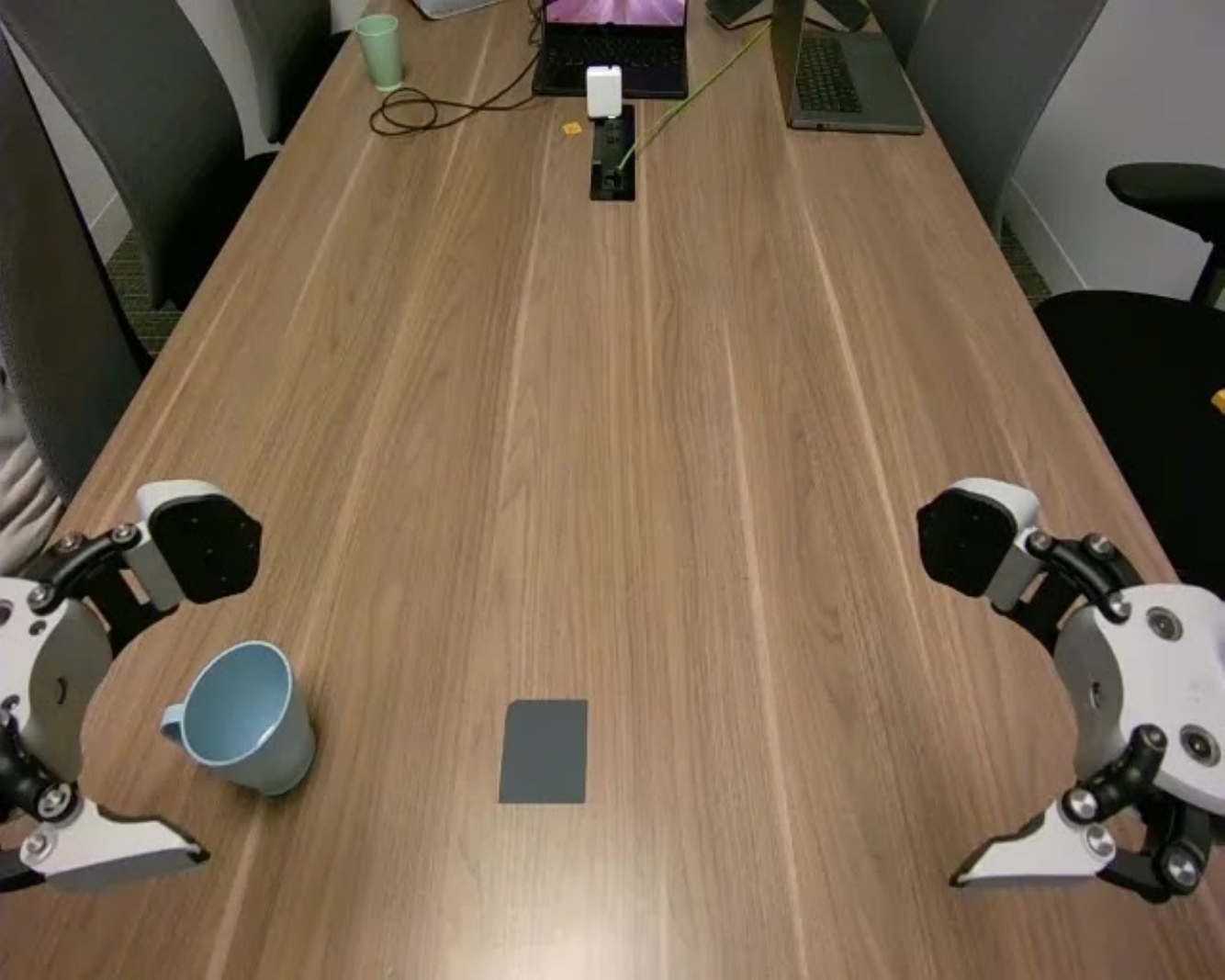} & \centering\arraybackslash\scriptsize{The left arm picks up the pink cup from the table and places it on the gray coaster.} \\
\cmidrule(lr){2-11}
 & 7 & \centering\textbf{Stack Bowls/Cups} & \includegraphics[width=3.2cm]{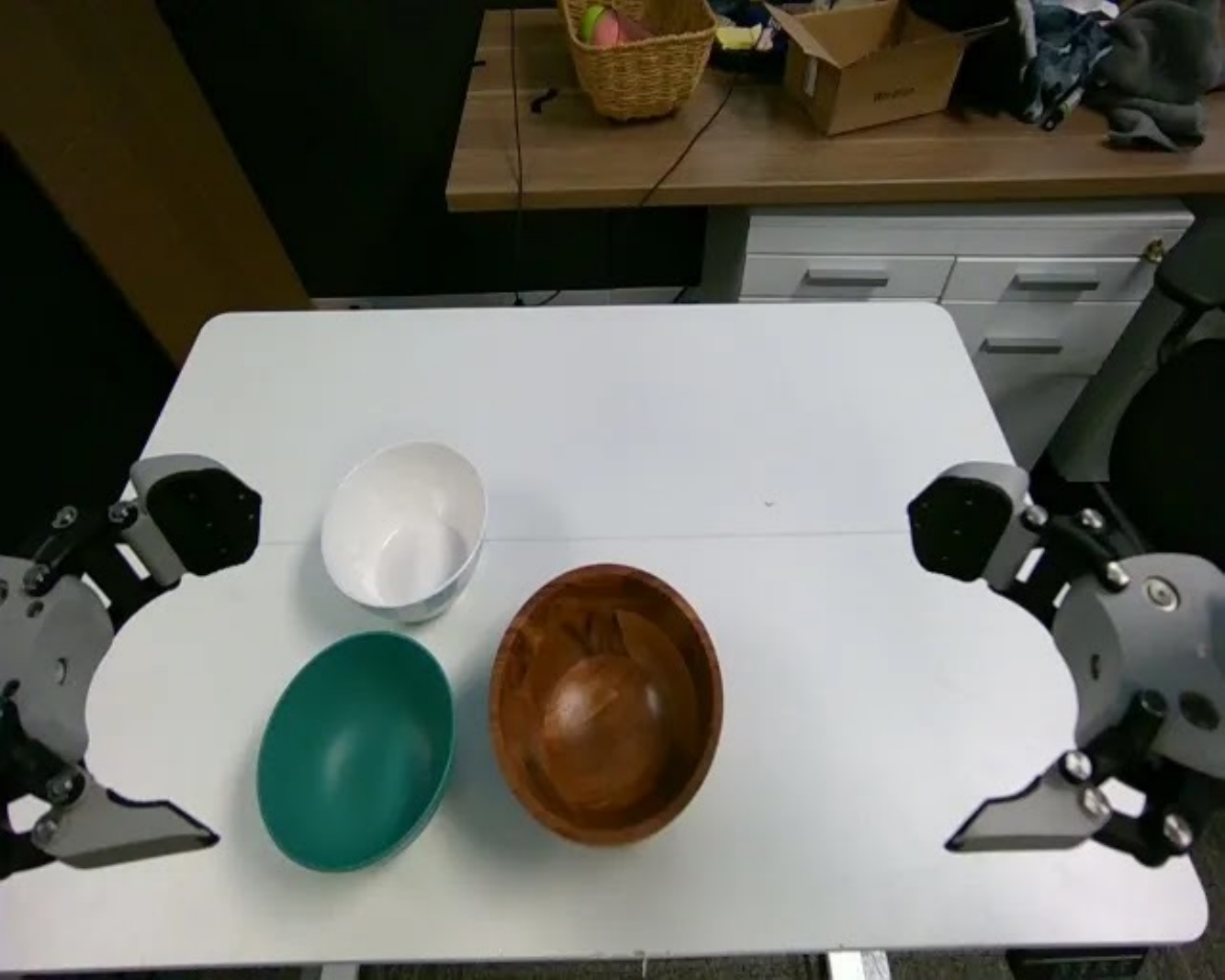} & \centering\scriptsize{The robot reaches to grip the green bowl, moves it to the middle wooden bowl, and releases it to stack. It then reaches to grip the white bowl, moves it to the same location, and releases it onto the stack.} & \includegraphics[width=3.2cm]{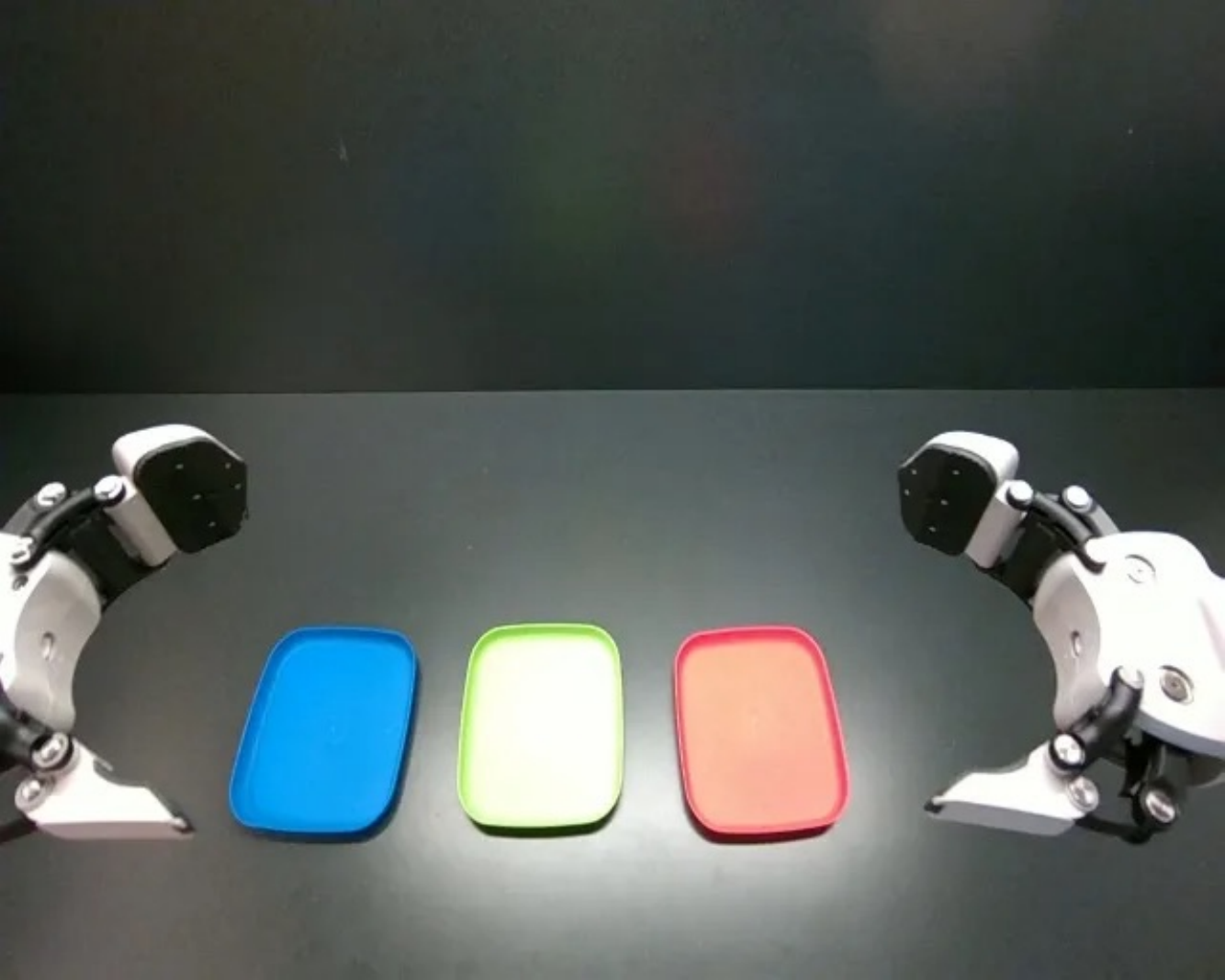} & \centering\scriptsize{The robot reaches its left arm to grip the blue plate, moves it to the middle light green plate, and releases it to stack. It then reaches its right arm to grip the red plate, moves it over the middle stack of plates, and releases it to finish the task.} & \includegraphics[width=3.2cm]{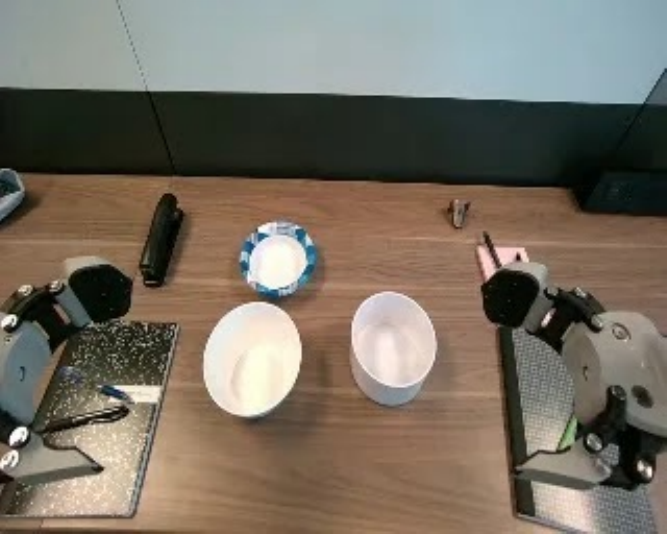} & \centering\scriptsize{The robot reaches its left arm to grip the paper bowl, moves it over the middle white bowl, and releases it to stack. It then reaches its left arm to grip the blue checkered bowl, moves it over the stack, and releases it to finish the task.} & \includegraphics[width=3.2cm]{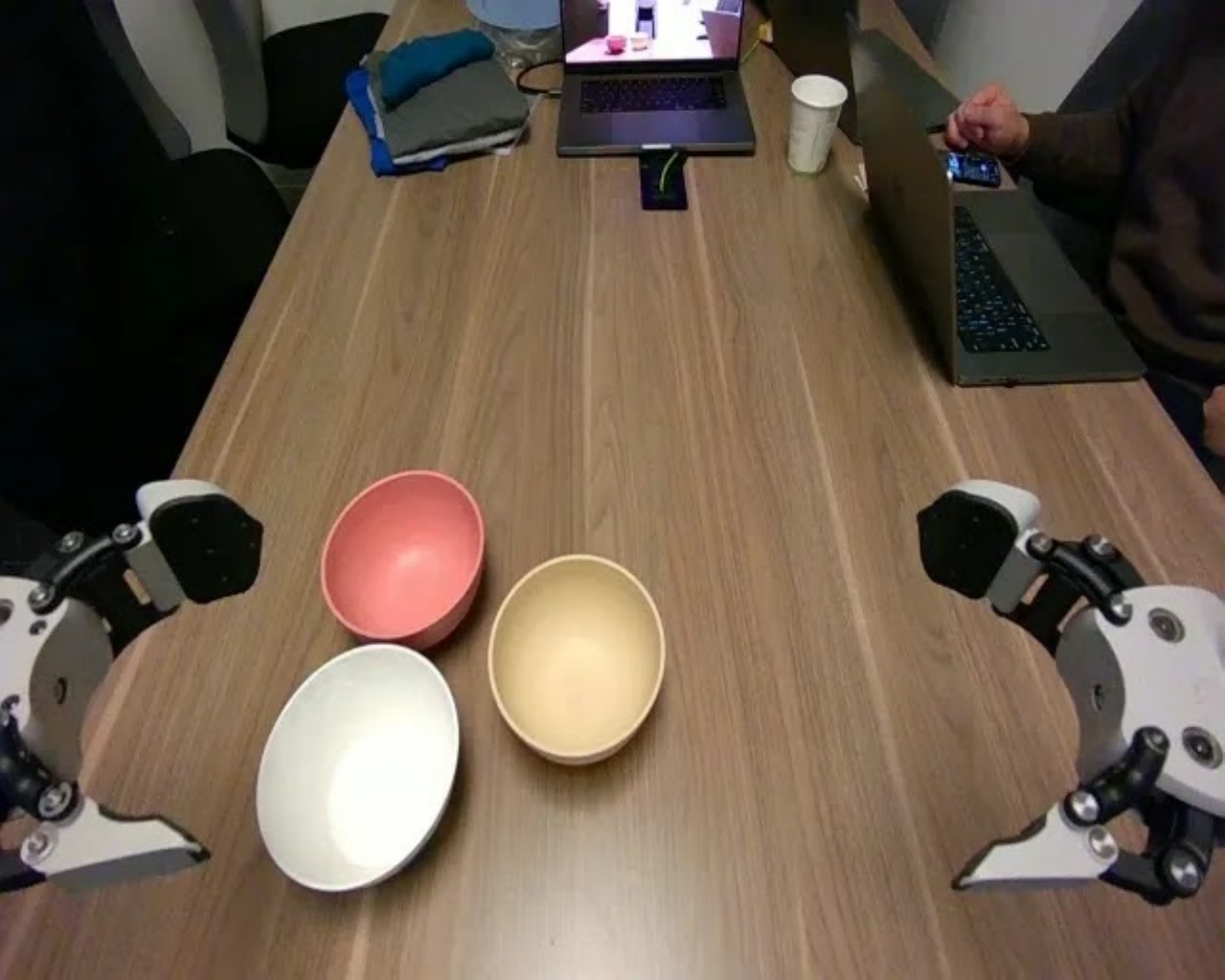} & \centering\arraybackslash\scriptsize{The robot reaches its left arm to grip the pink bowl, moves it over the middle beige bowl, and releases it to stack. It then reaches its left arm to grip the white bowl, moves it over the stack, and releases it onto the stack.} \\
\midrule
 & 8 & \centering\textbf{Folding Shirts} & \includegraphics[width=3.2cm]{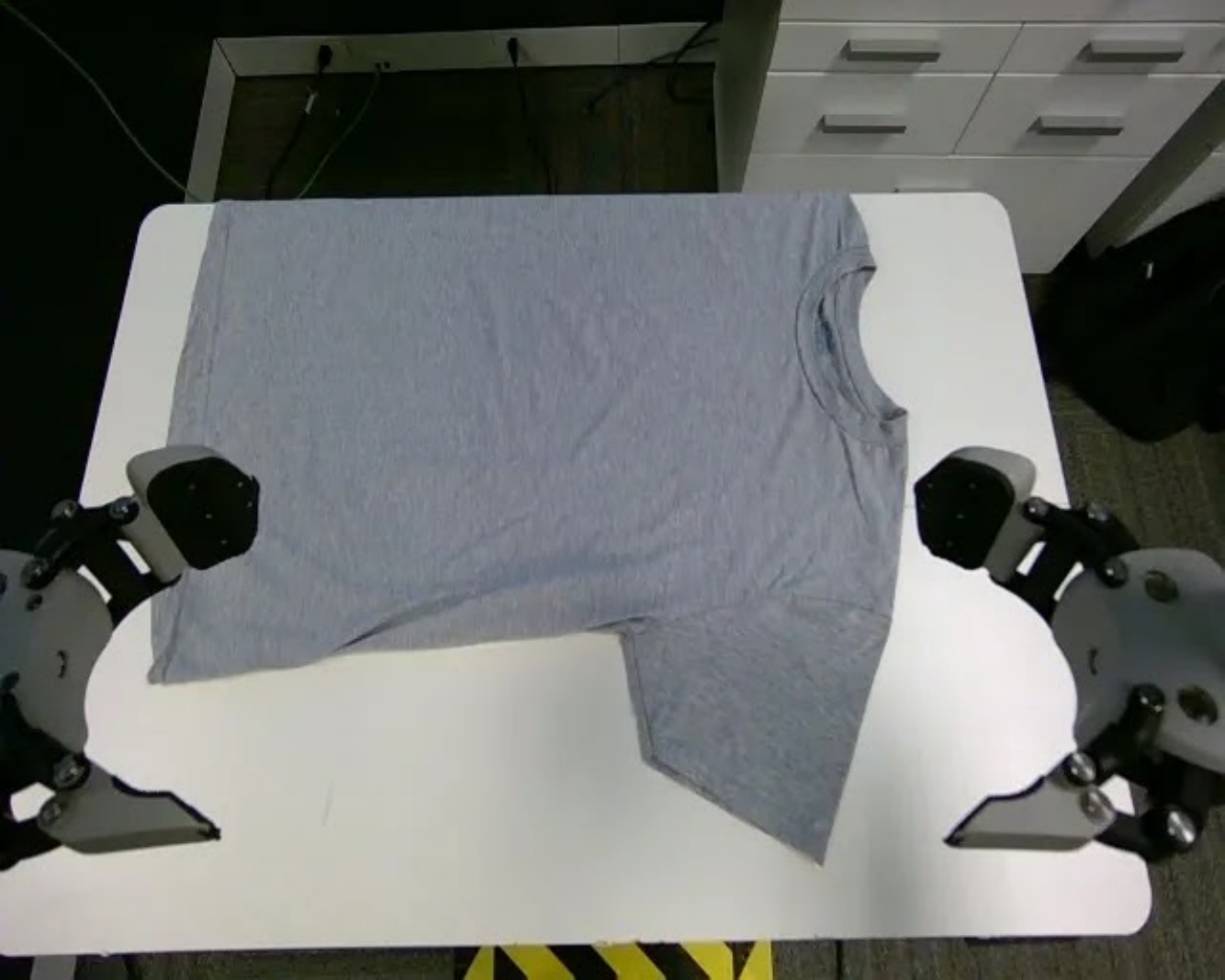} & \centering\scriptsize{Both arms grip the bottom of the light grey short sleeve and fold it toward the middle. They then pull the short sleeve across the table to the edge. Next, both arms grasp the top of the shirt and fold it down to the middle. Finally, the right arm grips the collar and folds it down to complete the task.} & \includegraphics[width=3.2cm]{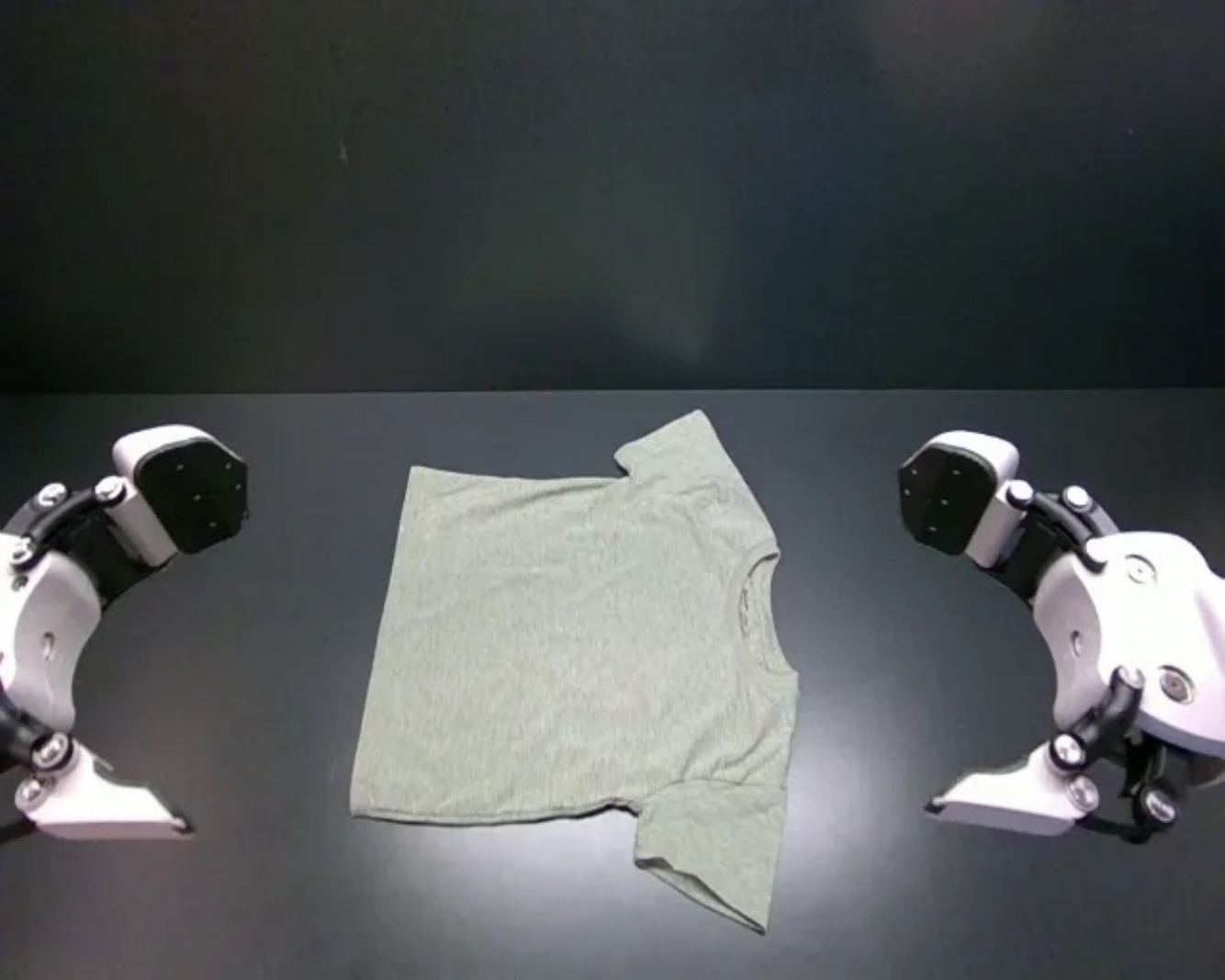} & \centering\scriptsize{Both arms fold the bottom of the green short sleeve to the middle. They then pull the shirt toward the edge of the table. Next, both arms fold the top of the shirt down to the middle. Finally, the right arm grasps the collar and folds it down to complete the task.} & \includegraphics[width=3.2cm]{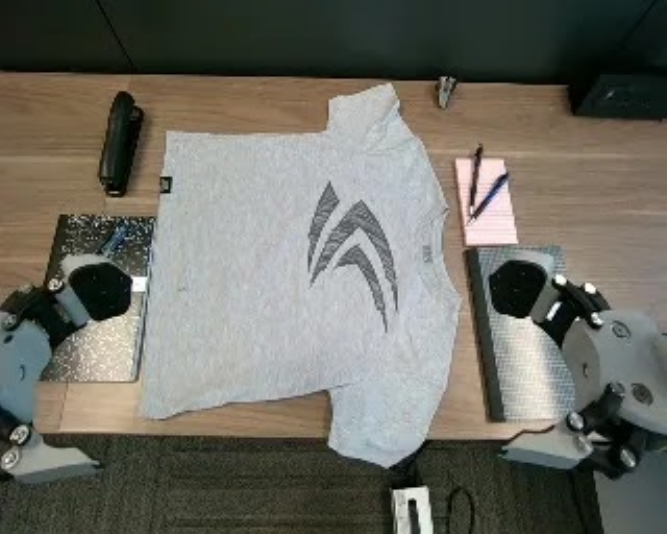} & \centering\scriptsize{Both arms fold the bottom of the logo short sleeve to the middle. They then pull the shirt toward the edge of the table. Next, both arms fold the top of the shirt down to the middle. Finally, the right arm grasps the collar and folds it down to complete the task.} & \includegraphics[width=3.2cm]{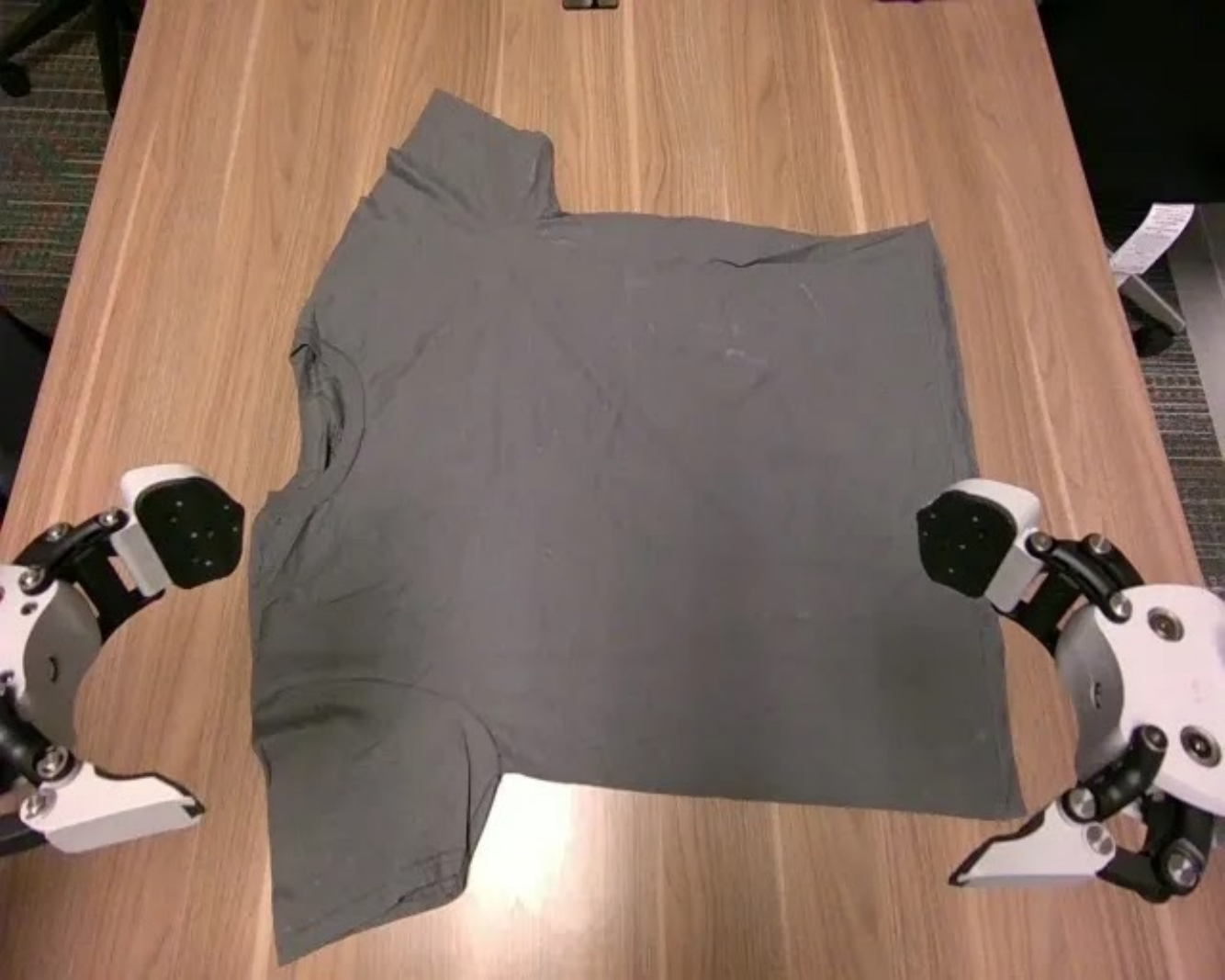} & \centering\arraybackslash\scriptsize{Both arms fold the bottom of the gray short sleeve to the middle. They then pull the shirt toward the edge of the table. Next, both arms fold the top of the shirt down to the middle. Finally, the left arm grasps the collar and folds it down to finish.} \\
\cmidrule(lr){2-11}
\centering\textbf{\shortstack{Contact \\ Rich}} & 9 & \centering\textbf{Folding Shorts} & \includegraphics[width=3.2cm]{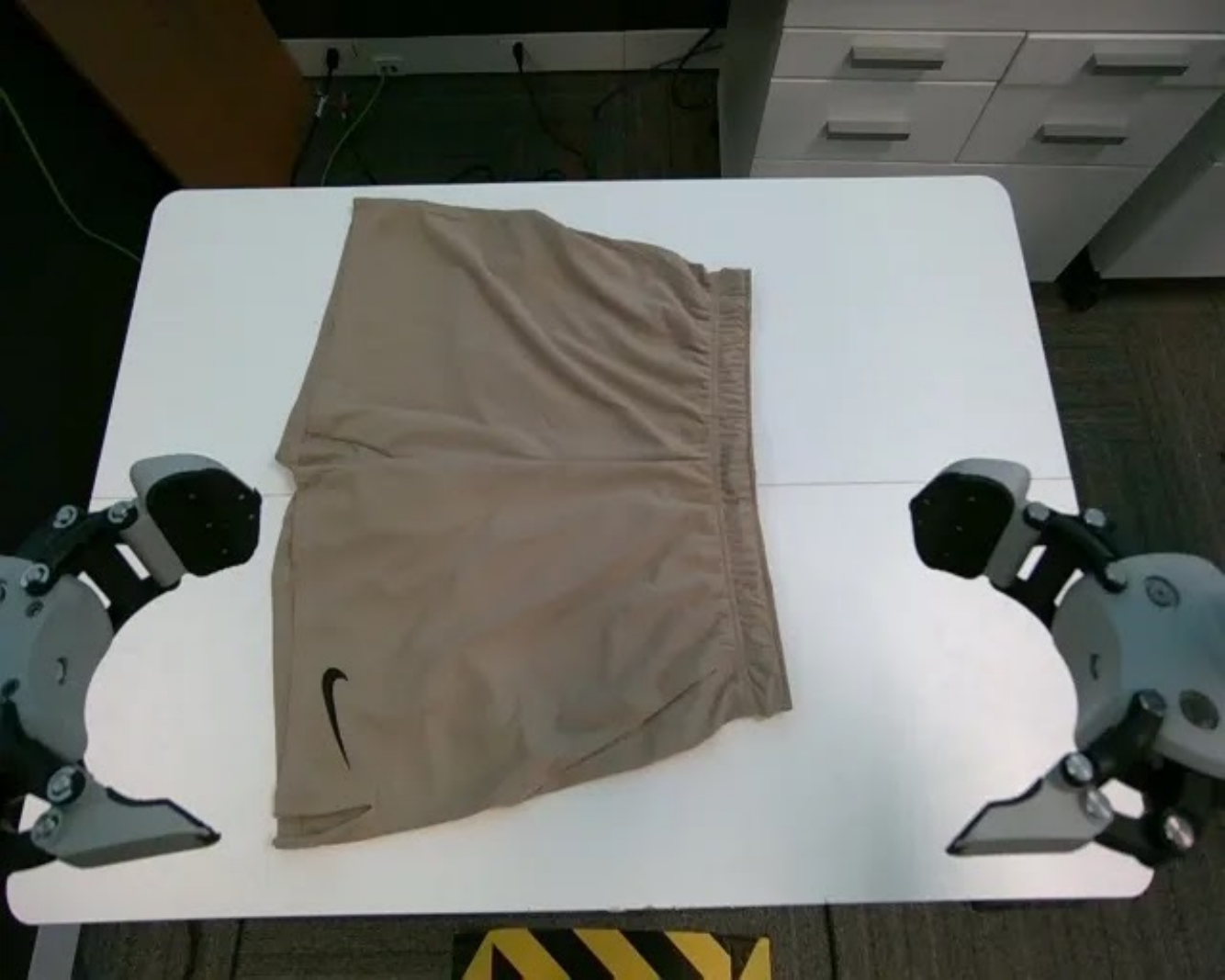} & \centering\scriptsize{Both arms fold the bottom of the tan shorts toward the middle. Then, the right arm folds the shorts in half from right to left to complete the fold.} & \includegraphics[width=3.2cm]{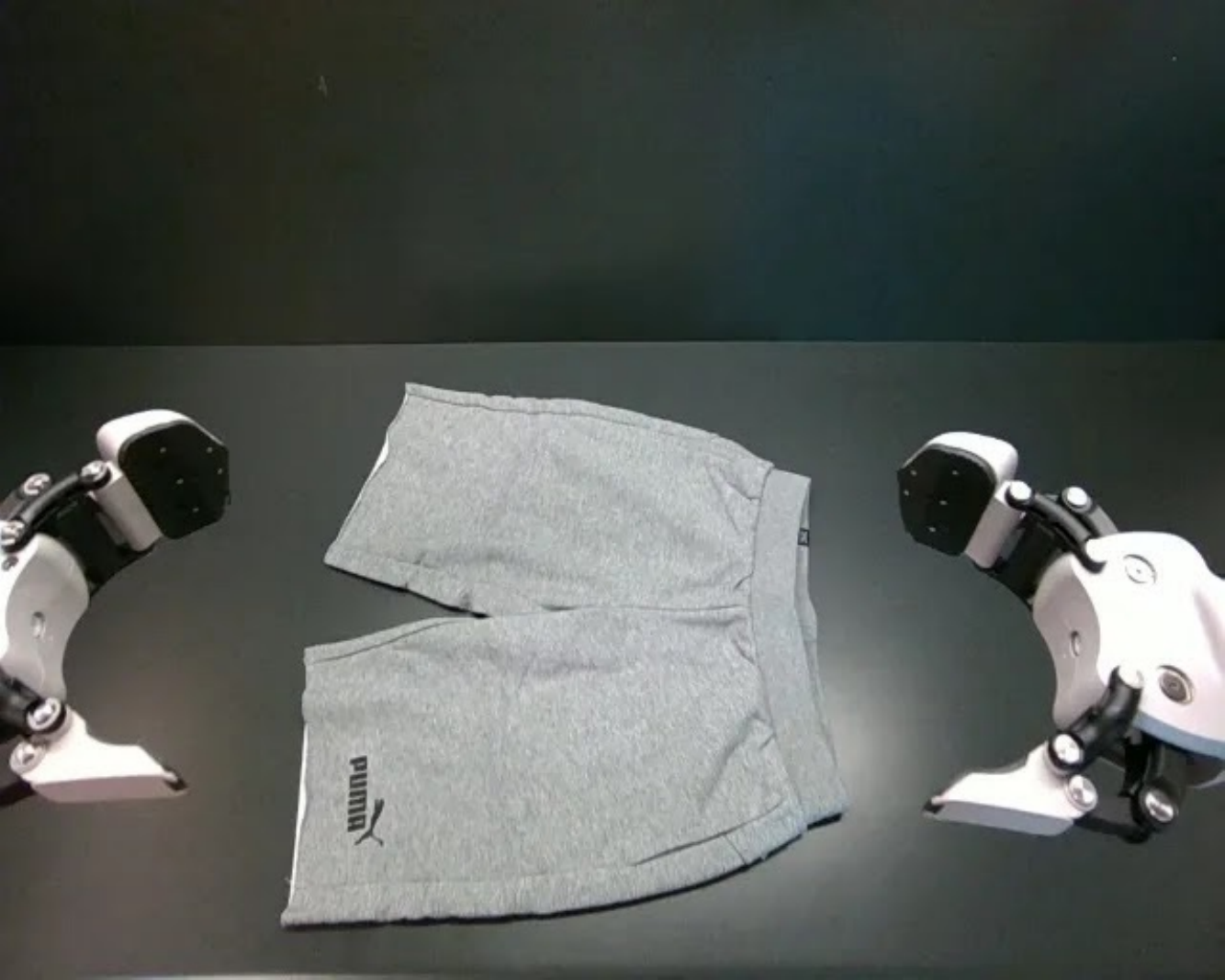} & \centering\scriptsize{Both arms fold the bottom of the grey shorts toward the middle. Then, the right arm folds the shorts in half from right to left to complete the fold.} & \includegraphics[width=3.2cm]{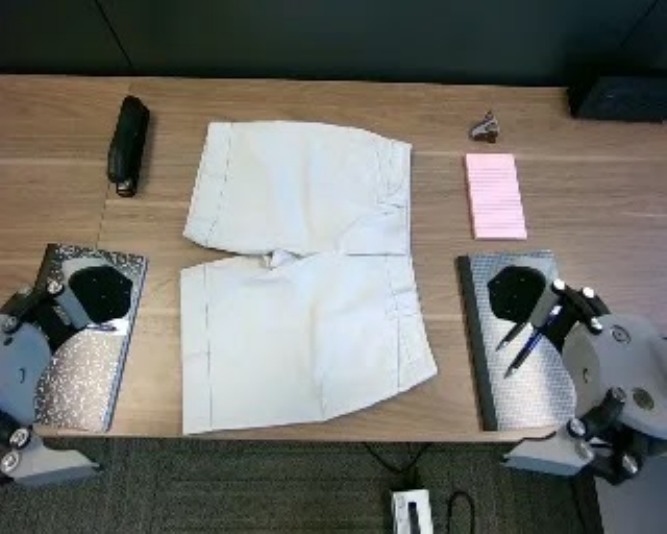} & \centering\scriptsize{Both arms fold the bottom of the white shorts toward the middle. Then, the right arm folds the shorts in half from right to left to complete the task.} & \includegraphics[width=3.2cm]{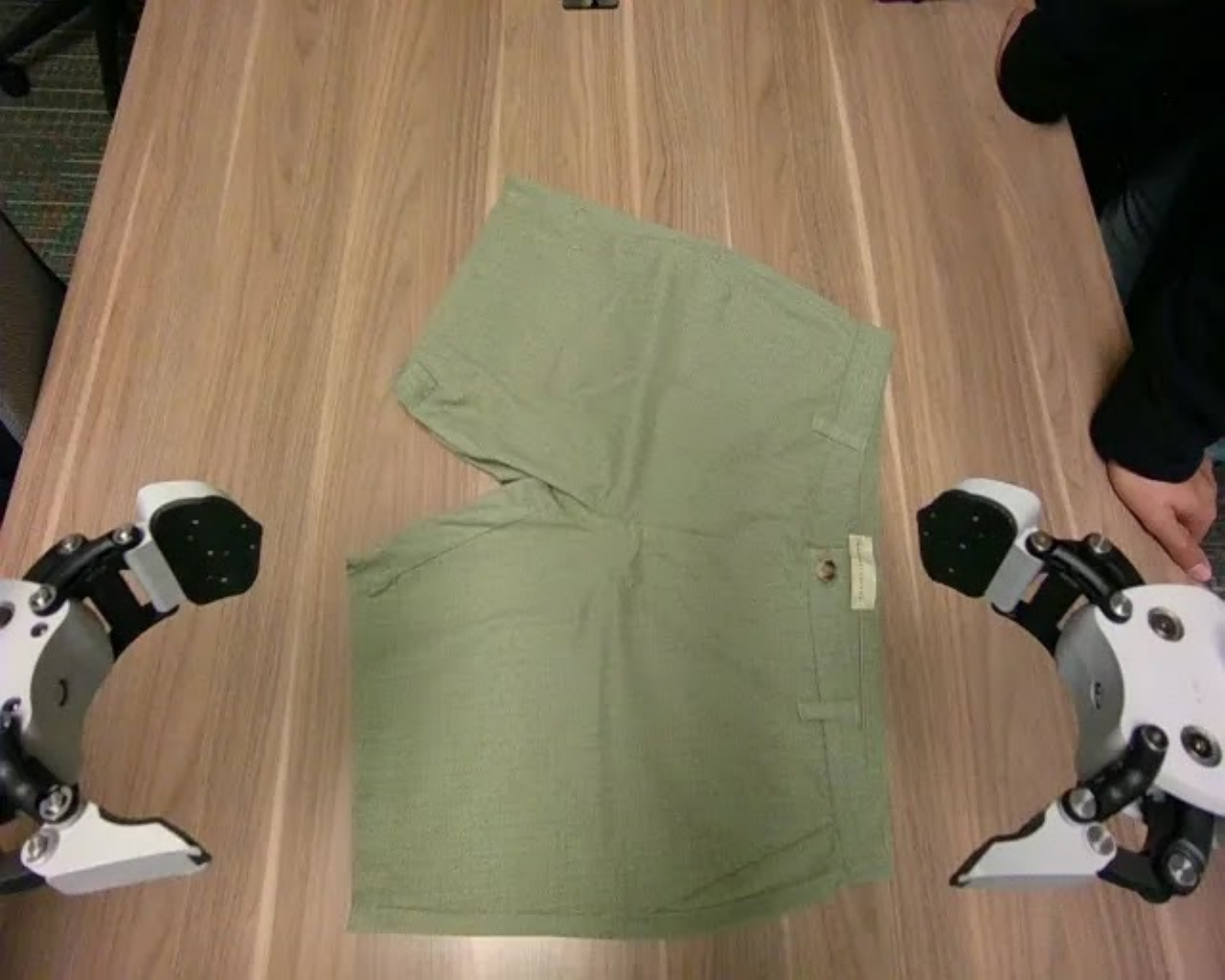} & \centering\arraybackslash\scriptsize{Both arms fold the bottom of the green shorts toward the middle. Then, the right arm folds the shorts in half from right to left to complete the fold.} \\
\cmidrule(lr){2-11}
 & 10 & \centering\textbf{Stacking Clothes} & \includegraphics[width=3.2cm]{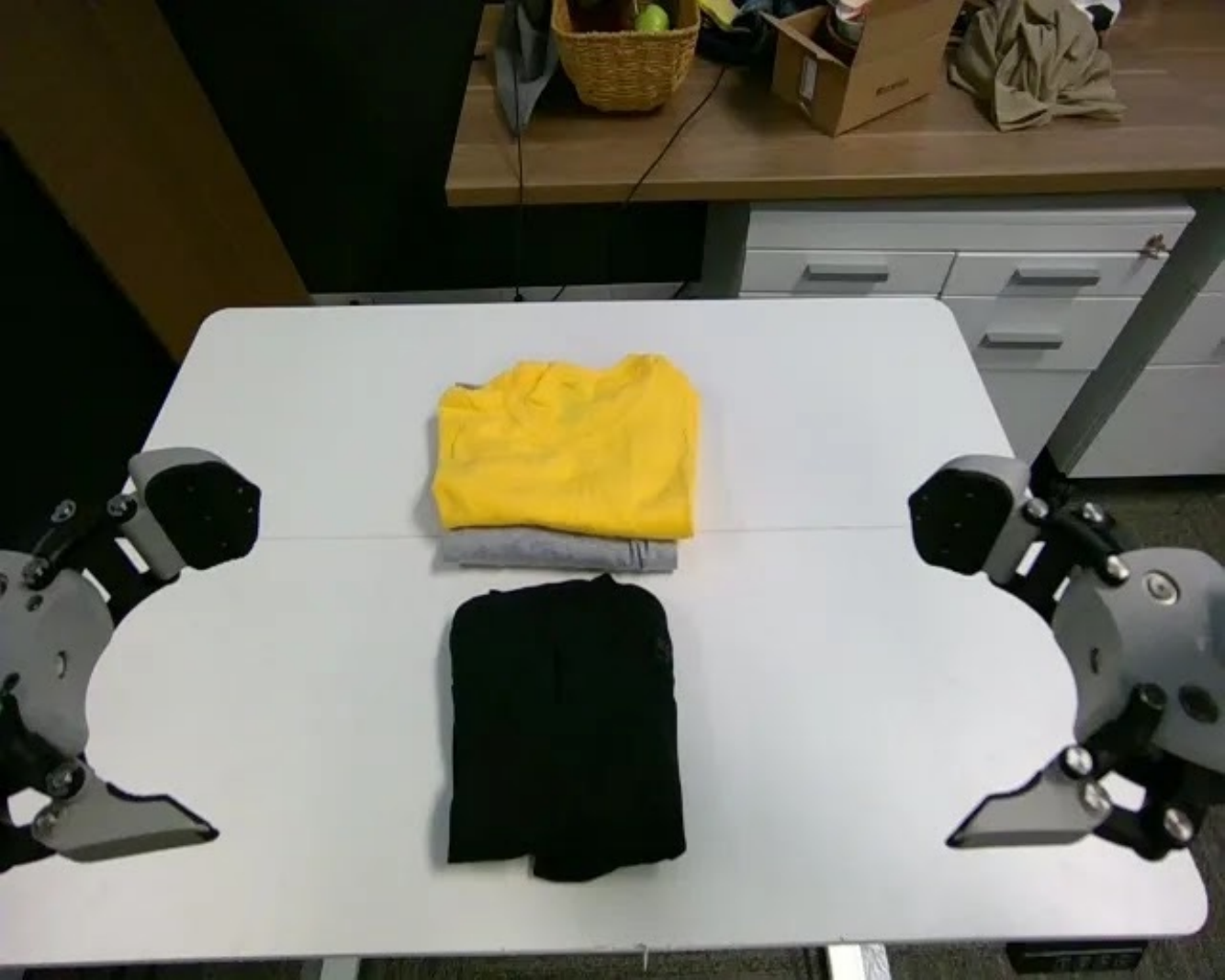} & \centering\scriptsize{Both arms pick up the black shirt and place it on the stack of clothes.} & \includegraphics[width=3.2cm]{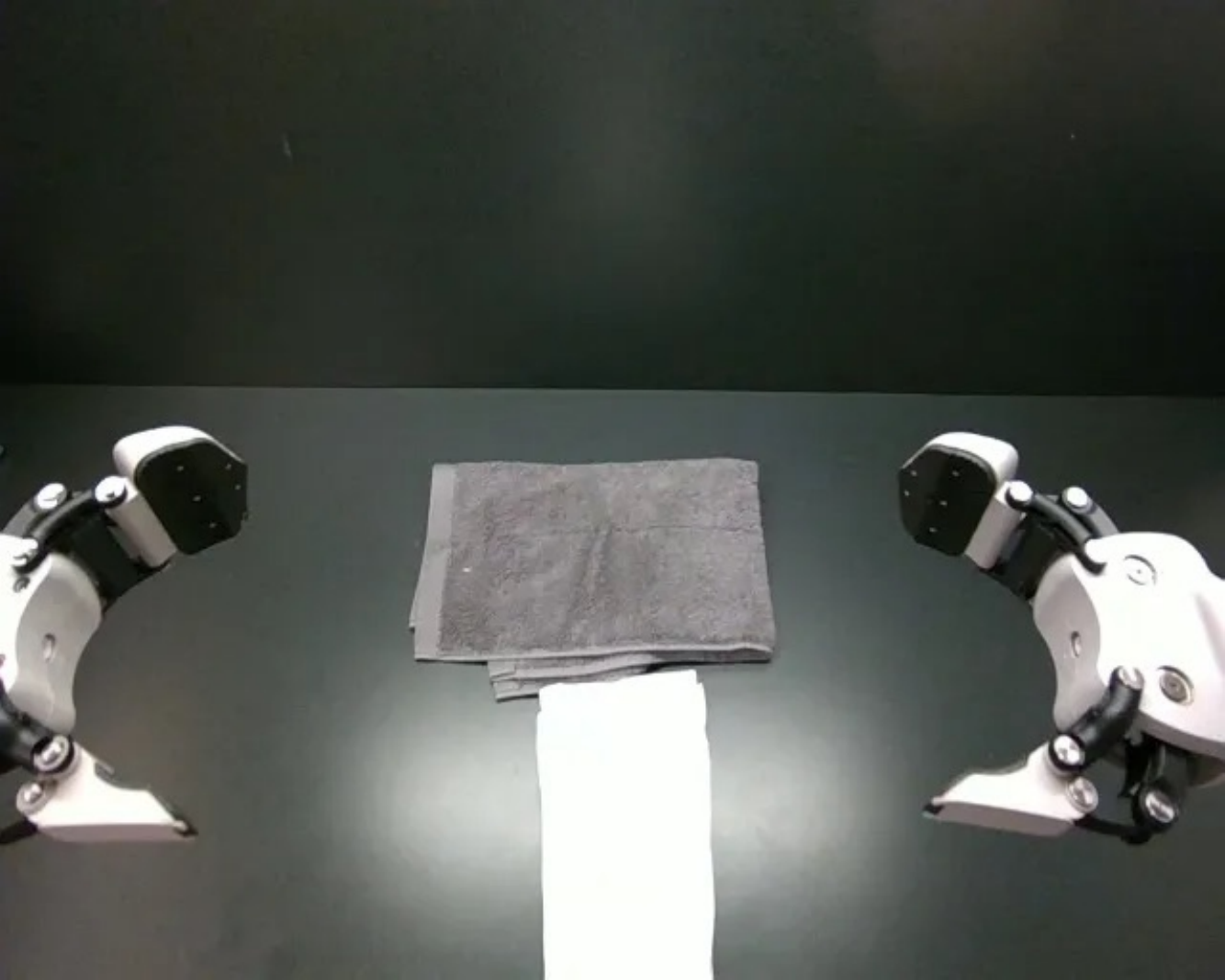} & \centering\scriptsize{Both arms pick up the white shirt and place it on the gray towel.} & \includegraphics[width=3.2cm]{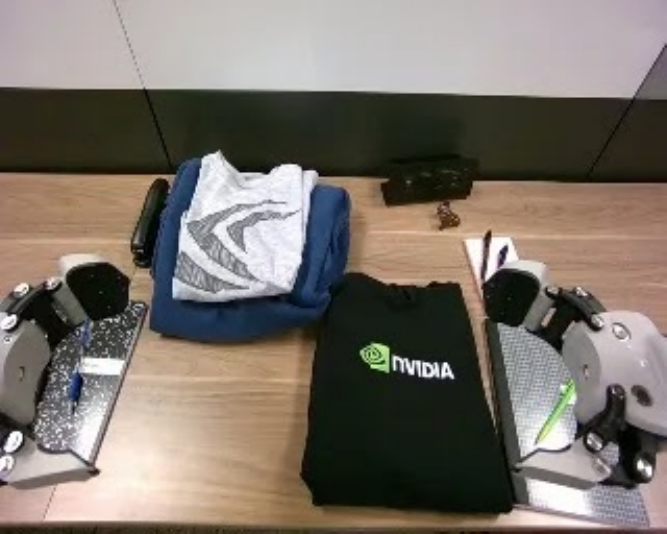} & \centering\scriptsize{Both arms pick up the black hoodie and place it on the stack of clothes.} & \includegraphics[width=3.2cm]{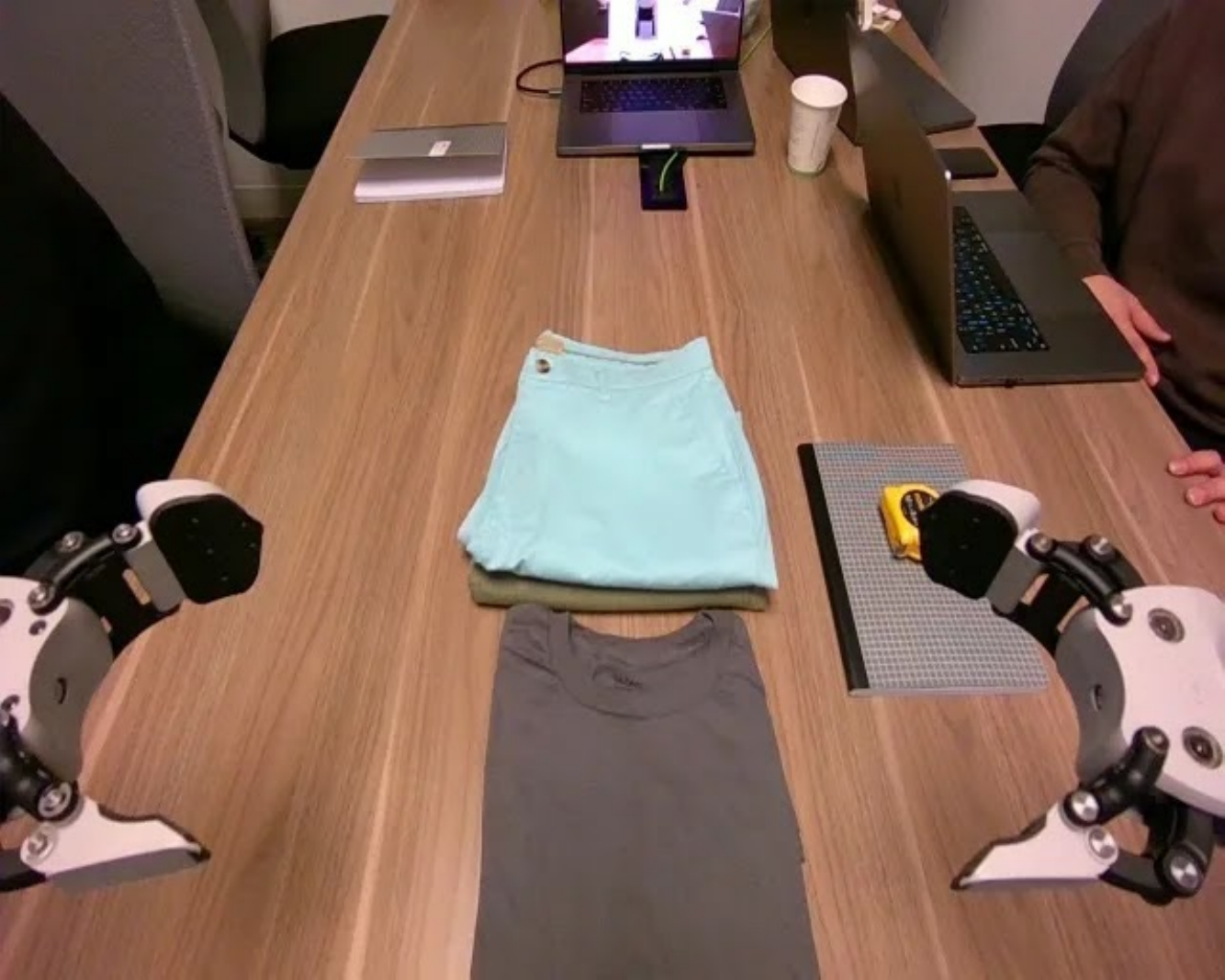} & \centering\arraybackslash\scriptsize{Both arms pick up the dark gray shirt and place it on the stack of clothes.} \\
\bottomrule
\end{tabular}
}
\end{table*}

\begin{table*}[htbp]
\centering
\caption{\textbf{Unseen Tasks Evaluation Setup for AgiBot G1}}
\label{tab:unseen_tasks}
\setlength{\tabcolsep}{2pt}
\resizebox{\textwidth}{!}{
\begin{tabular}{cm{1.8cm}m{3.5cm}m{3.2cm}m{3.2cm}m{3.2cm}m{3.2cm}m{3.2cm}m{3.2cm}m{3.2cm}}
\toprule
\centering\textbf{\#} & \centering\textbf{Task} & \centering\textbf{Image} & \centering\textbf{Instruction} & \centering\textbf{Image} & \centering\textbf{Instruction} & \centering\textbf{Image} & \centering\textbf{Instruction} & \centering\textbf{Image} & \centering\arraybackslash\textbf{Instruction} \\
\midrule
1 & \centering\textbf{Untie Shoelaces} & \includegraphics[width=3.2cm]{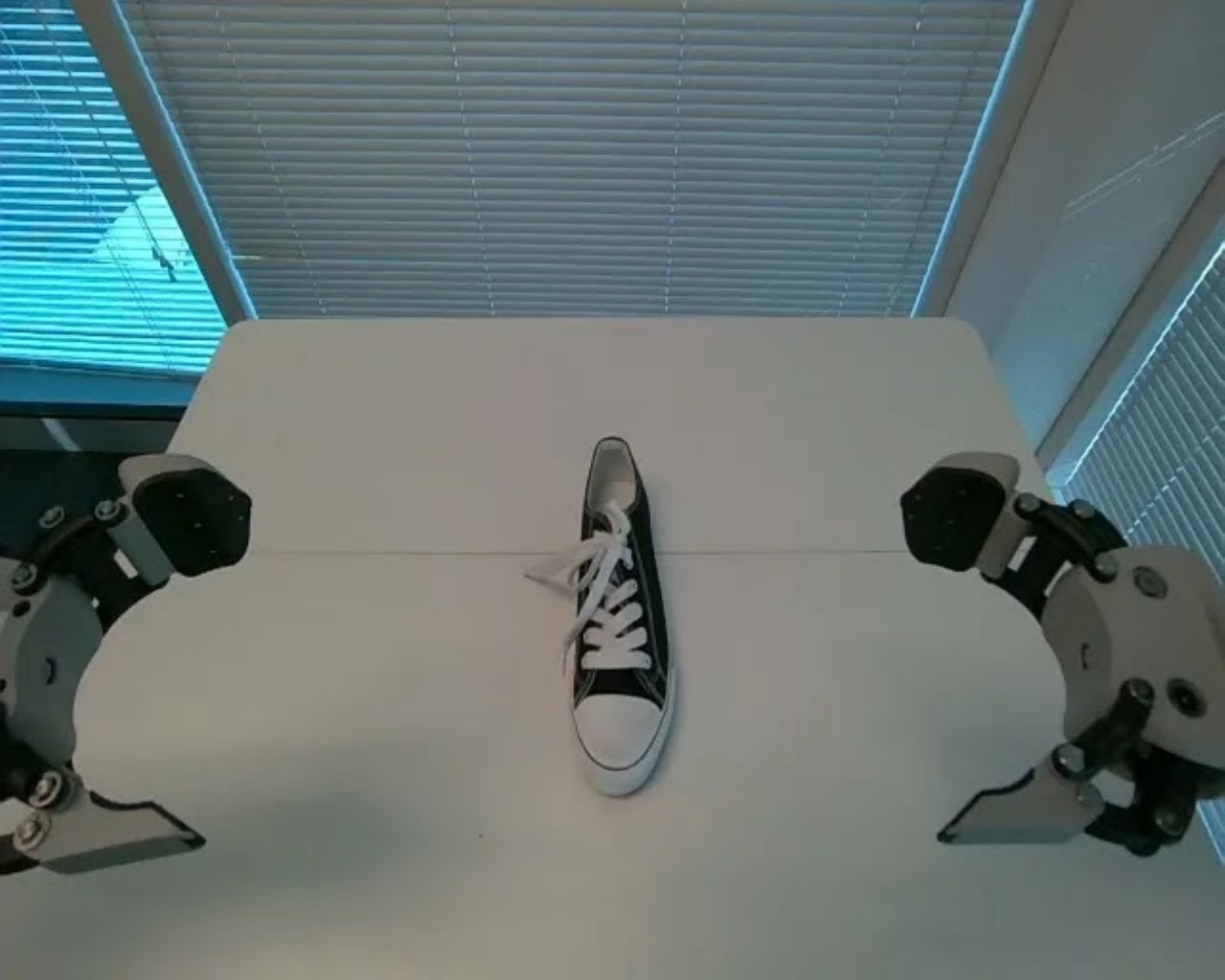} & \centering\scriptsize{The robot coordinates both arms to simultaneously grasp the two loops of the shoelace. It then moves both arms outward in synchronized, opposing directions until the shoelace is fully untied.} & \includegraphics[width=3.2cm]{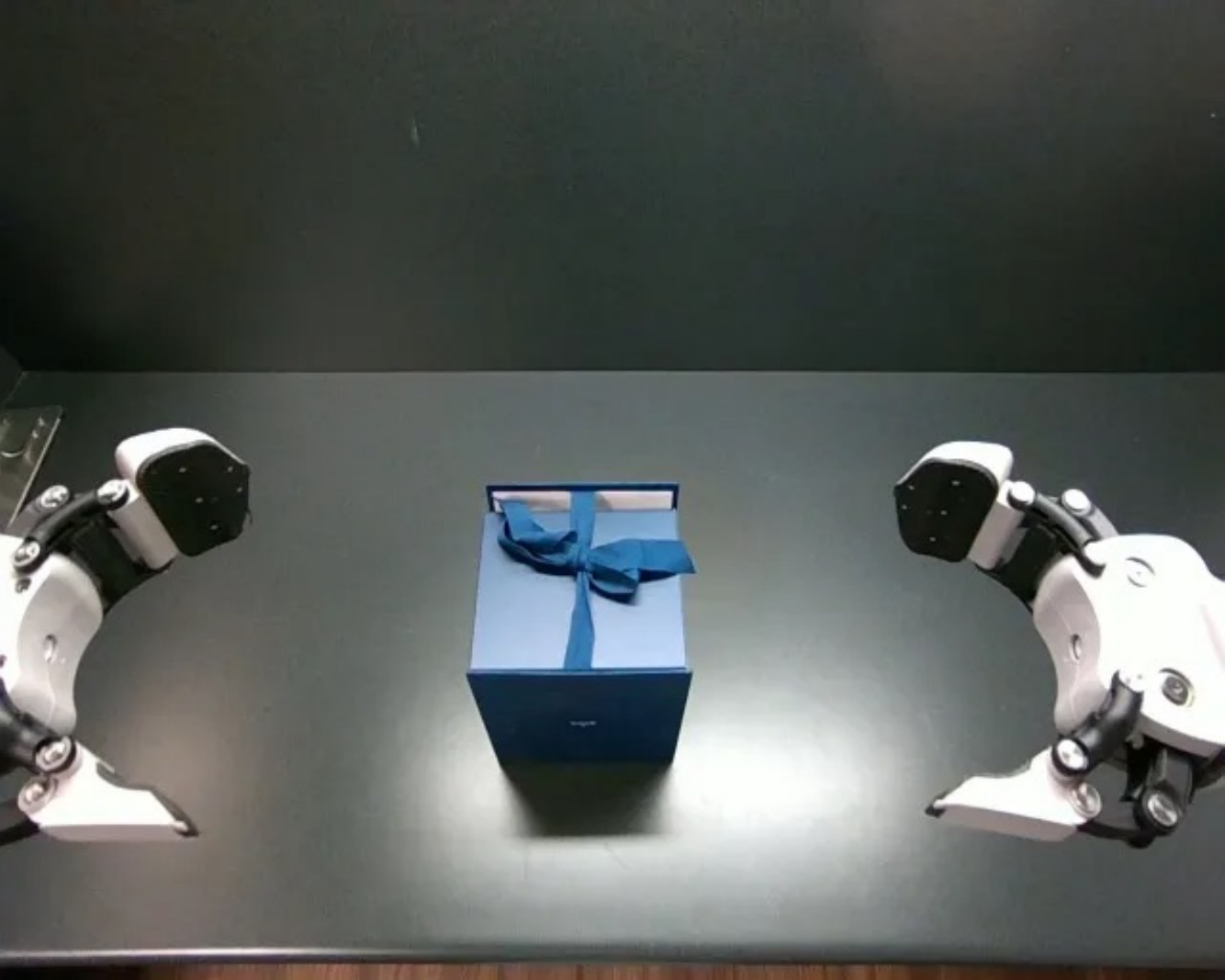} & \centering\scriptsize{The robot reaches its left arm to grasp the blue box and hold it steady. It then reaches its right arm toward the blue ribbon and moves it to untie the knot.} & \includegraphics[width=3.2cm]{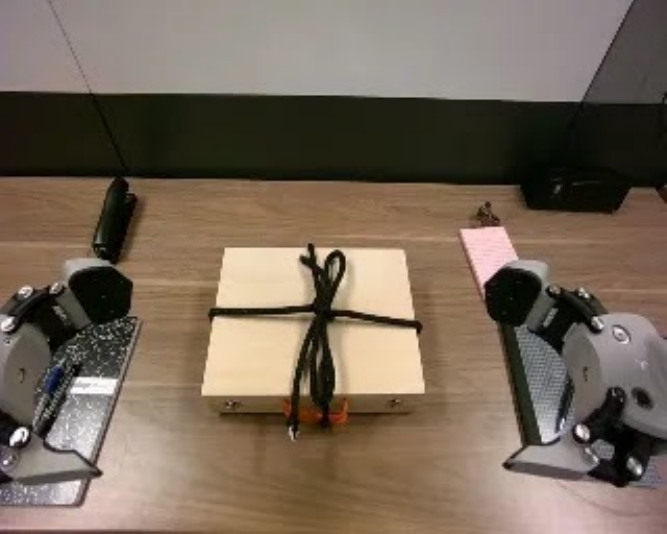} & \centering\scriptsize{The robot coordinates both arms to simultaneously grasp the two loops of the knot of the box. It then moves both arms outward in synchronized, opposing directions until the knot of the box is fully untied.} & \includegraphics[width=3.2cm]{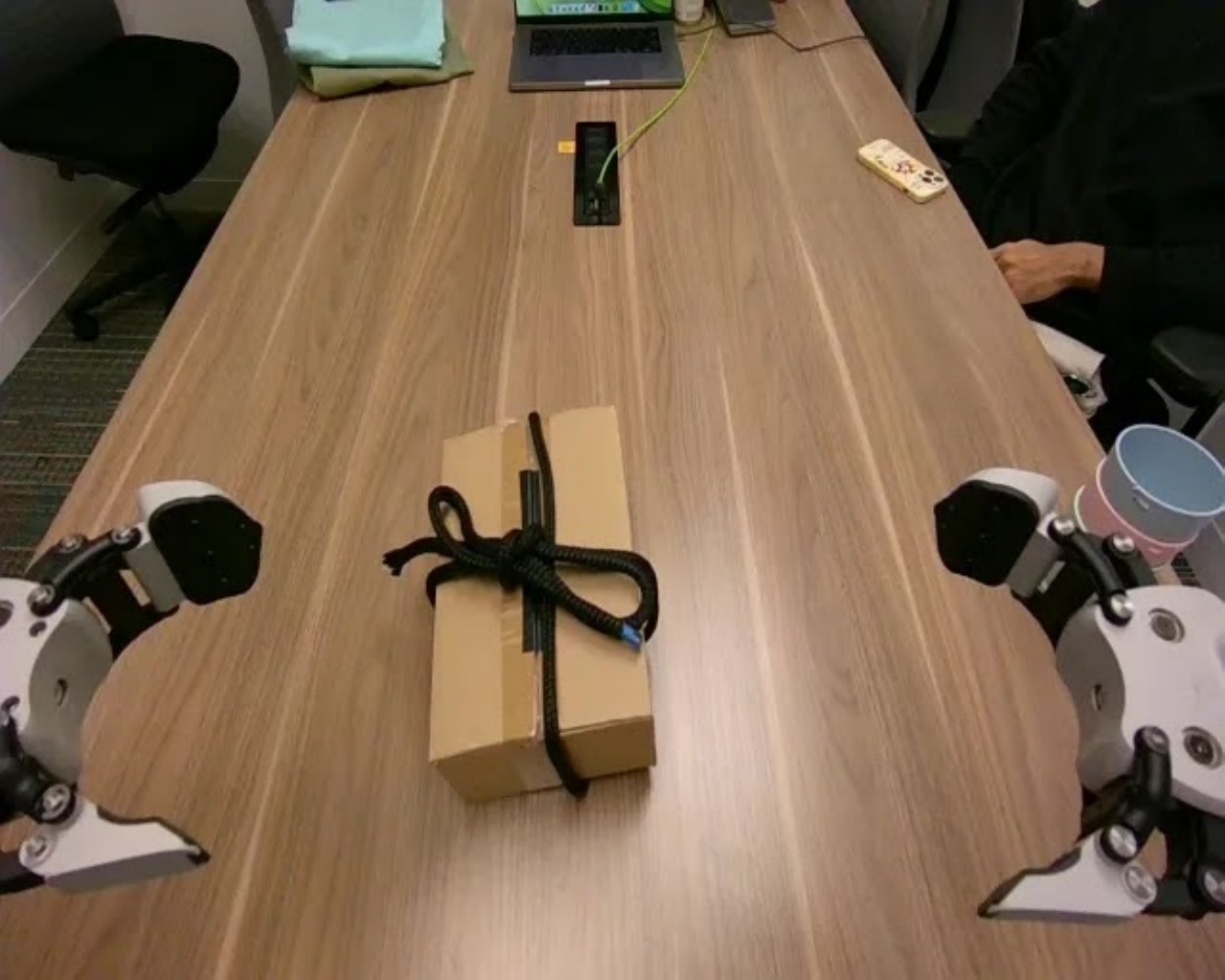} & \centering\arraybackslash\scriptsize{The robot coordinates both arms to simultaneously grasp the two loops of the knot of the package. It then moves both arms outward in synchronized, opposing directions until the knot of the package is fully untied.} \\
\midrule
2 & \centering\textbf{Remove\\/Put Hat} & \includegraphics[width=3.2cm]{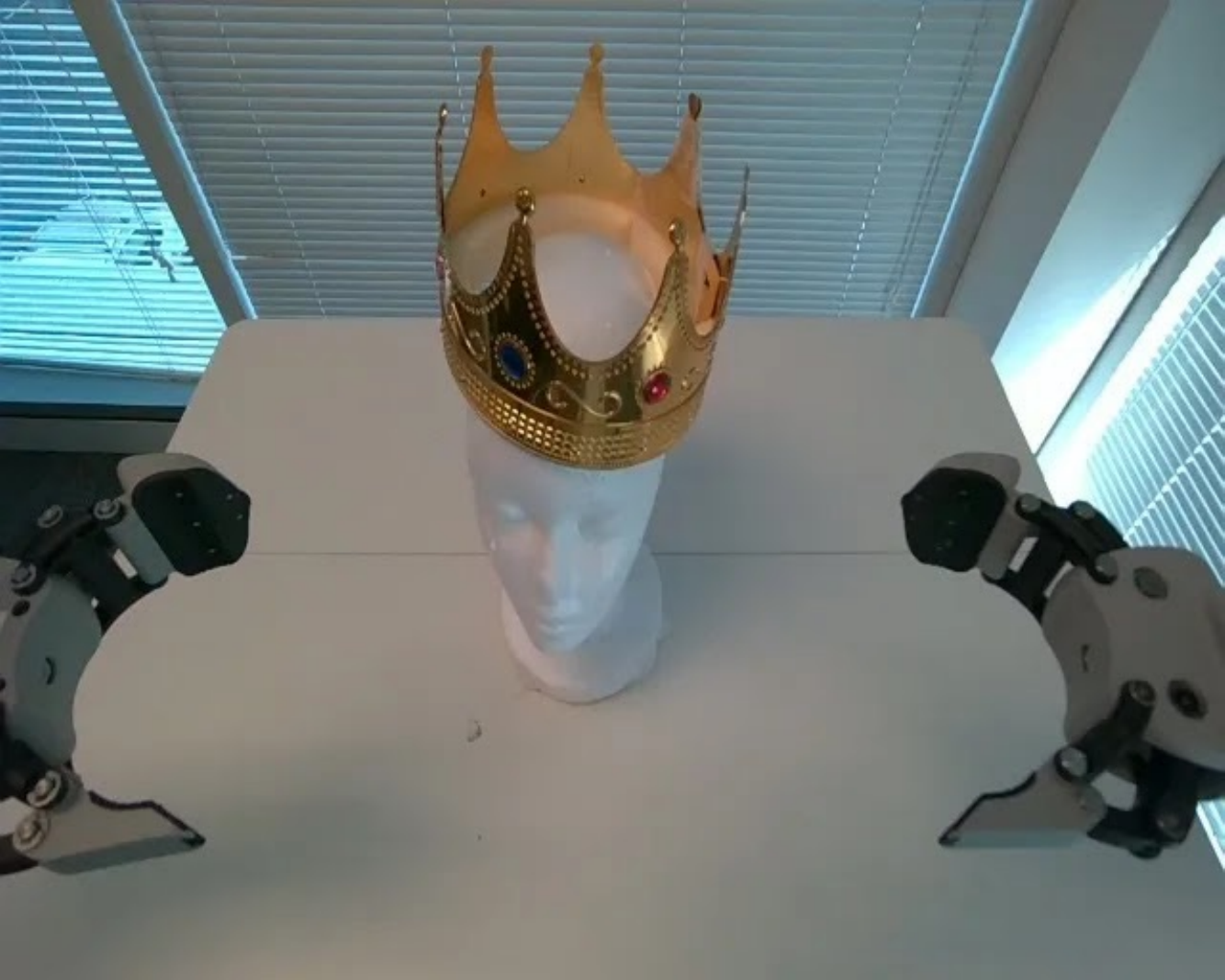} & \centering\scriptsize{The robot reaches its right arm to gasp the crown and lifts it to remove it from the mannequin's head.} & \includegraphics[width=3.2cm]{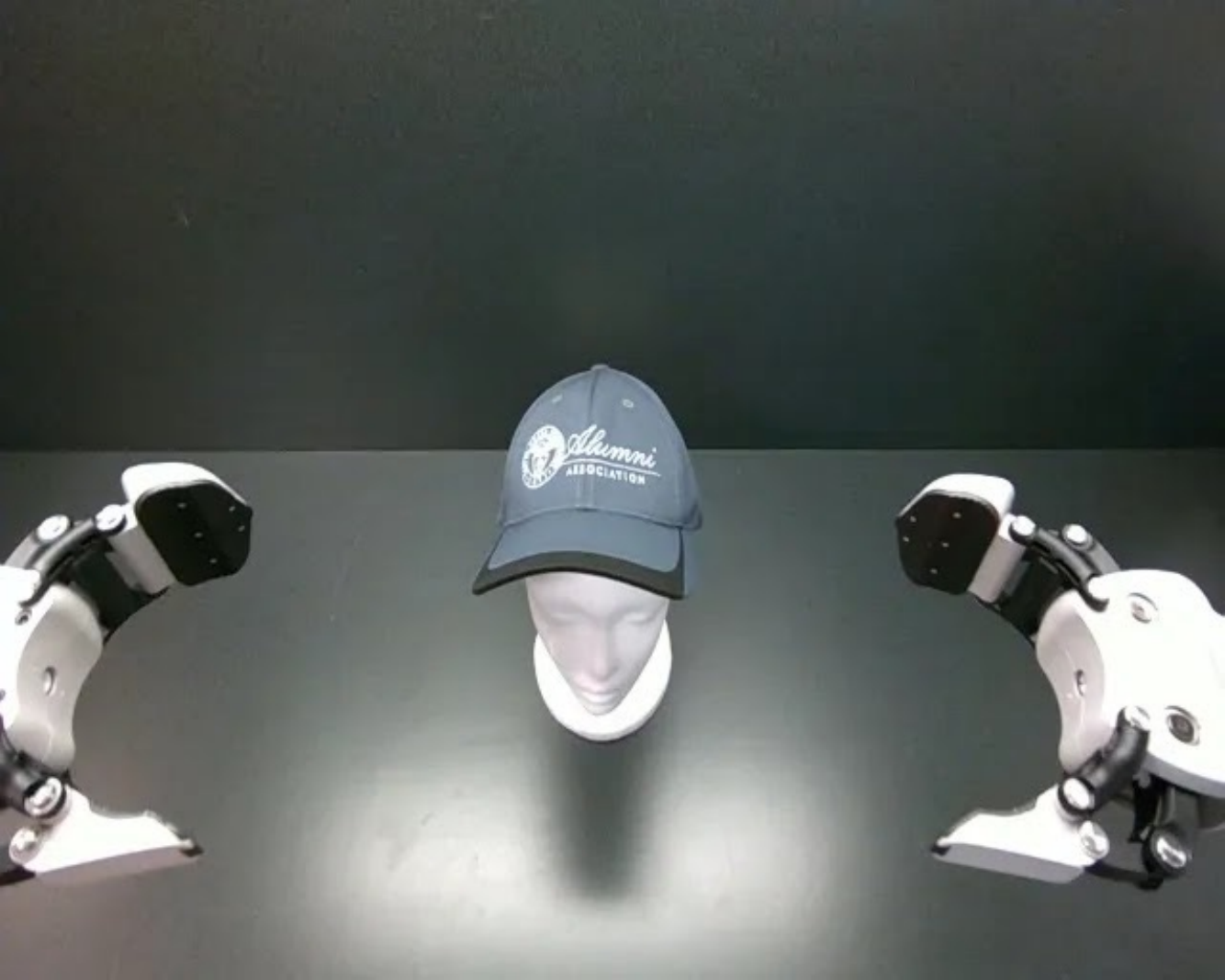} & \centering\scriptsize{The robot reaches its left arm to grasp the hat and lifts it to remove it from the mannequin's head.} & \includegraphics[width=3.2cm]{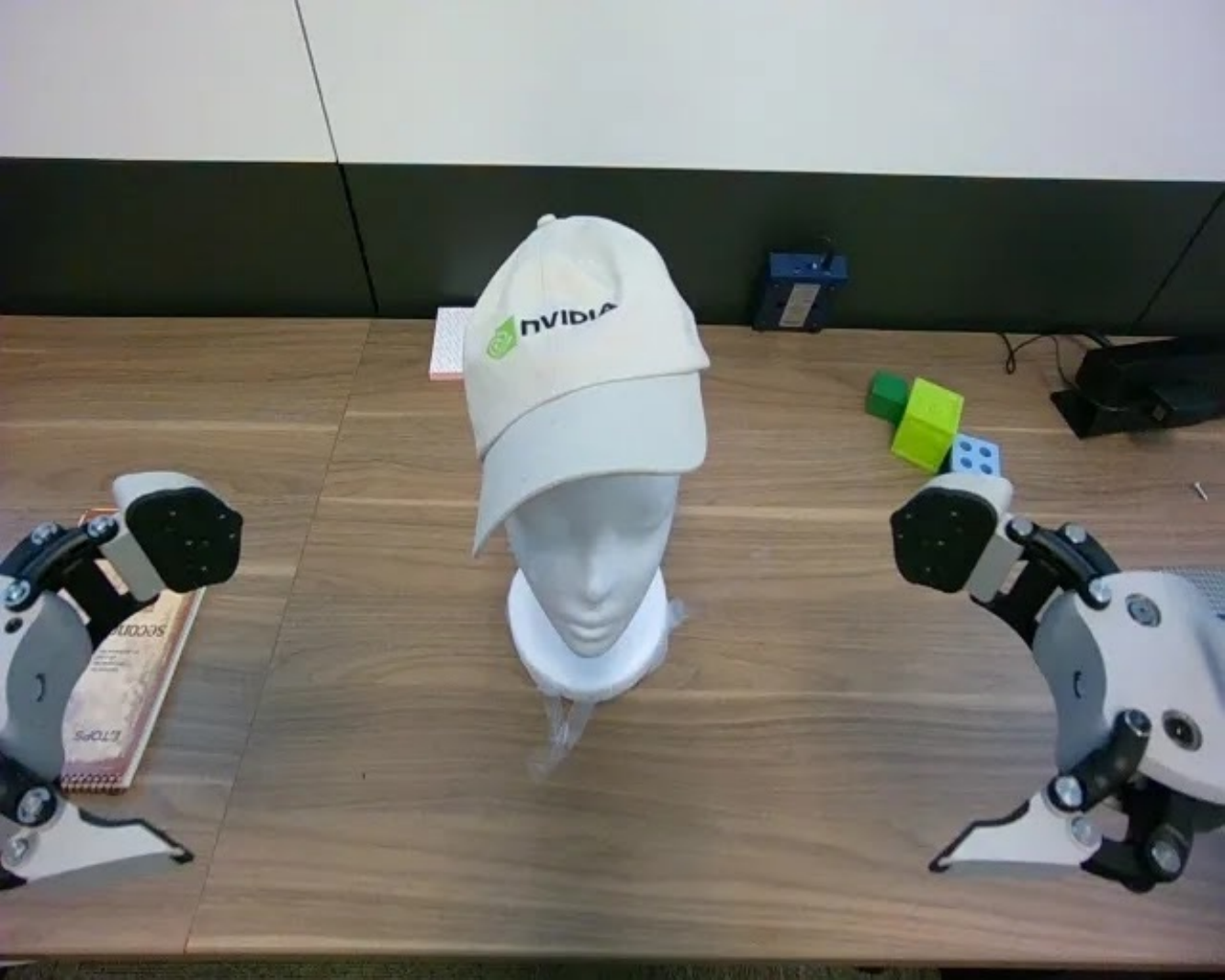} & \centering\scriptsize{The robot reaches its left arm to grasp the hat and lifts it to remove it from the mannequin's head.} & \includegraphics[width=3.2cm]{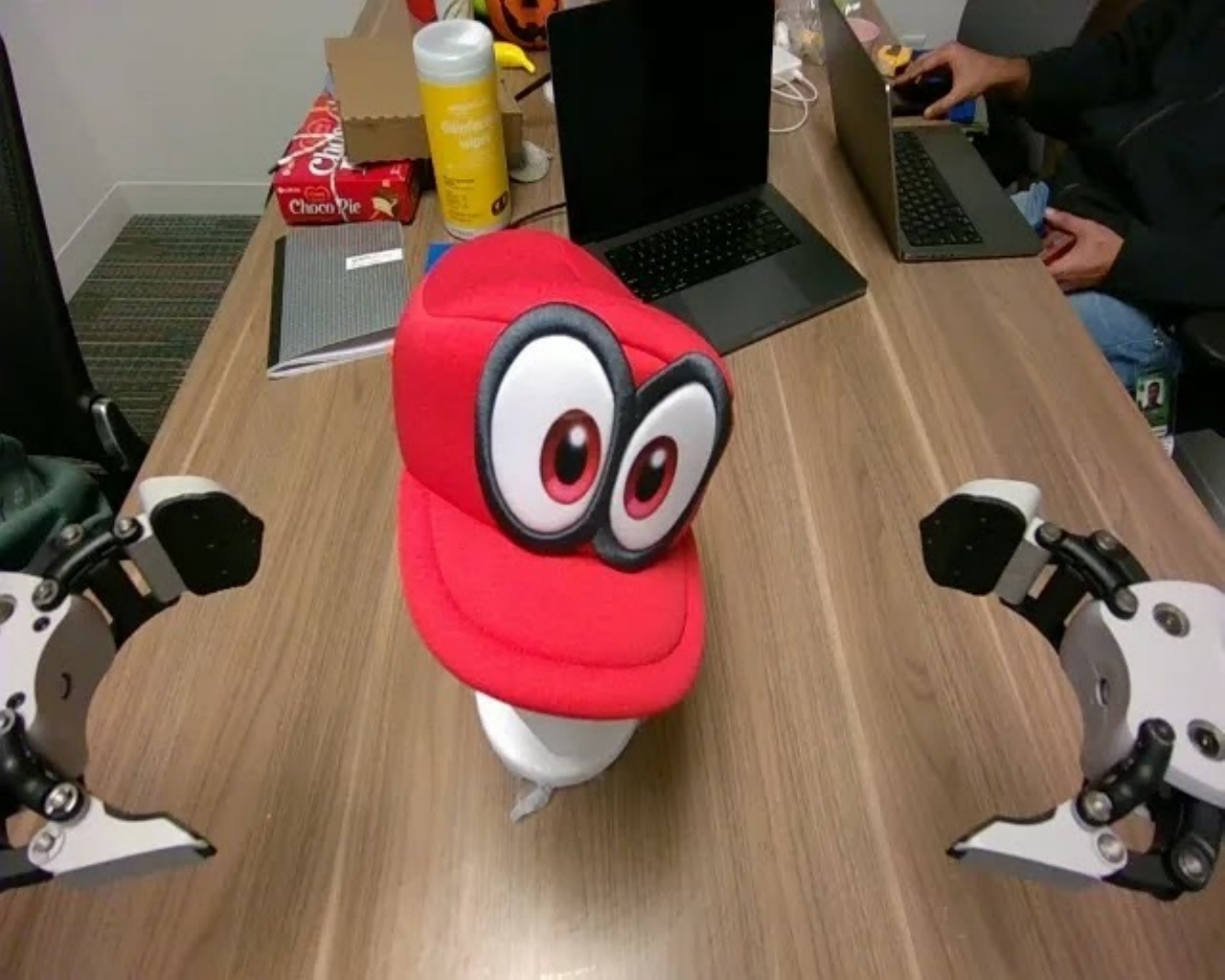} & \centering\arraybackslash\scriptsize{The robot reaches its left arm to grasp the hat and lifts it to remove it from the mannequin's head.} \\
\midrule
3 & \centering\textbf{Draw Circle} & \includegraphics[width=3.2cm]{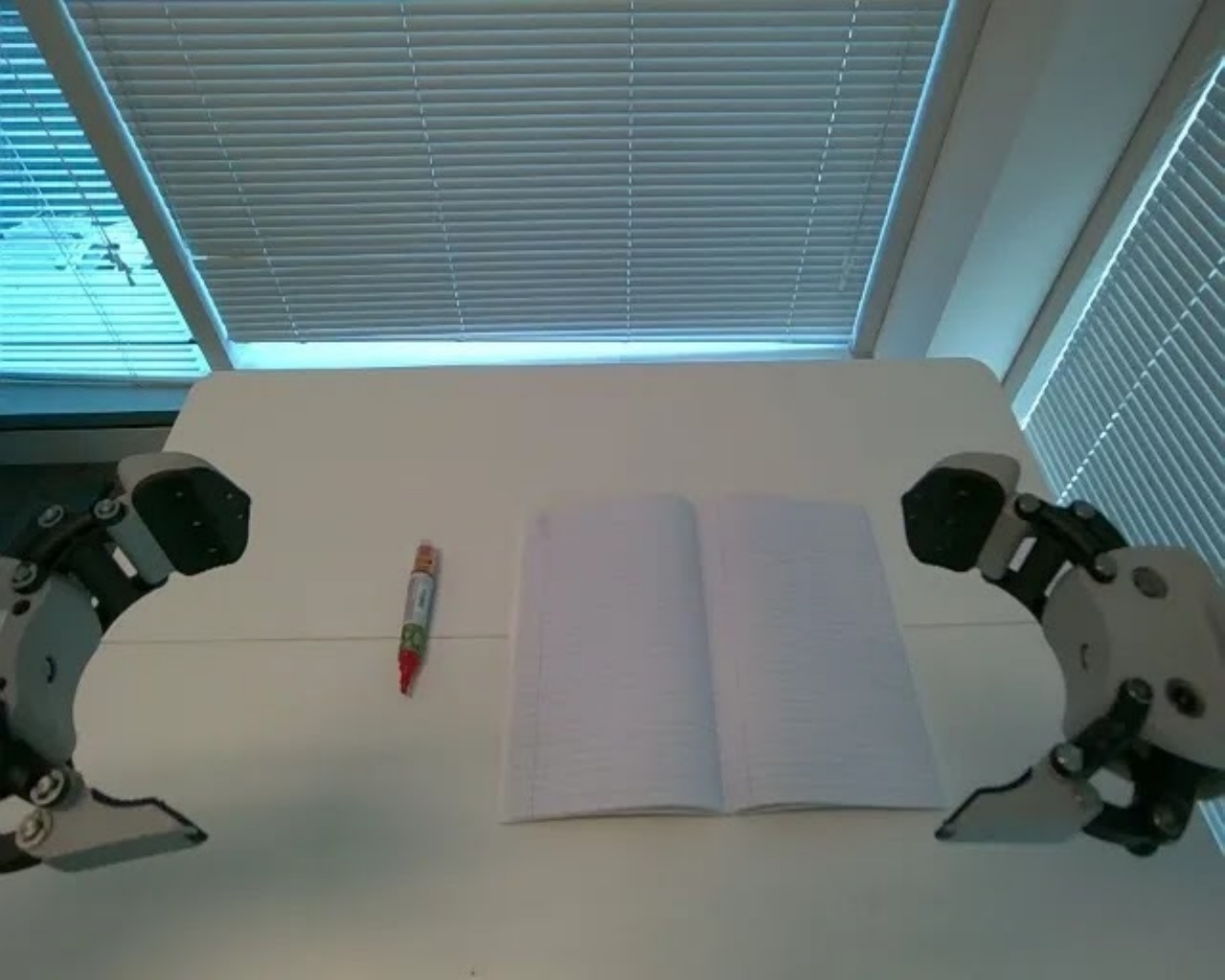} & \centering\scriptsize{The robot reaches its left arm to pick up the red marker and the left arm moves the marker to draw a circle on the book.} & \includegraphics[width=3.2cm]{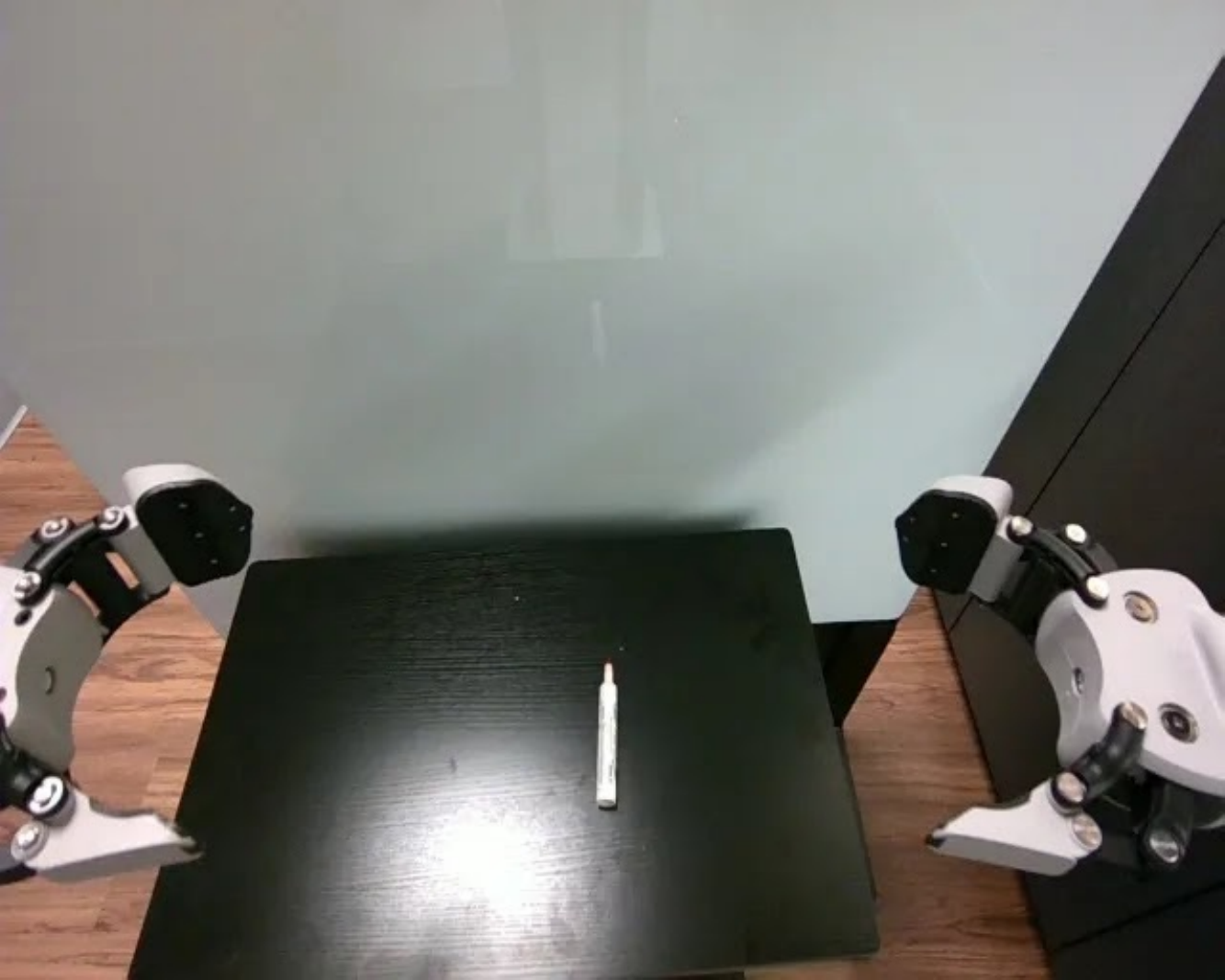} & \centering\scriptsize{The robot reaches its right arm to pick up the marker and the right arm  draws a circle on the whiteboard with the marker.} & \includegraphics[width=3.2cm]{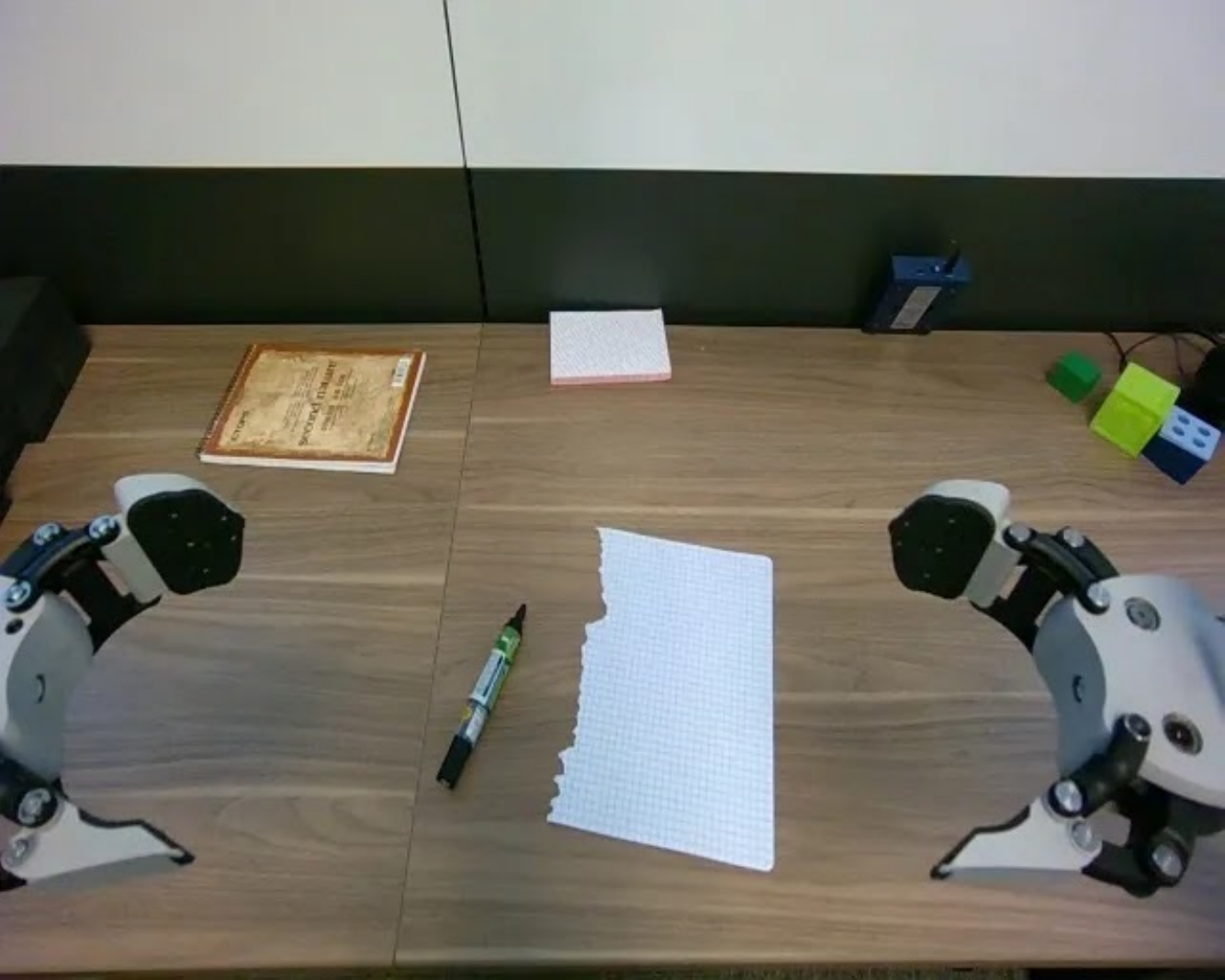} & \centering\scriptsize{The robot reaches its left arm to pick up the black marker and the left arm moves the marker to draw a circle on the paper} & \includegraphics[width=3.2cm]{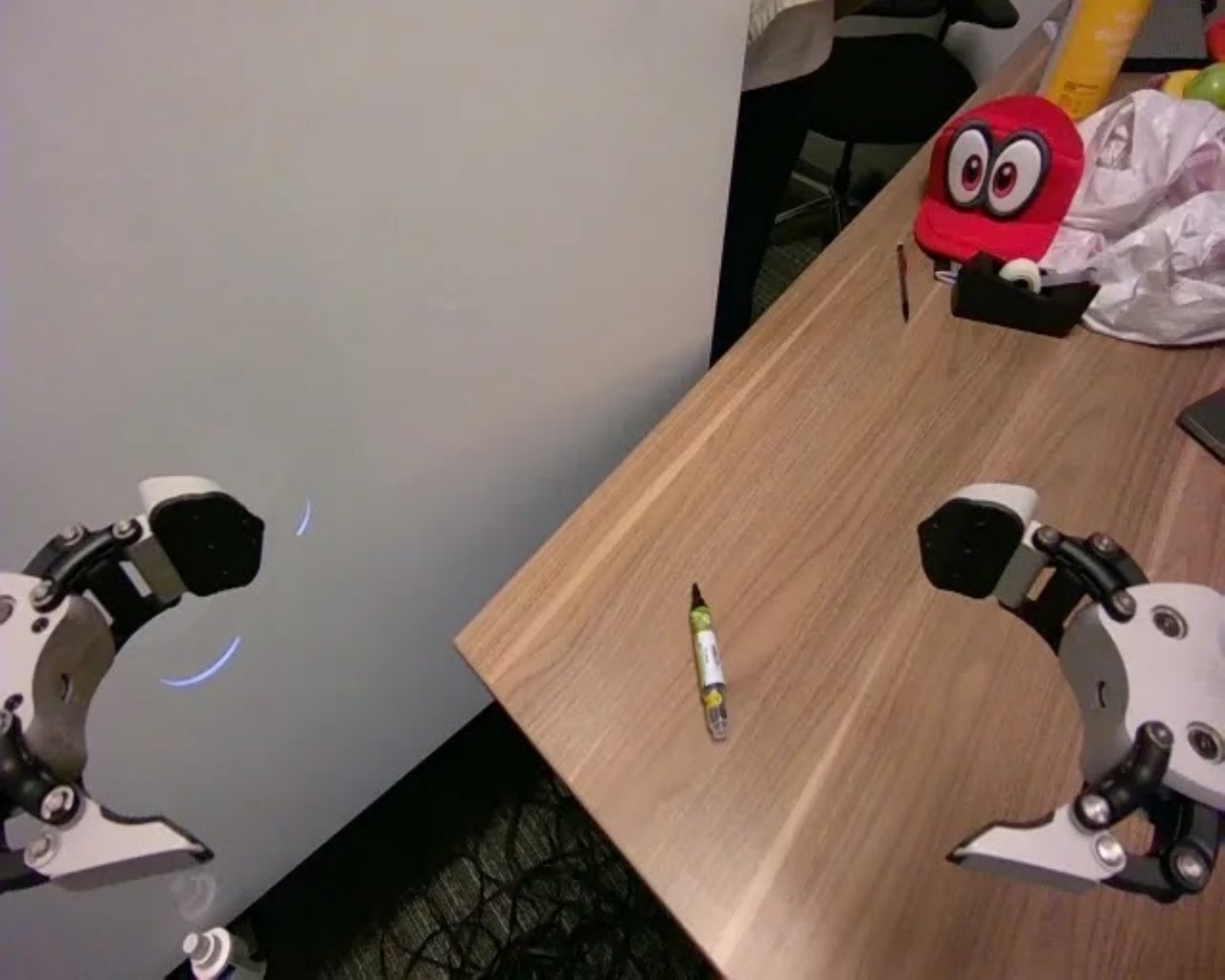} & \centering\arraybackslash\scriptsize{The robot reaches its right arm to pick up the marker and the right arm moves the marker to draw a llne on the whiteboard.} \\
\midrule
4 & \centering\textbf{Take out Straw} & \includegraphics[width=3.2cm]{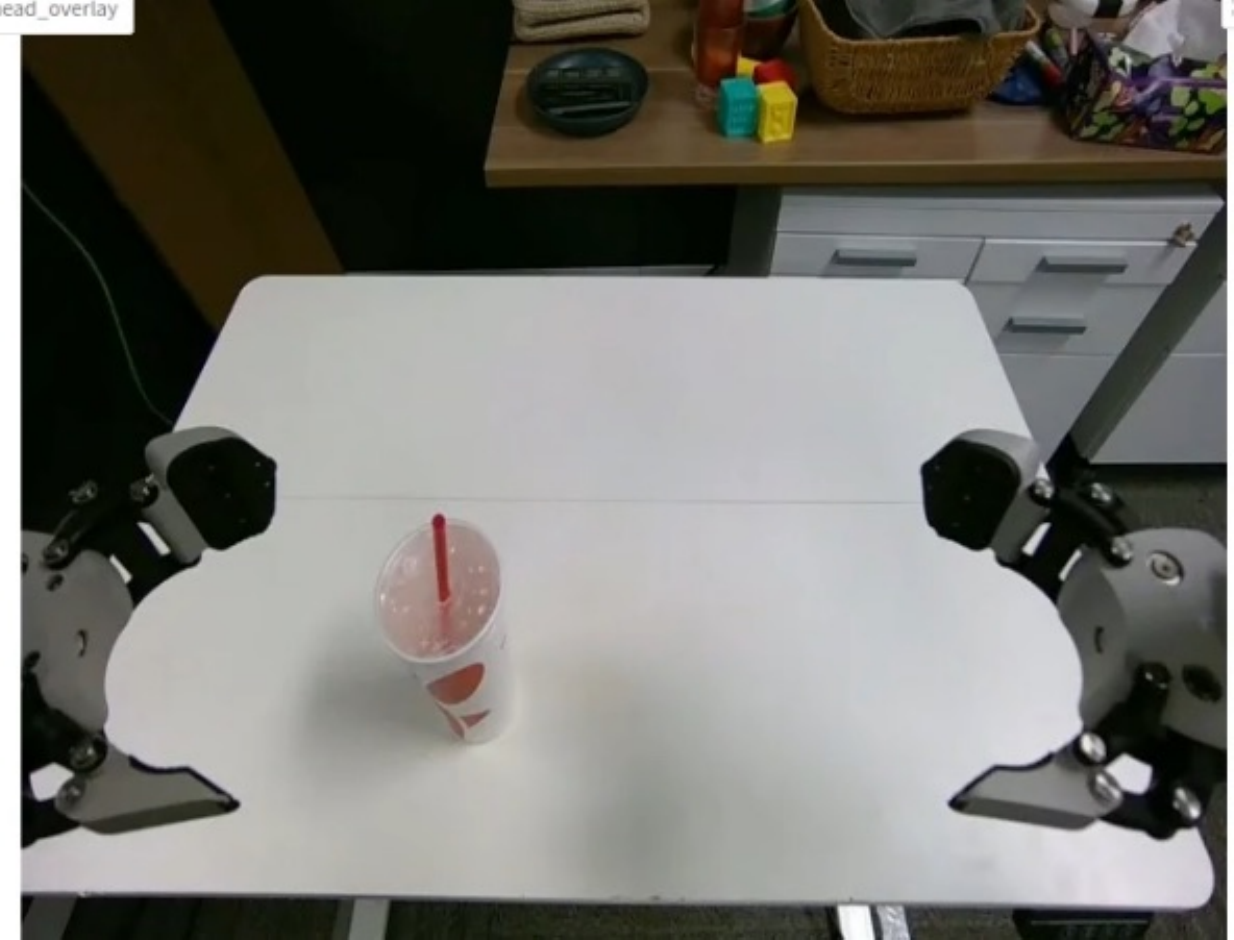} & \centering\scriptsize{The left arm holds the cup on the table. Then the right arm pulls the straw out of the cup.} & \includegraphics[width=3.2cm]{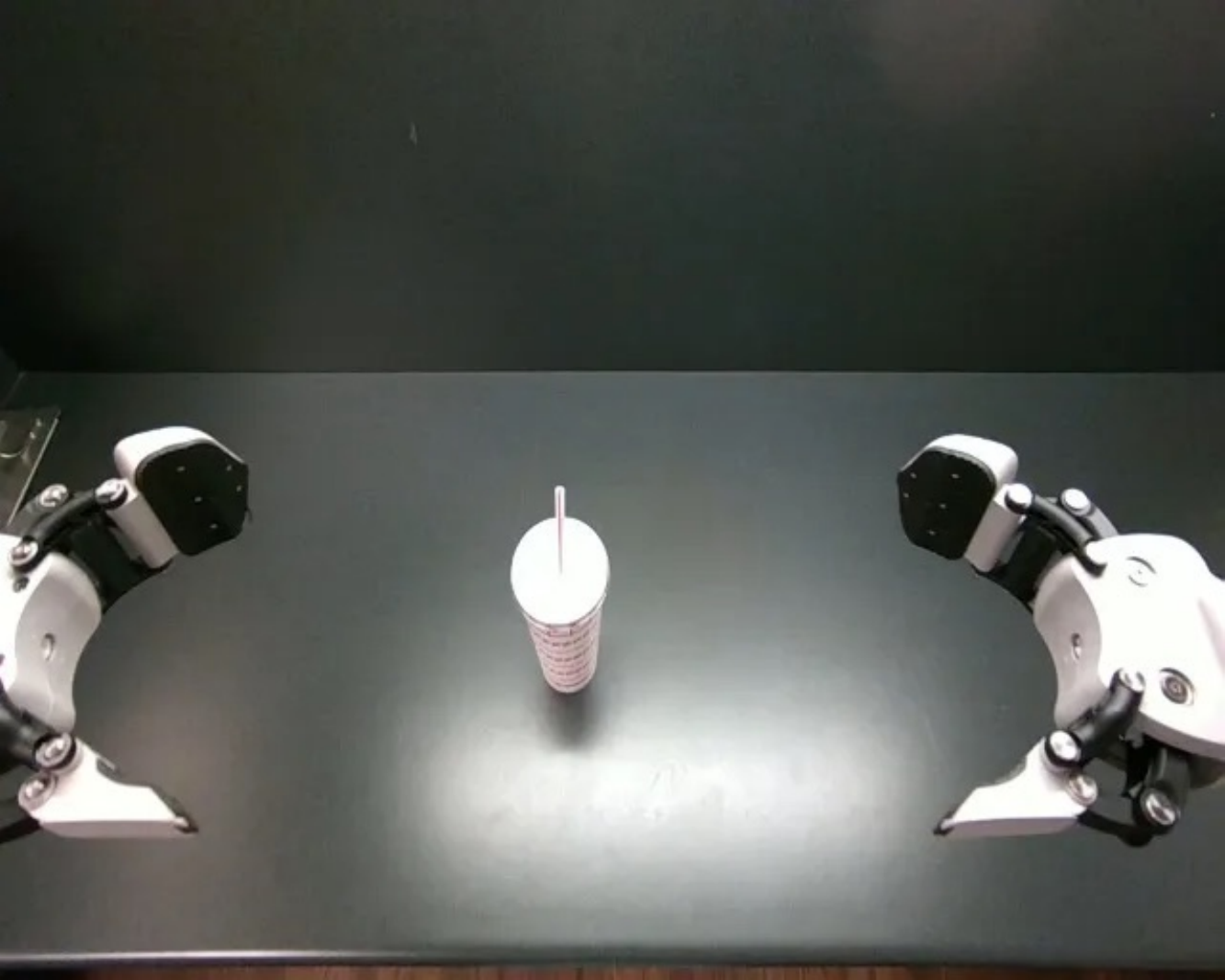} & \centering\scriptsize{The left arm holds the cup on the table. Then the right arm pulls the straw out of the cup.} & \includegraphics[width=3.2cm]{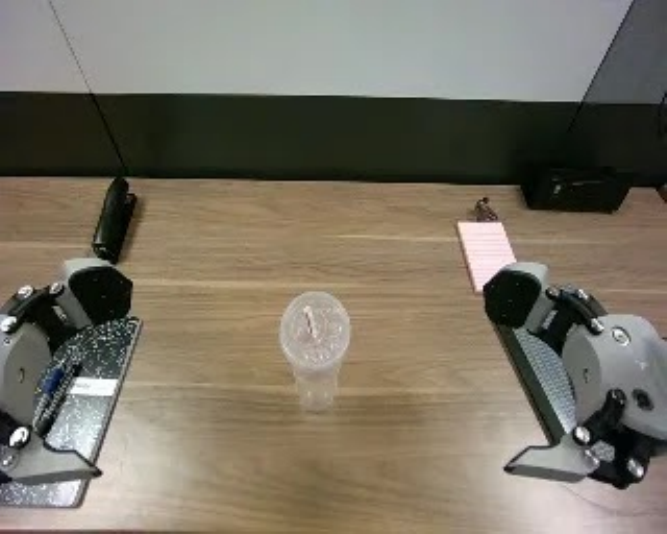} & \centering\scriptsize{The left arm holds the cup on the table. Then the right arm pulls the straw out of the cup.} & \includegraphics[width=3.2cm]{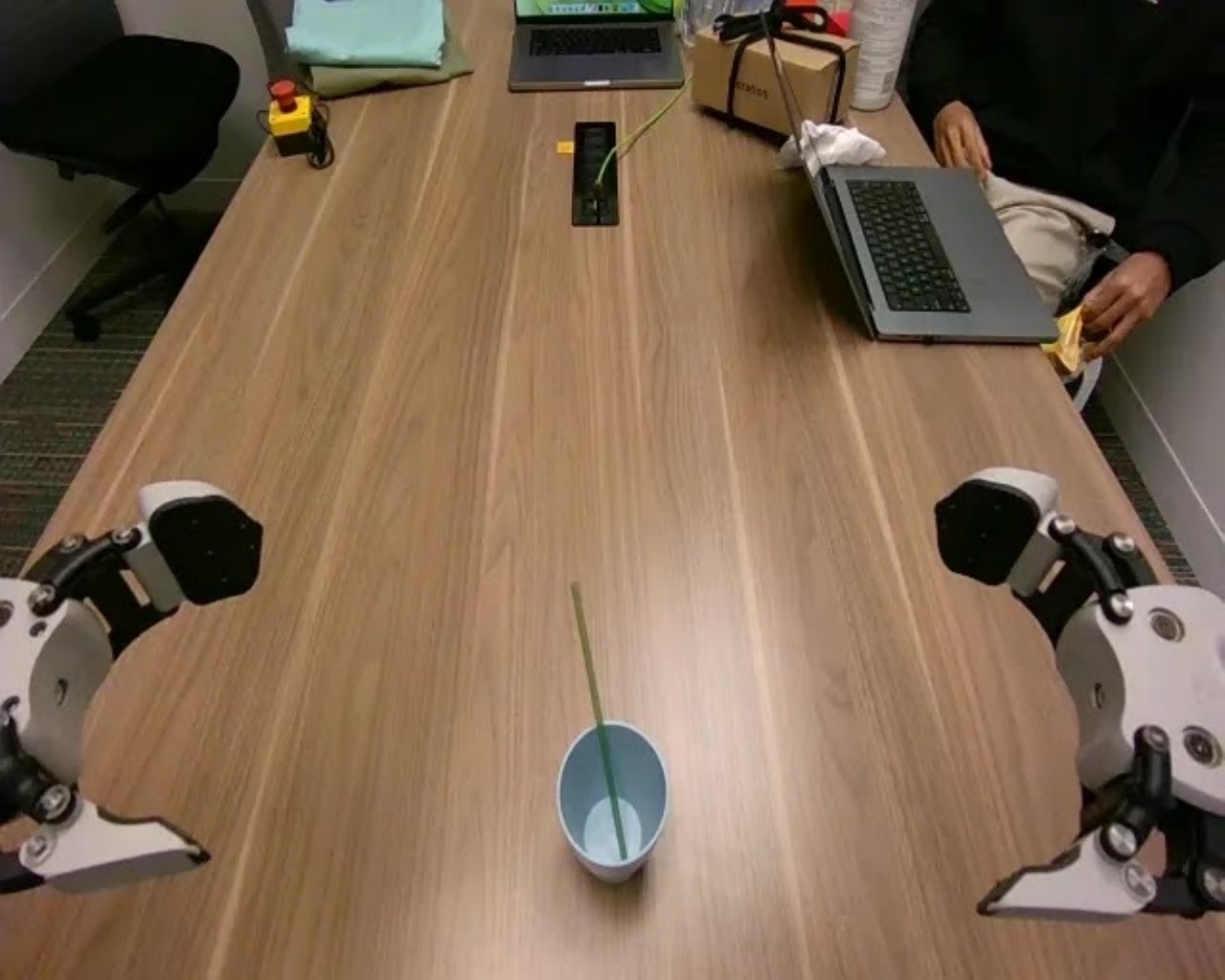} & \centering\arraybackslash\scriptsize{The right arm holds the cup on the table.Then the left arm pulls the straw out of the cup.} \\
\midrule
5 & \centering\textbf{Cube Stacking} & \includegraphics[width=3.2cm]{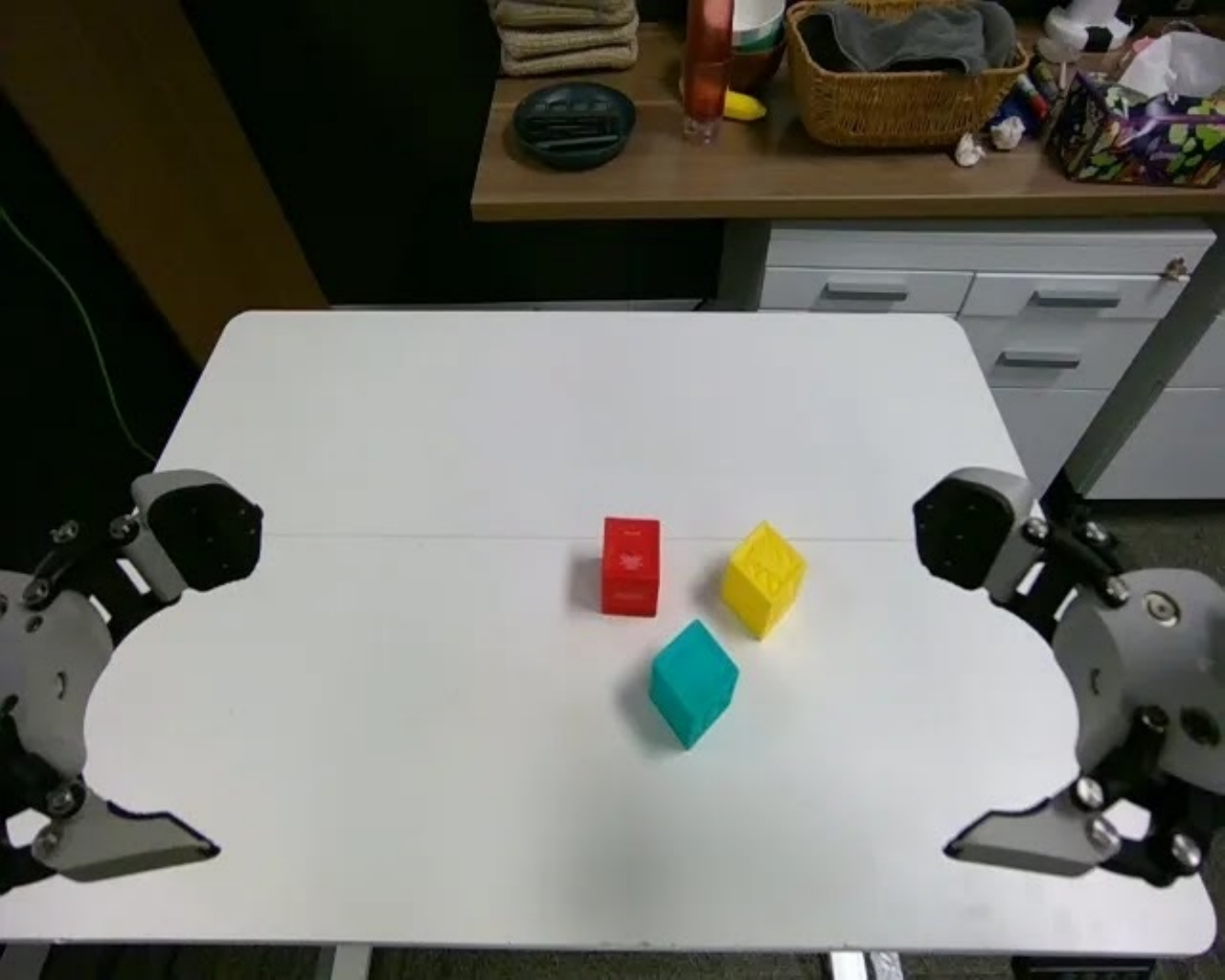} & \centering\scriptsize{The robot reaches its right arm to pick up the green cube, moves it over the red cube, and releases it to stack. It then reaches its right arm to pick up the yellow cube, moves it over the stack, and releases it onto the green cube to finish the task.} & \includegraphics[width=3.2cm]{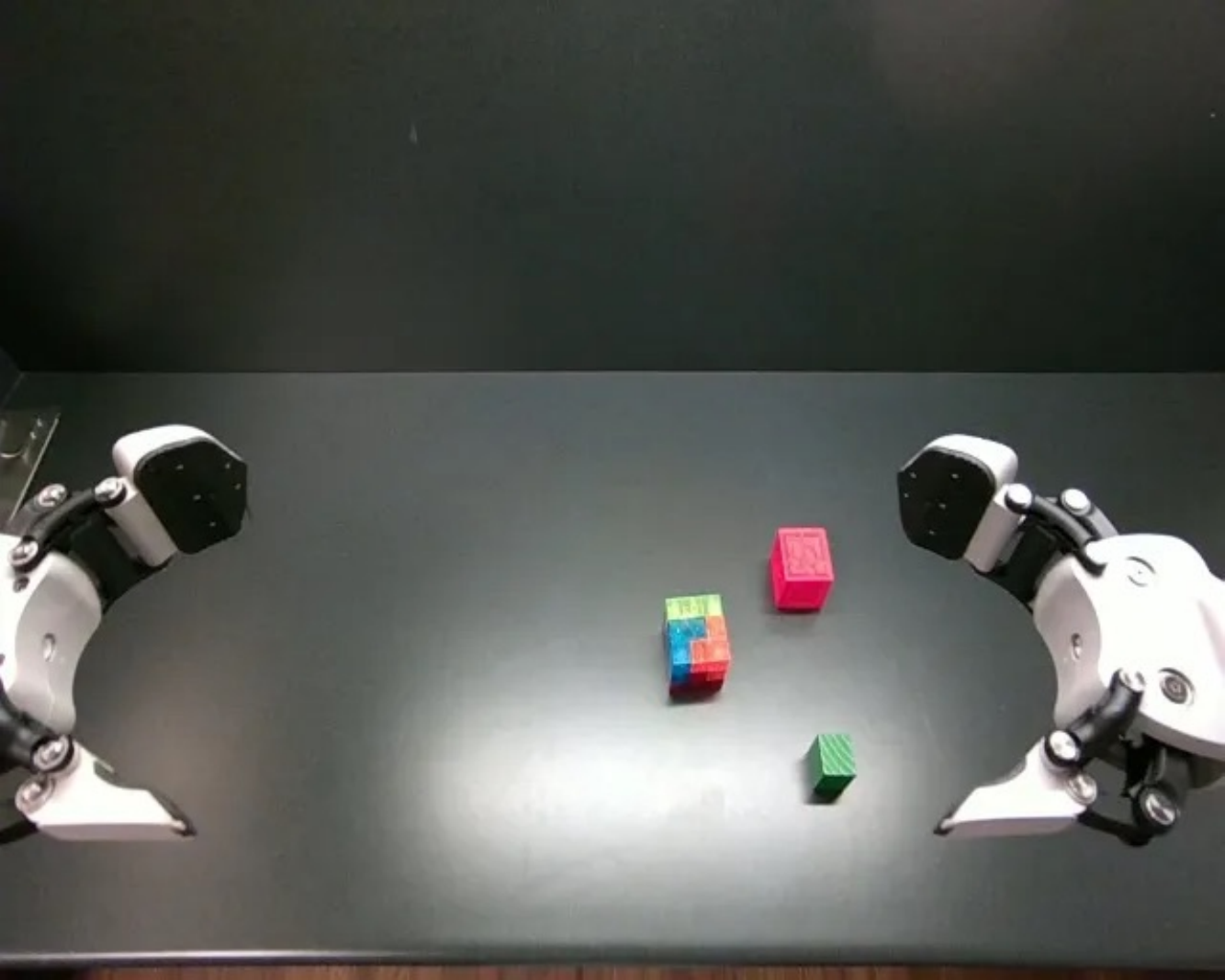} & \centering\scriptsize{The robot reaches its right arm to pick up the red cube, moves it over the colorful cube, and releases it to stack. It then reaches its right arm to pick up the green cube, moves it over the red cube, and releases it to finish the tower.} & \includegraphics[width=3.2cm]{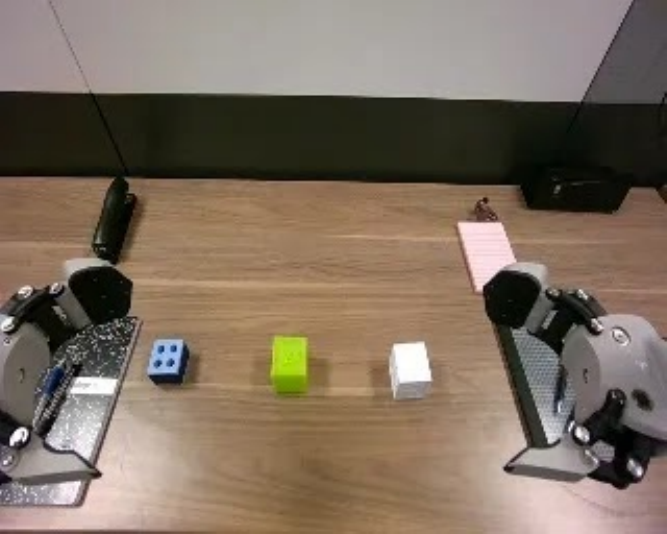} & \centering\scriptsize{The robot reaches its right arm to pick up the white cube, moves it over the green cube, and releases it to stack. It then reaches its left arm to pick up the blue cube, moves it over the stack, and releases it onto the white cube to finish the task.} & \includegraphics[width=3.2cm]{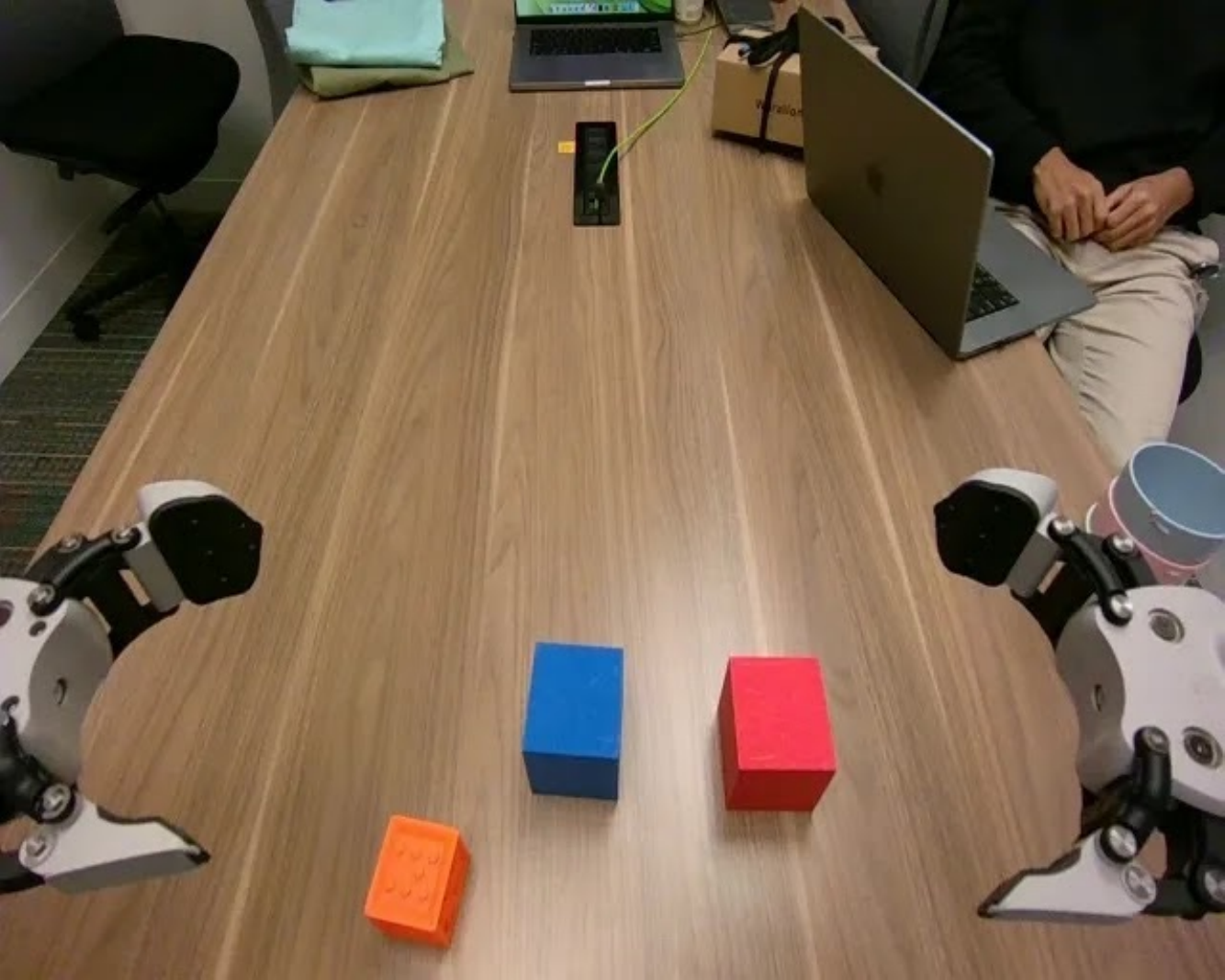} & \centering\arraybackslash\scriptsize{The robot reaches its right arm to pick up the red cube, moves it over the blue cube, and releases it to begin the stack. It then reaches its left arm to pick up the orange cube, moves it over the stack, and releases it onto the red cube to complete the three-tier structure.} \\
\midrule
6 & \centering\textbf{Painting} & \includegraphics[width=3.2cm]{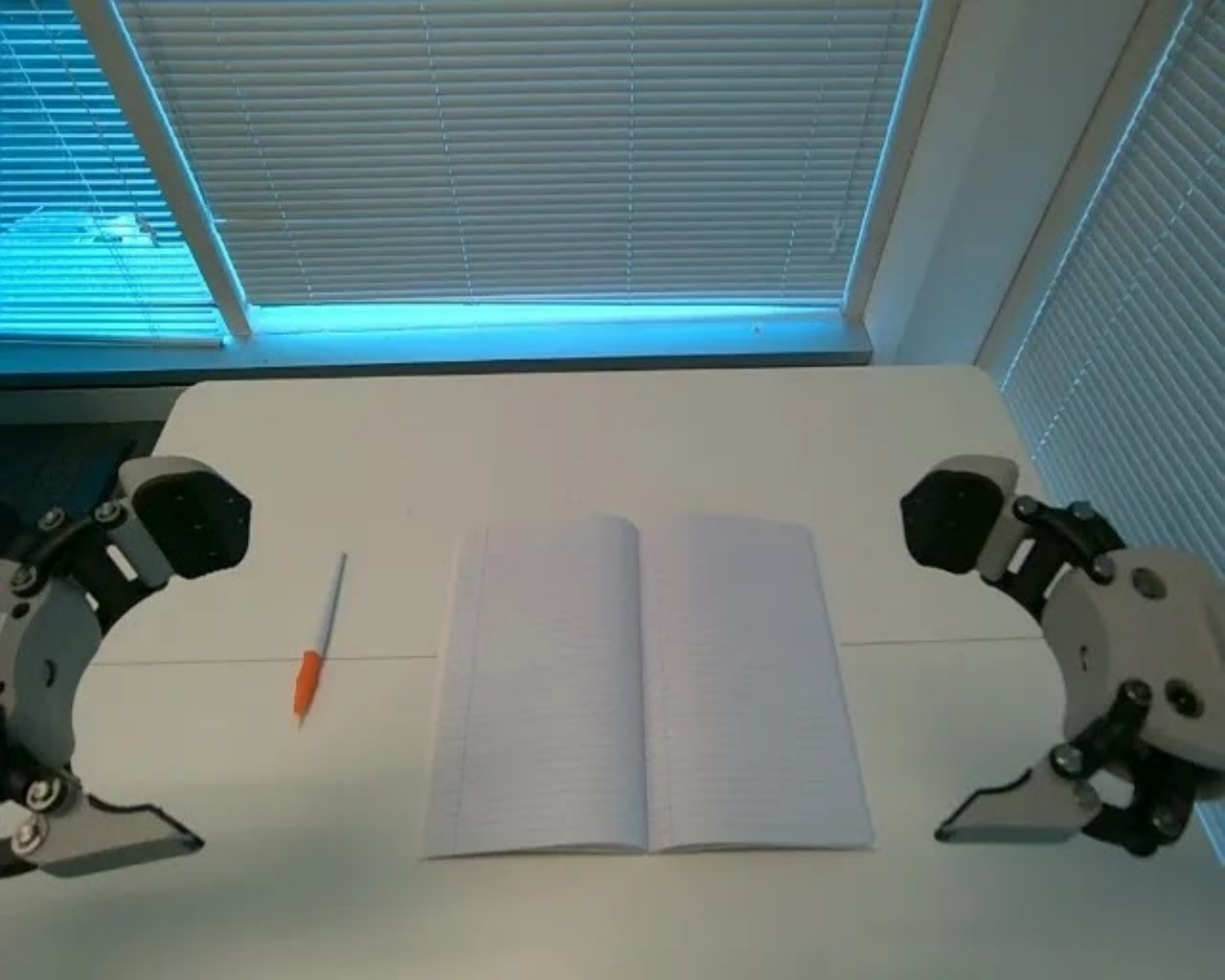} & \centering\scriptsize{The left arm grabs the brush. Then left arm paints with the brush on the notebook} & \includegraphics[width=3.2cm]{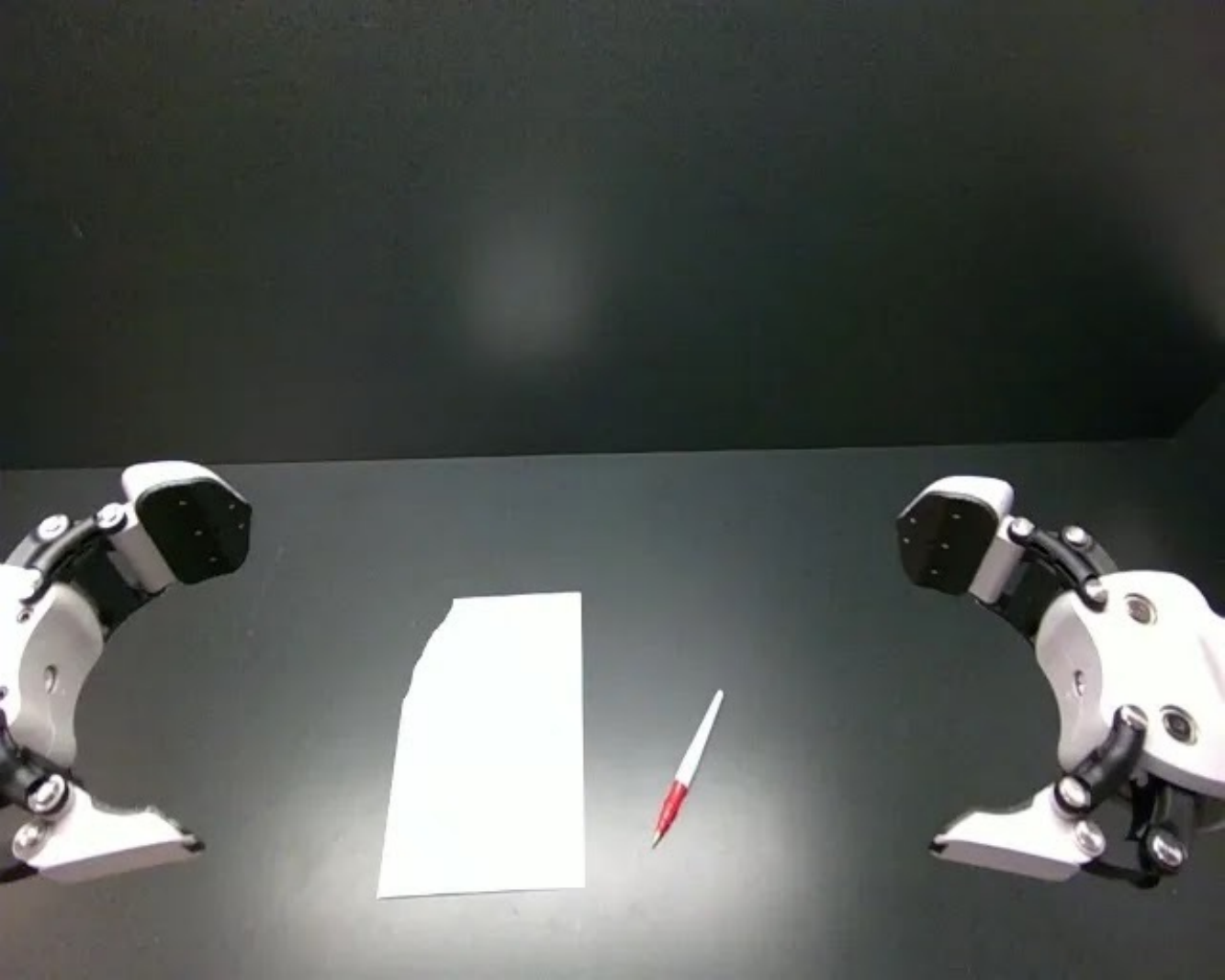} & \centering\scriptsize{The right arm grabs the brush. Then right arm paints with the brush on the paper} & \includegraphics[width=3.2cm]{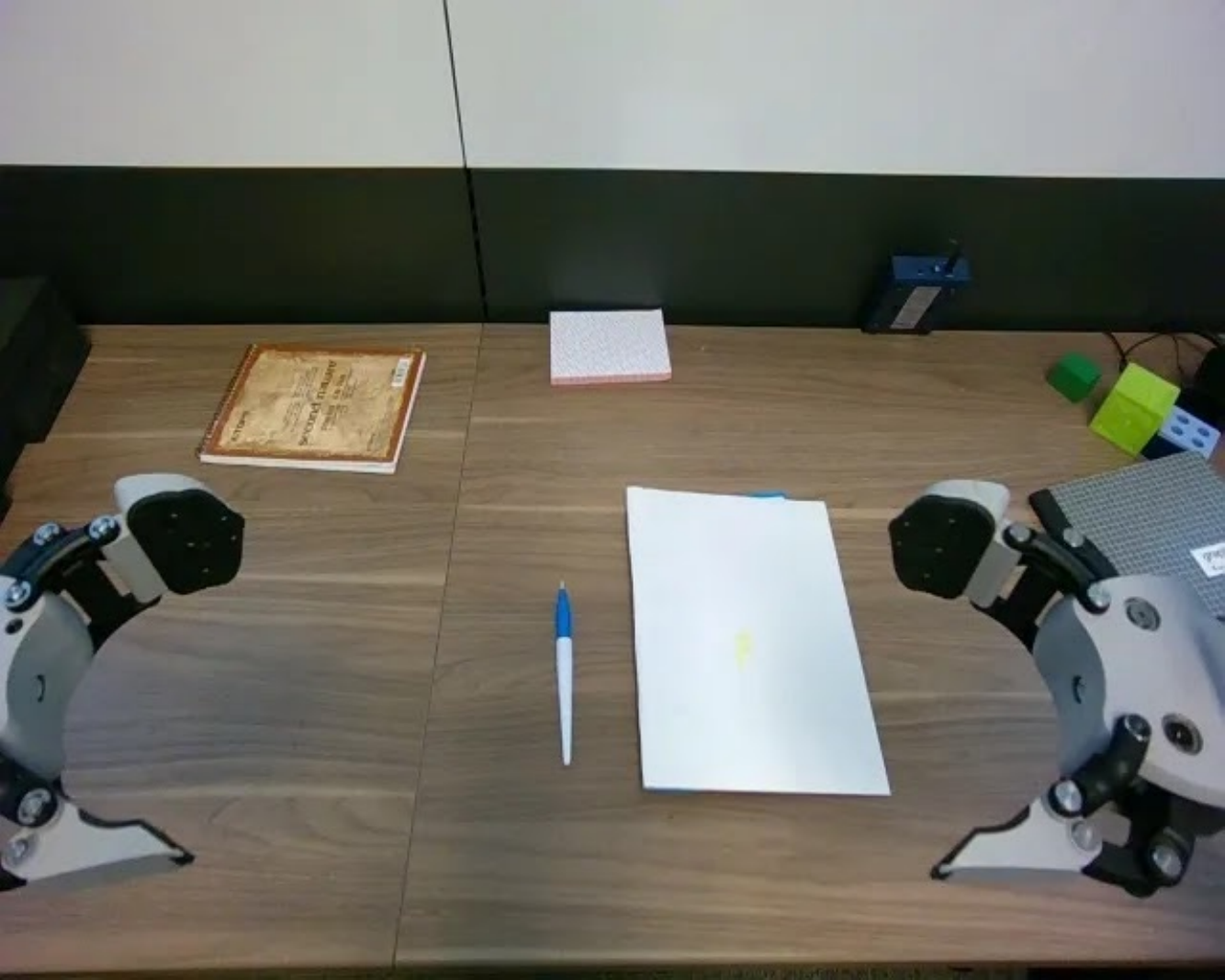} & \centering\scriptsize{The left arm grabs the brush. Then left arm paints with the brush on the notebook} & \includegraphics[width=3.2cm]{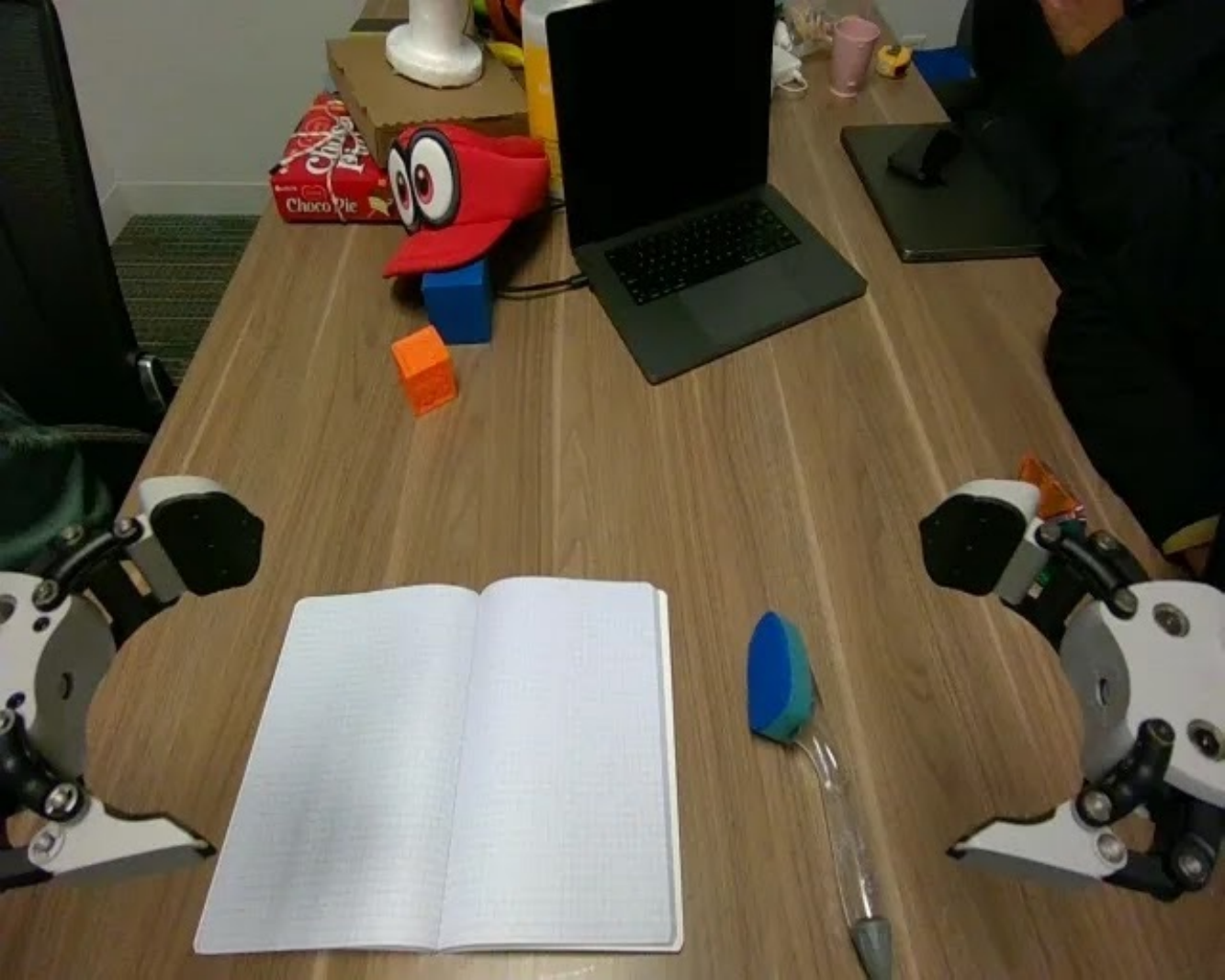} & \centering\arraybackslash\scriptsize{The right arm grabs the brush. Then right arm paints with the brush on the notebook} \\
\midrule
7 & \centering\textbf{Ironing} & \includegraphics[width=3.2cm]{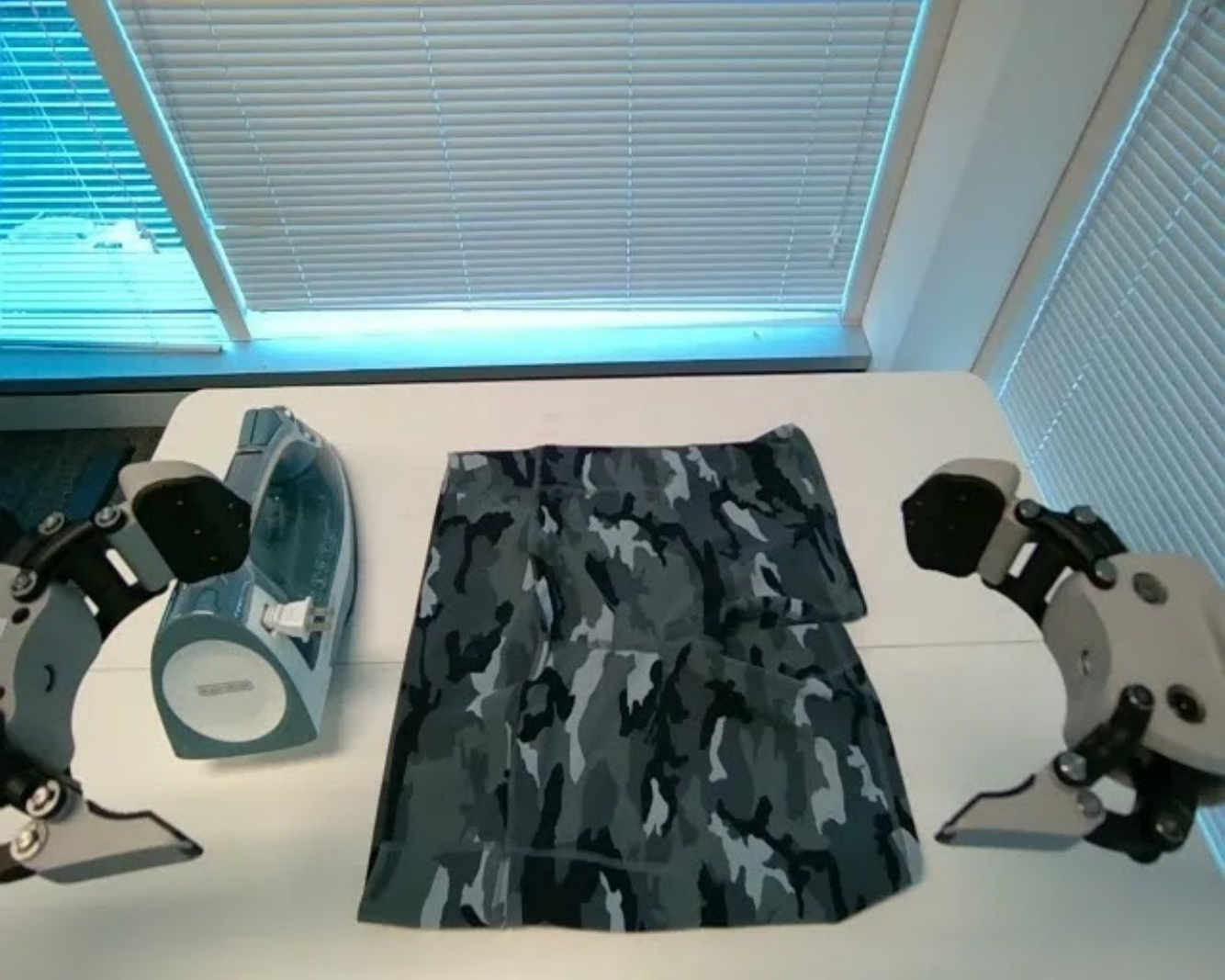} & \centering\scriptsize{The robot reaches its left arm to grasp the iron and moves it across the shorts to iron it.} & \includegraphics[width=3.2cm]{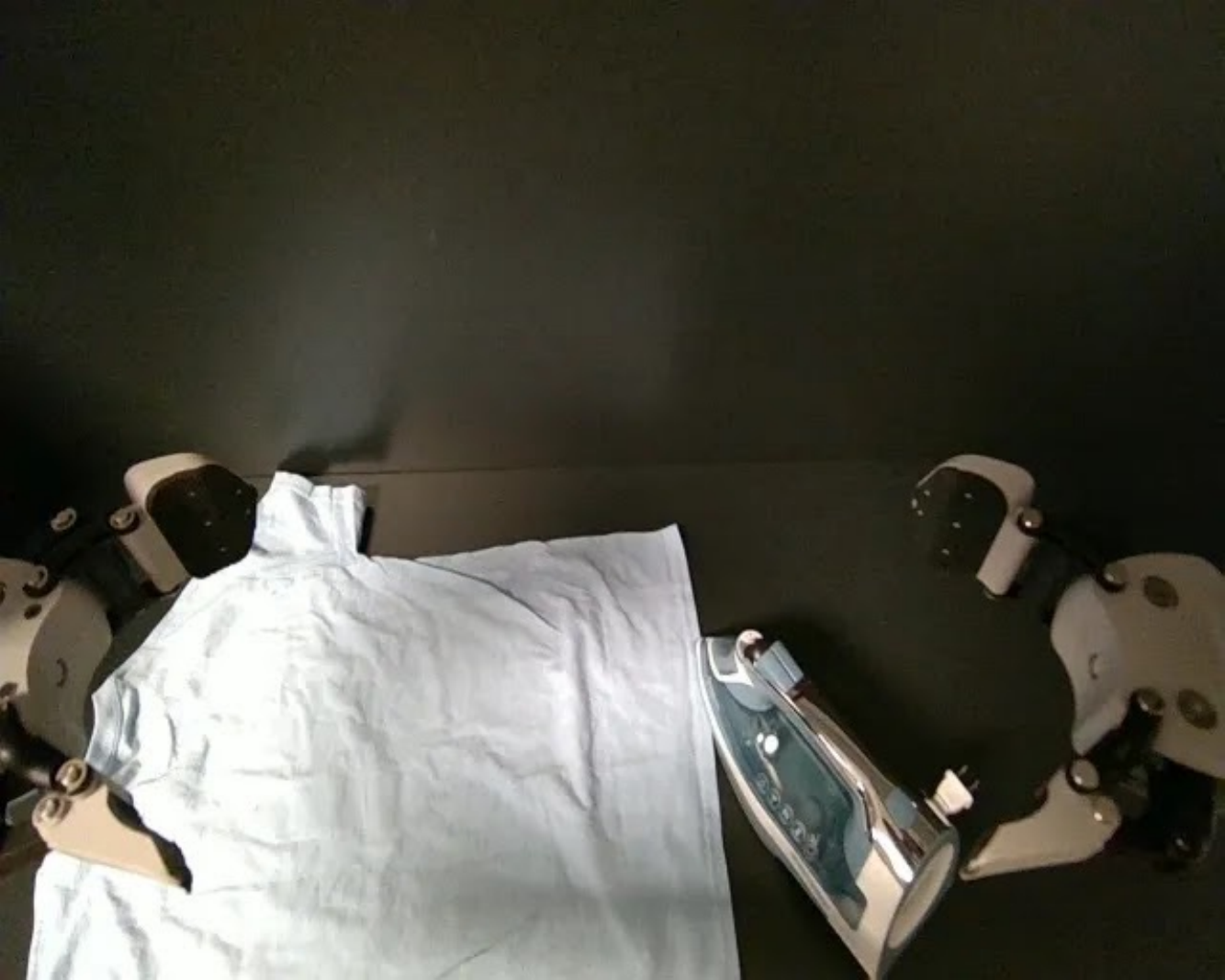} & \centering\scriptsize{The robot reaches its right arm to grasp the iron and moves it across the shirt to iron it.} & \includegraphics[width=3.2cm]{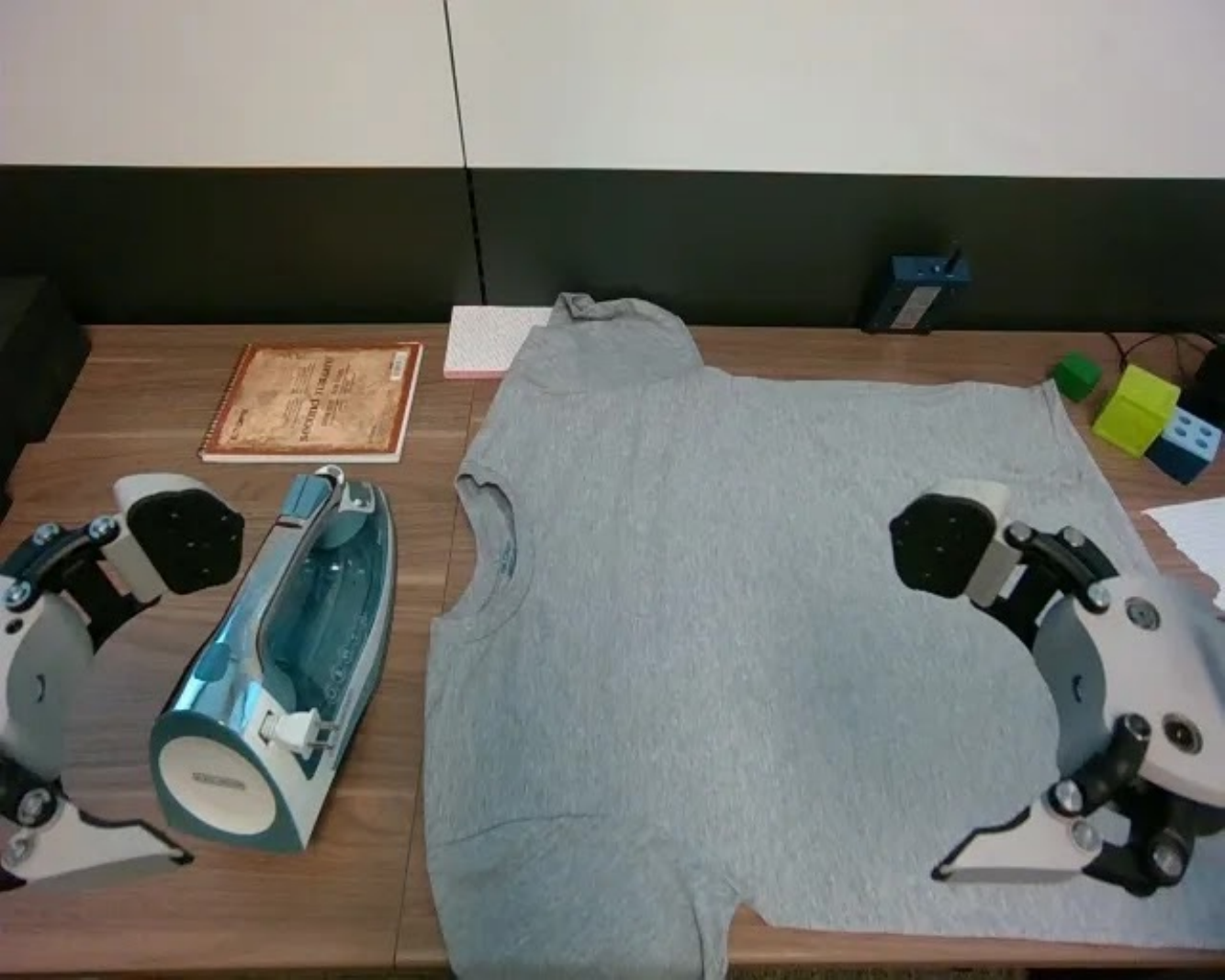} & \centering\scriptsize{The robot reaches its left arm to grasp the iron and moves it across the shirt to iron it.} & \includegraphics[width=3.2cm]{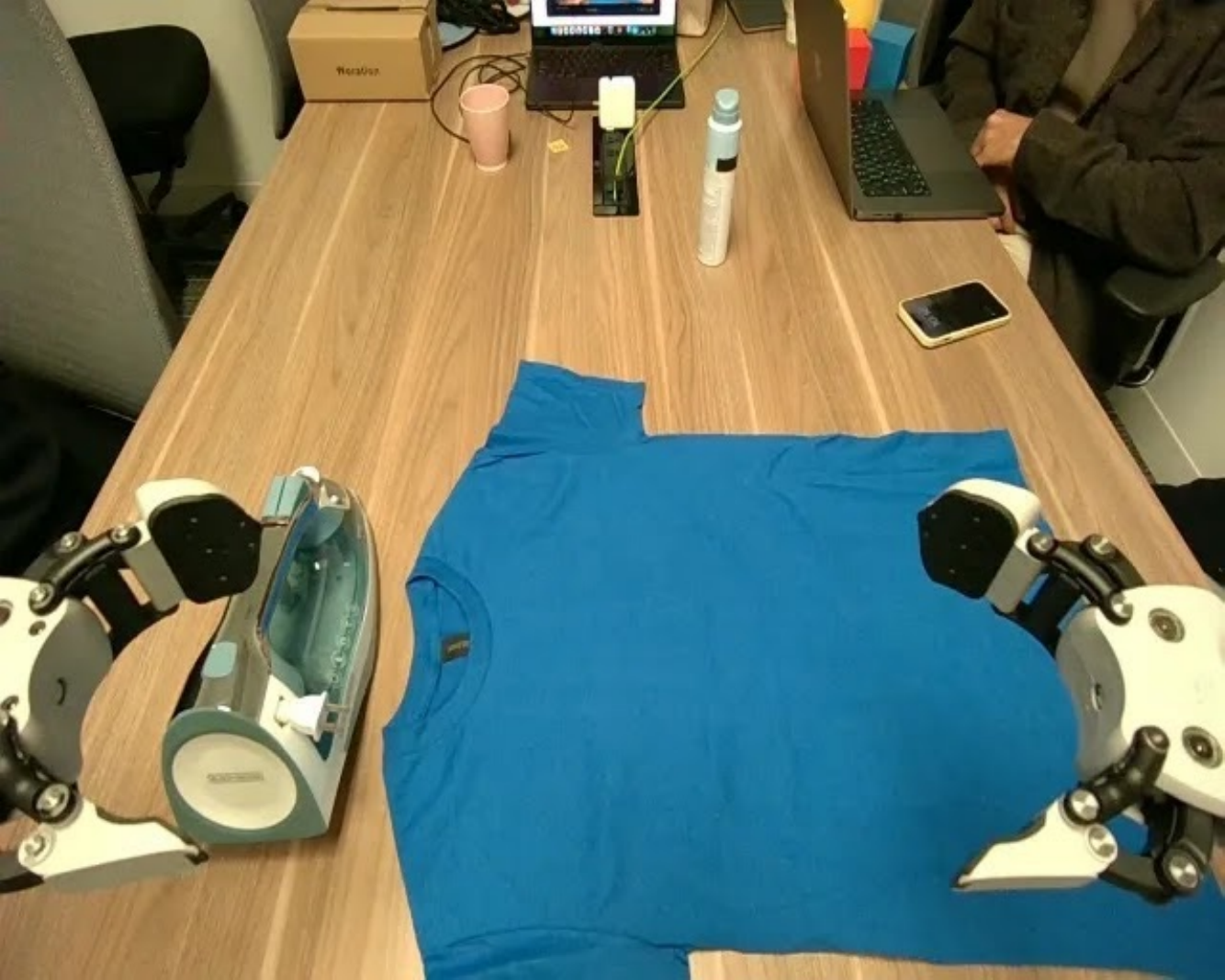} & \centering\arraybackslash\scriptsize{The robot reaches its left arm to grasp the iron and moves it across the shirt to iron it.} \\
\midrule
8 & \centering\textbf{Shake Hands} & \includegraphics[width=3.2cm]{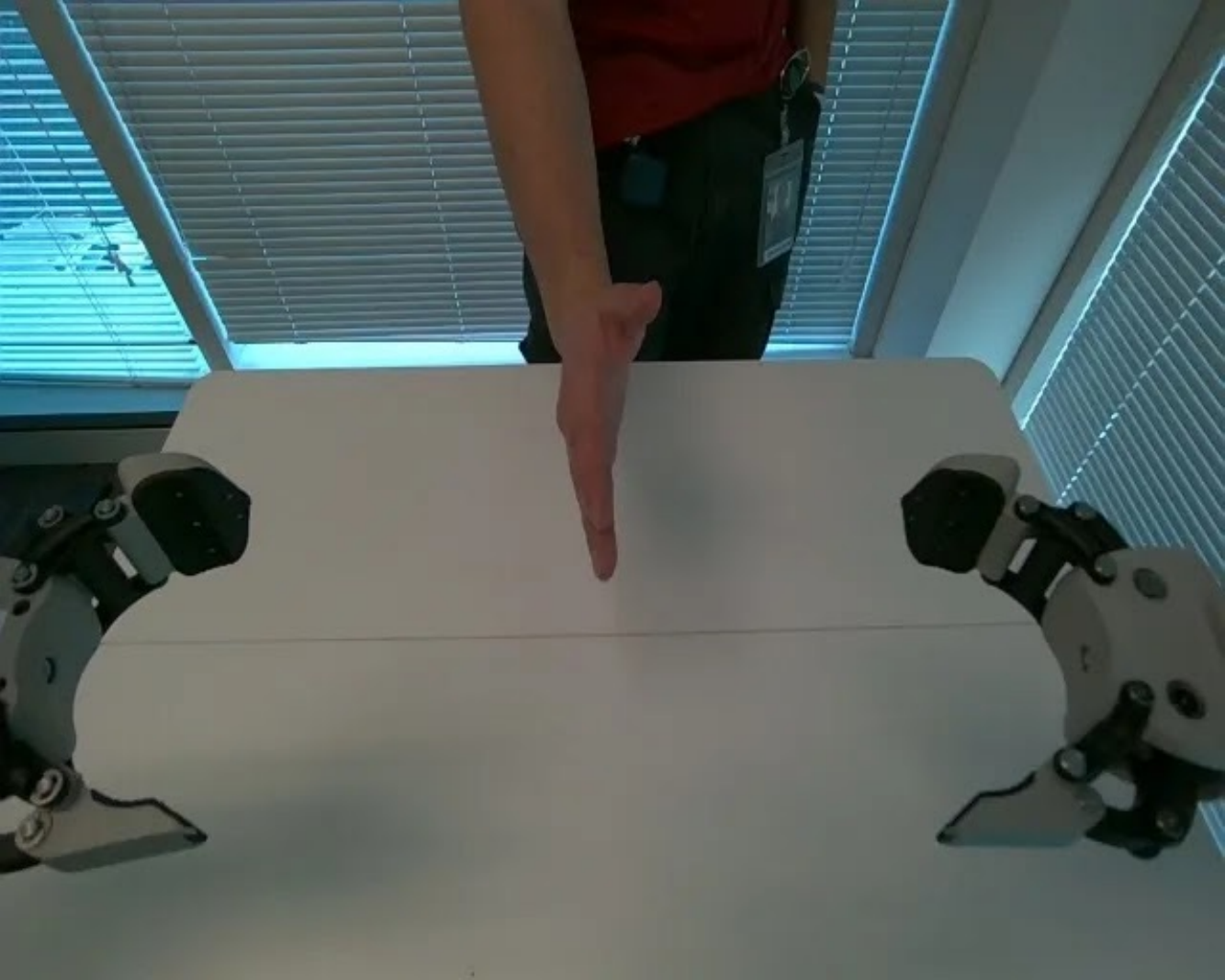} & \centering\scriptsize{The right arm of the robot grasp the human hand to shake hands. It then initiates a rhythmic up-and-down motion to perform the handshake.} & \includegraphics[width=3.2cm]{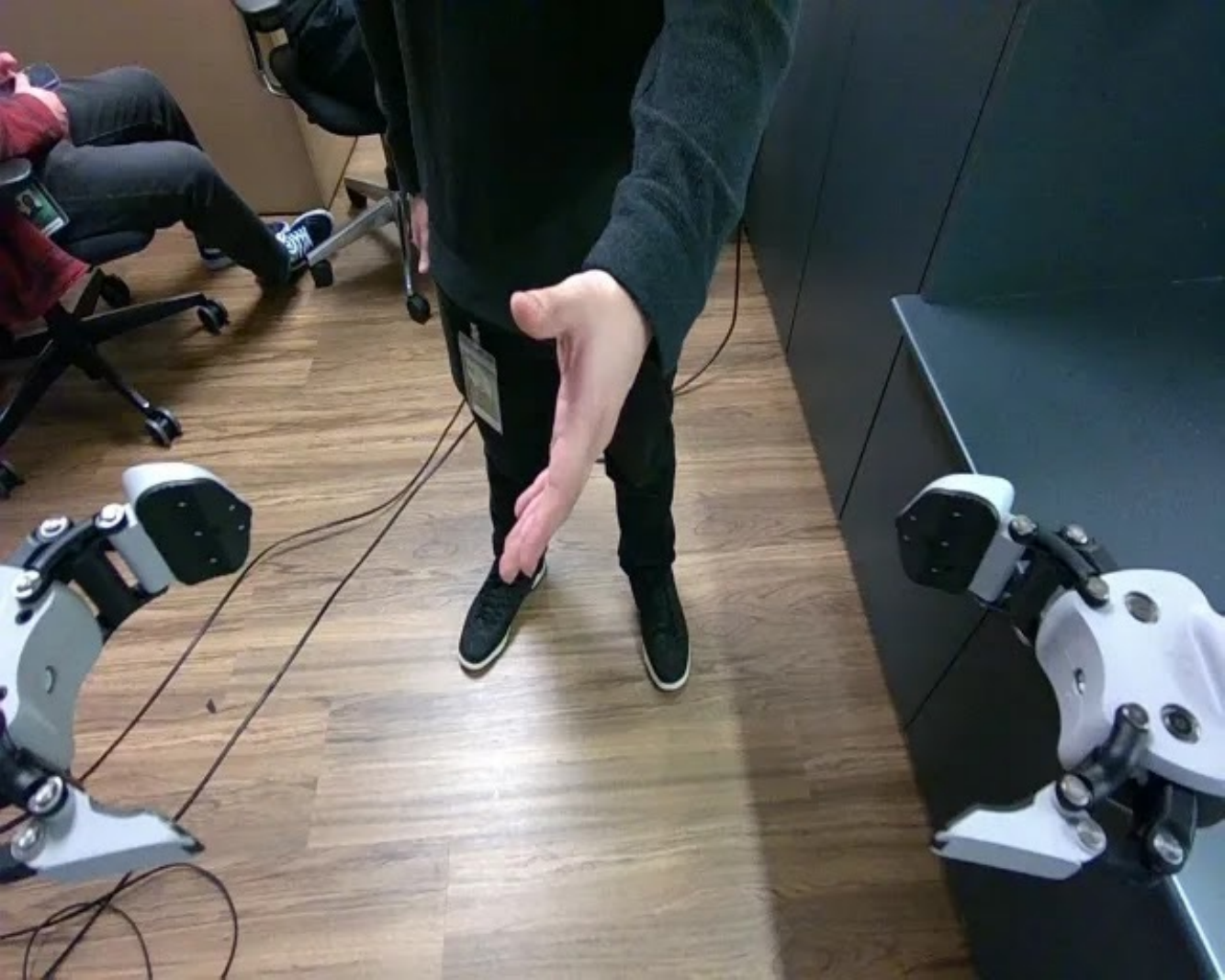} & \centering\scriptsize{The left arm shakes the hand of the human up and down} & \includegraphics[width=3.2cm]{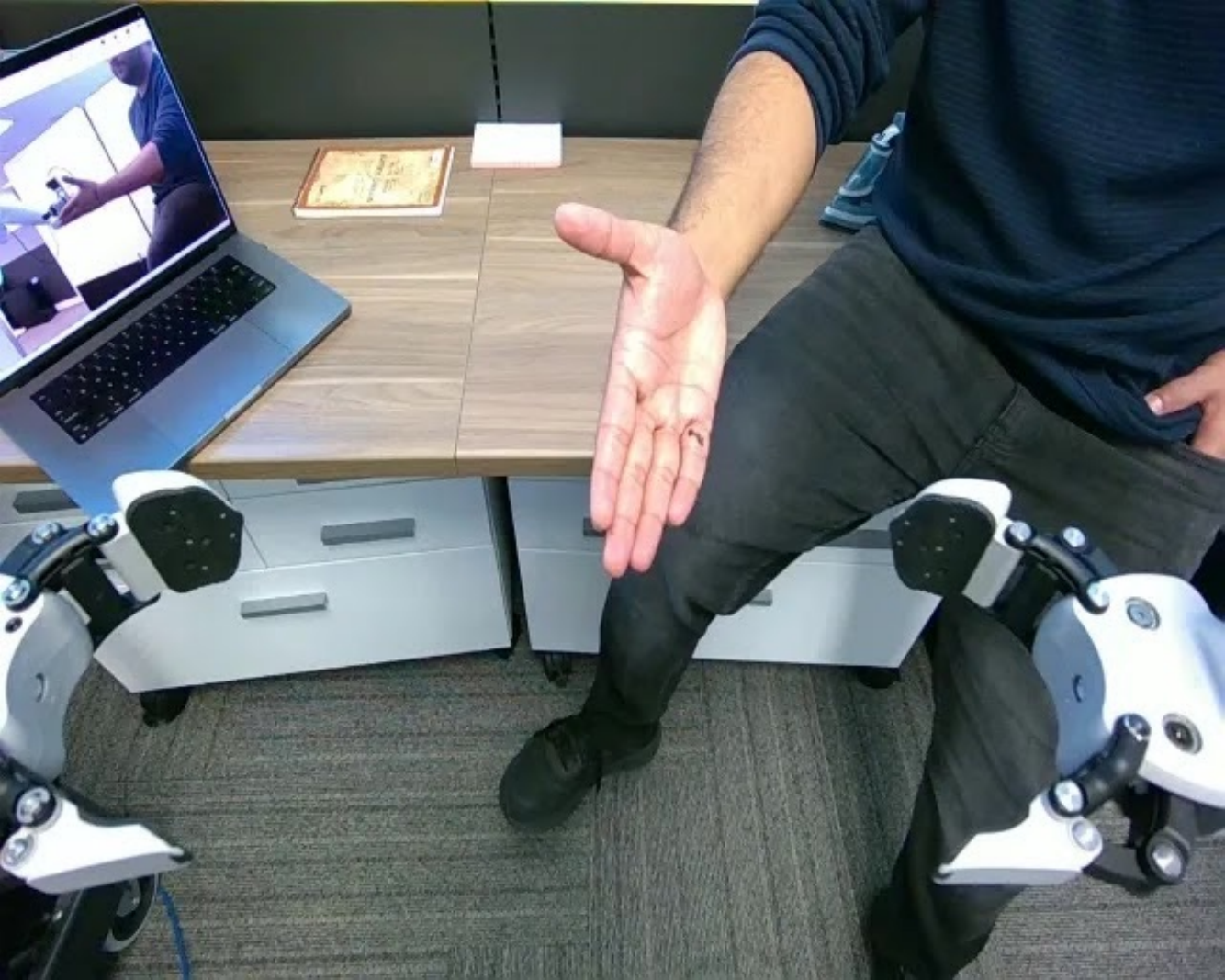} & \centering\scriptsize{The right arm shakes the hand of the human up and down} & \includegraphics[width=3.2cm]{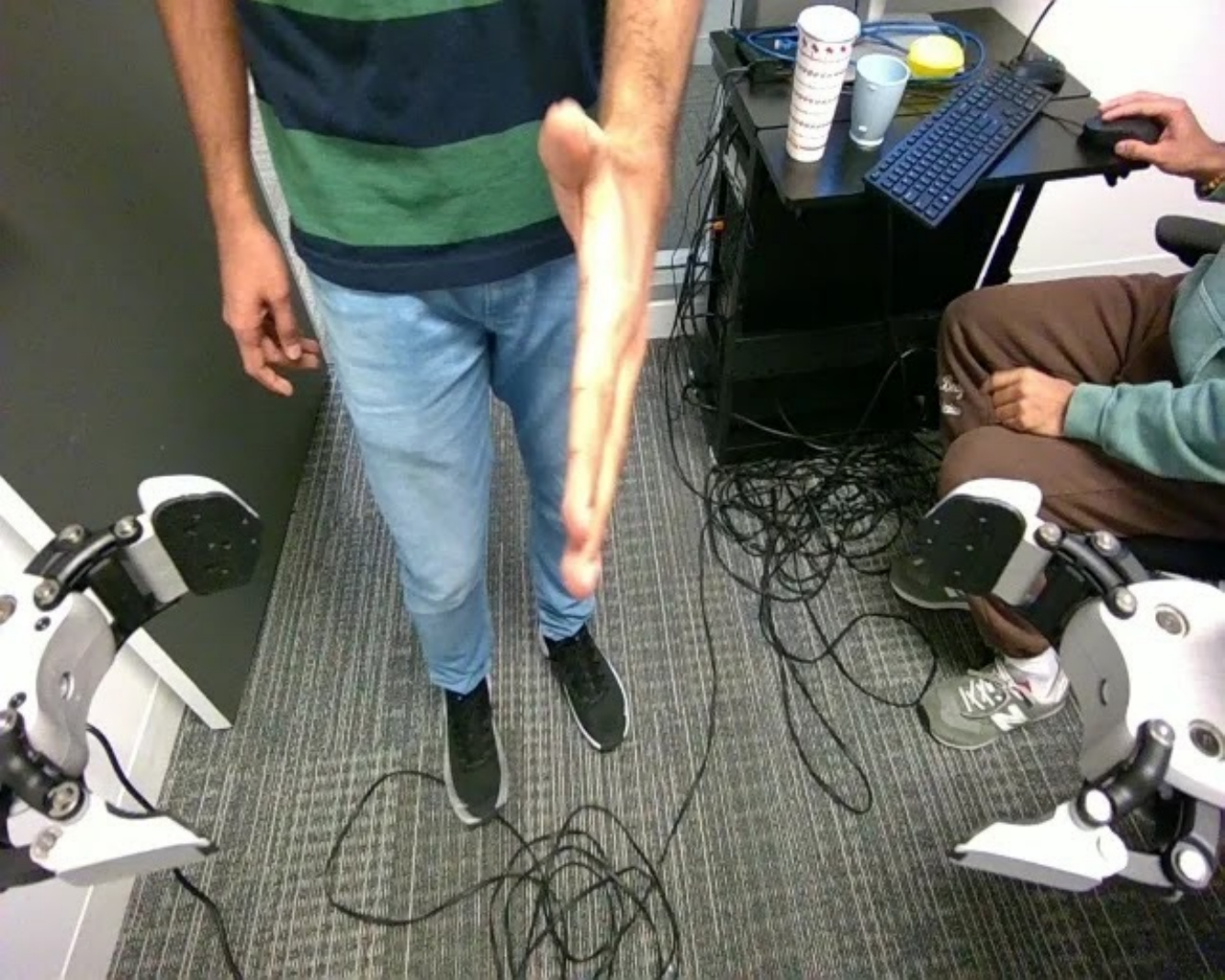} & \centering\arraybackslash\scriptsize{The left arm of the robot grasp the human hand to shake hands. It then initiates a rhythmic up-and-down motion to perform the handshake.} \\
\midrule
9 & \centering\textbf{Folding (Map)} & \includegraphics[width=3.2cm]{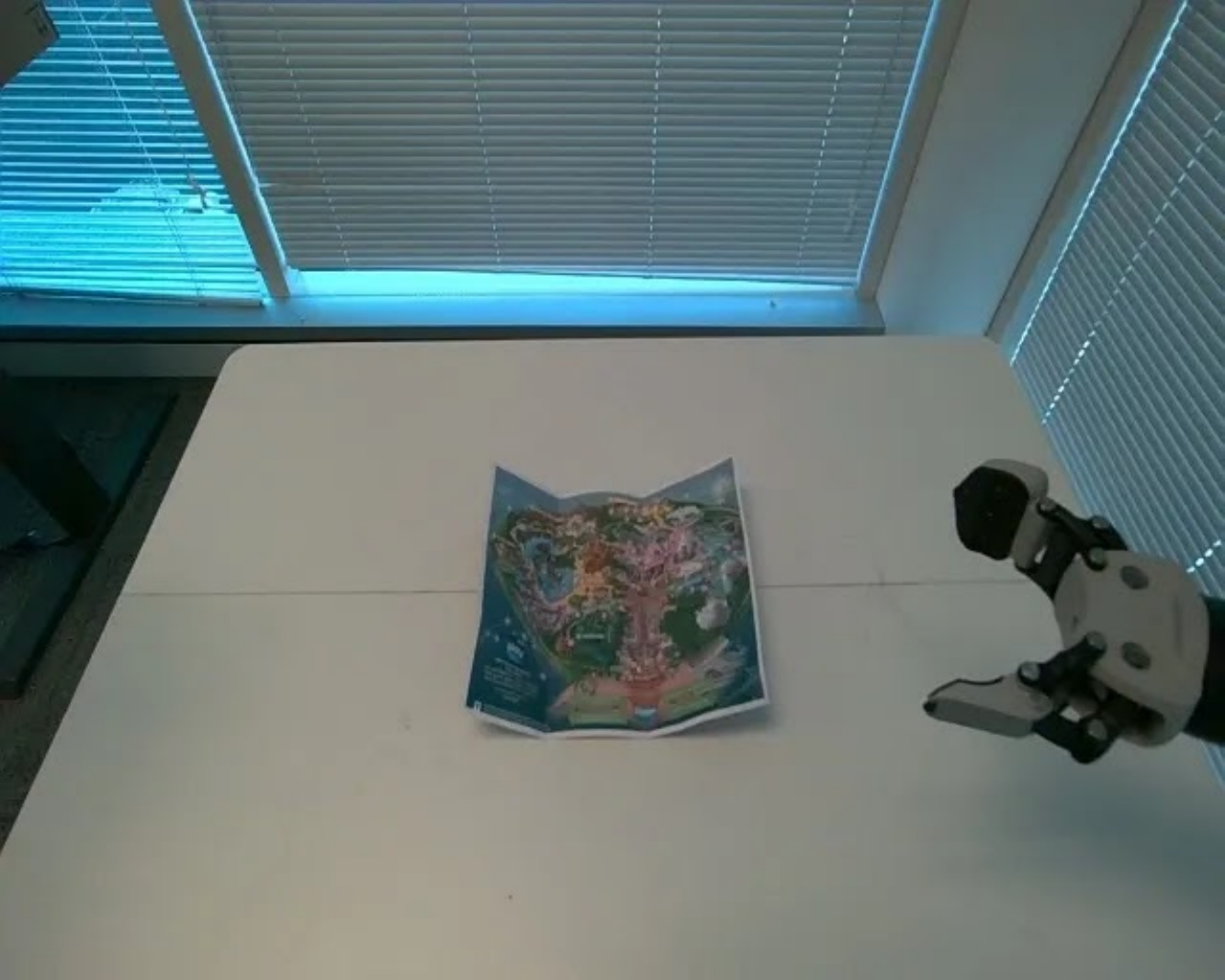} & \centering\scriptsize{The left arm grabs the left side of the map. The right arm folds the right side of the map. The left arm folds the left side of the map.} & \includegraphics[width=3.2cm]{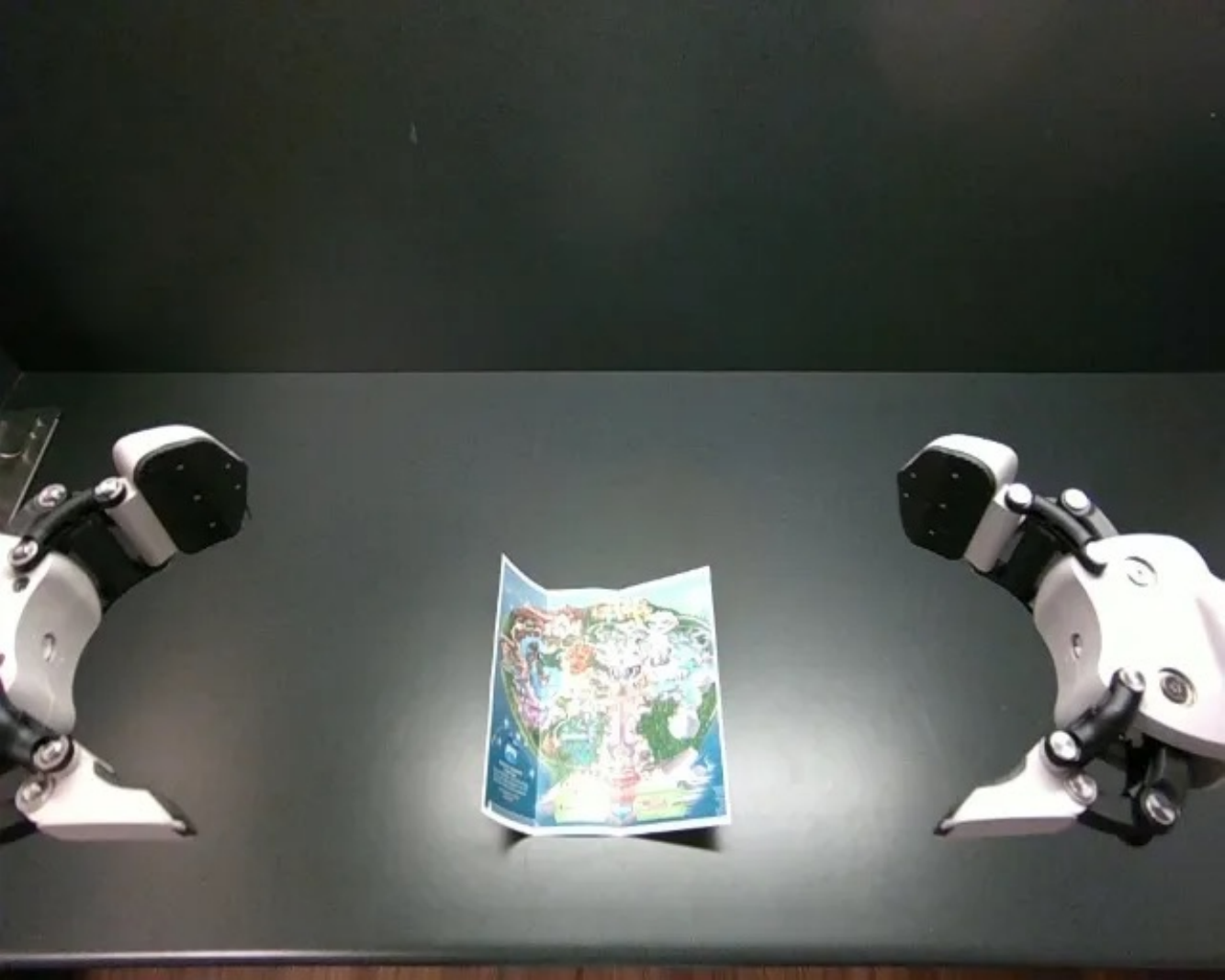} & \centering\scriptsize{The left arm grabs the left side of the map. The right arm folds the right side of the map. The left arm folds the left side of the map.} & \includegraphics[width=3.2cm]{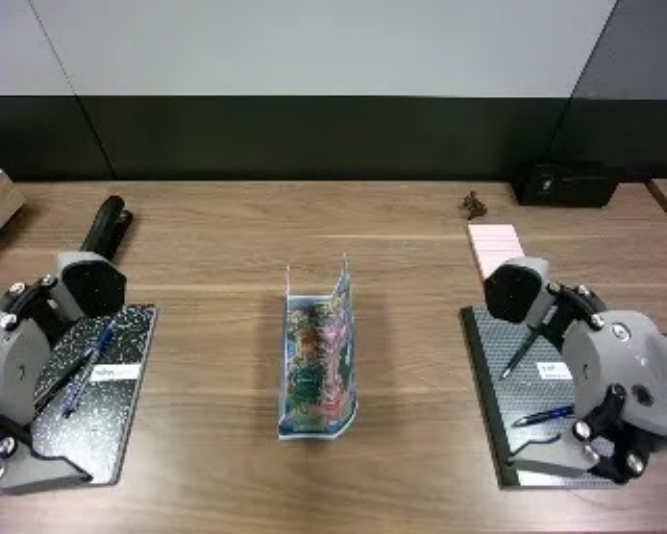} & \centering\scriptsize{The right arm grabs the right side of the map. The left arm folds the left side of the map. The right arm folds the right side of the map.} & \includegraphics[width=3.2cm]{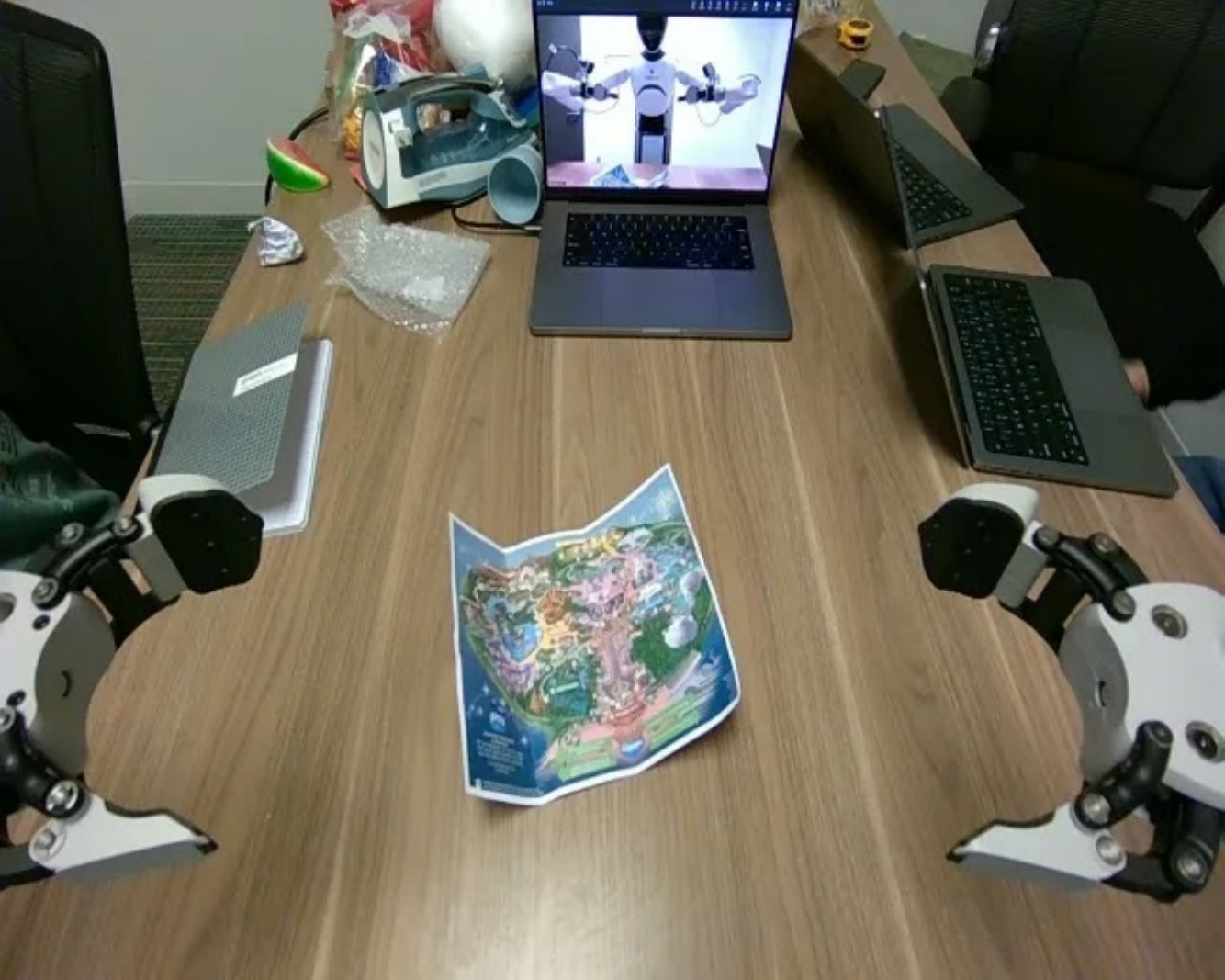} & \centering\arraybackslash\scriptsize{The right arm grabs the right side of the map.The left arm folds the left side of the map. The right arm folds the right side of the map.} \\
\midrule
10 & \centering\textbf{Pulling Cart} & \includegraphics[width=3.2cm]{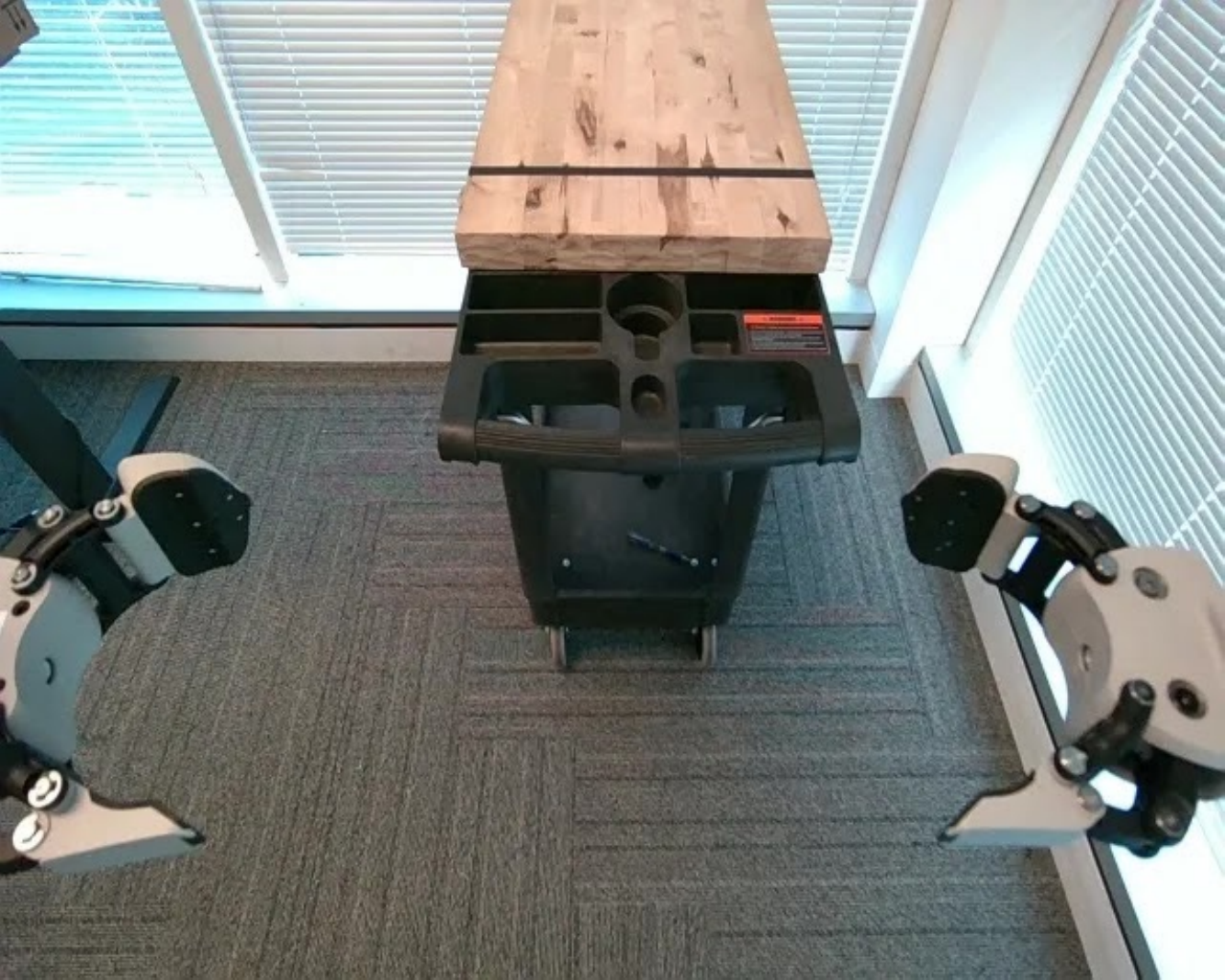} & \centering\scriptsize{The robot reaches its right arm to grasp the cart and pulls it.} & \includegraphics[width=3.2cm]{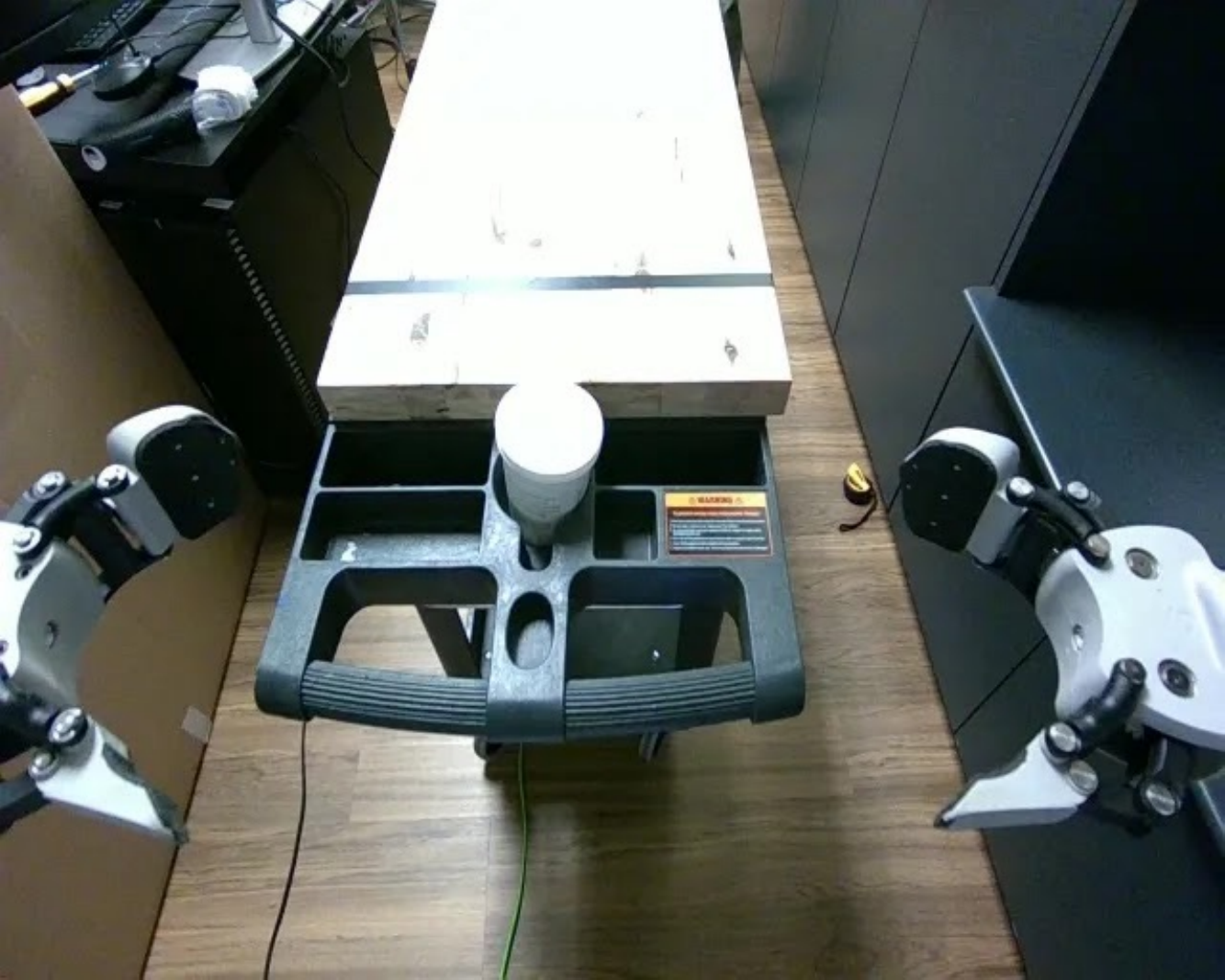} & \centering\scriptsize{The robot reaches its left arm to grasp the cart and pulls it.} & \includegraphics[width=3.2cm]{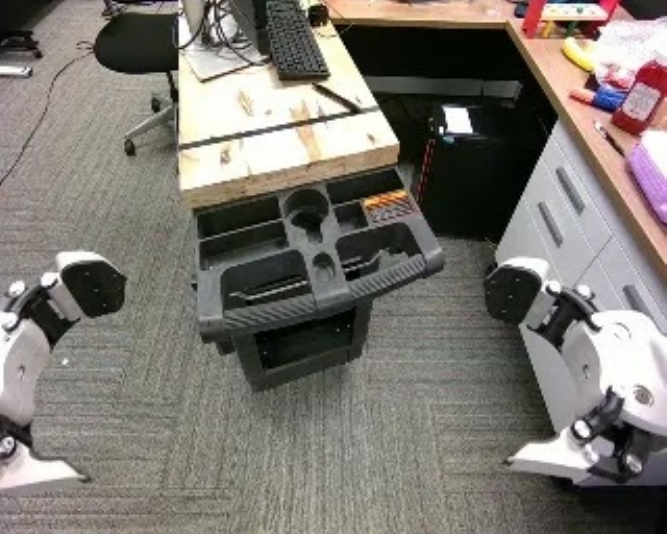} & \centering\scriptsize{The robot reaches its right arm to grasp the cart and pulls it.} & \includegraphics[width=3.2cm]{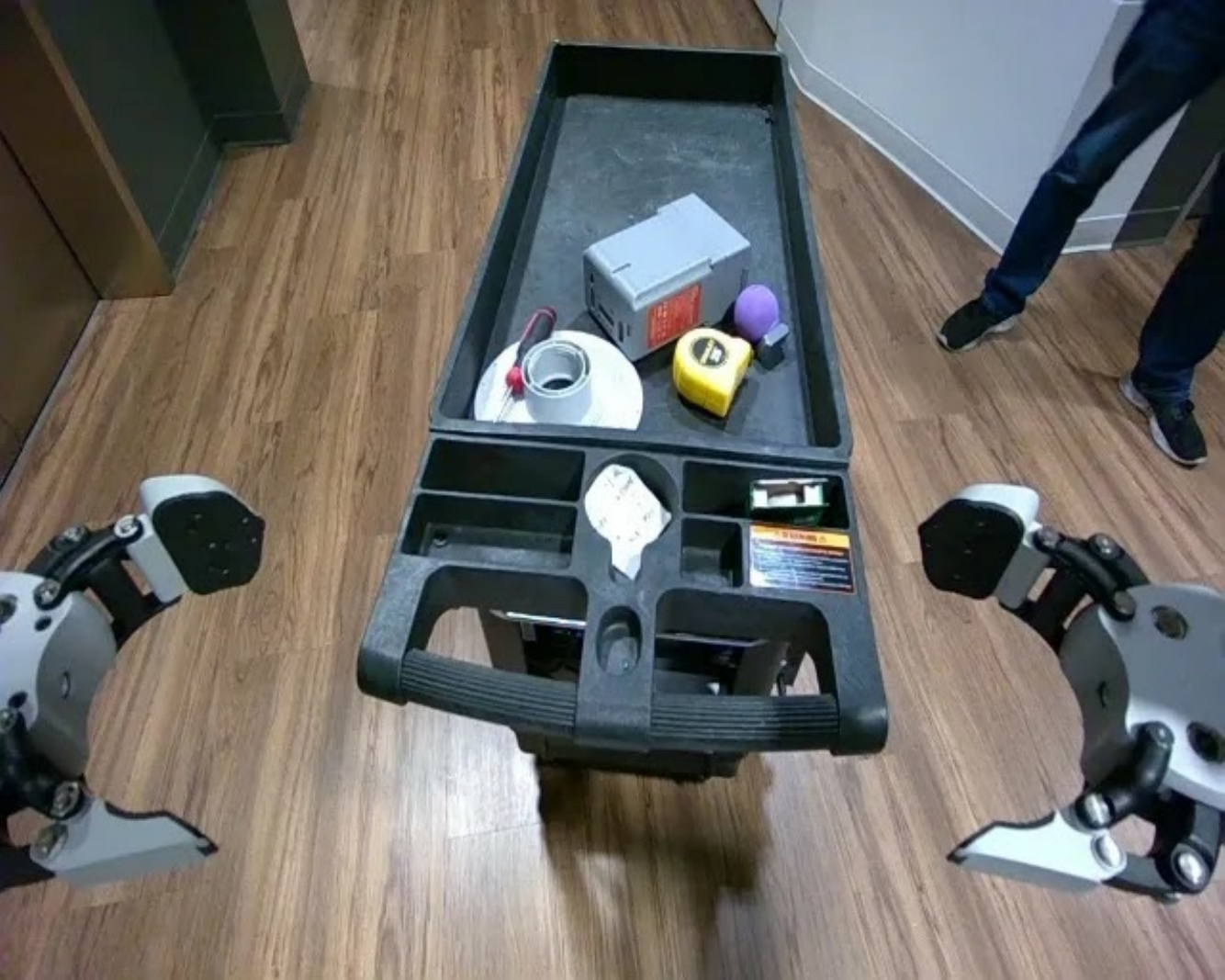} & \centering\arraybackslash\scriptsize{The robot reaches its left arm to grasp the cart and pulls it forward.} \\
\bottomrule
\end{tabular}
}
\end{table*}

\section{DROID Evaluation Details}
\label{appen:droid_evals}
We provide the evaluation setup for both seen and unseen tasks in Tables~\ref{tab:droid_seen} and ~\ref{tab:droid_unseen} for DROID. We conduct 2 rollouts per task by varying the location of the objects.

\begin{table*}[t]
\centering
\setlength{\tabcolsep}{2pt}
\begin{subtable}[t]{0.49\textwidth}
\centering
\caption{\textbf{Seen tasks on DROID}}
\label{tab:droid_seen_tasks}
\resizebox{\linewidth}{!}{%
\begin{tabular}{c m{2.1cm} >{\centering\arraybackslash}m{3.6cm} c m{2.1cm} >{\centering\arraybackslash}m{3.6cm}} 
\toprule
\centering\textbf{\#} & \centering\textbf{Image} & \centering\textbf{Instruction} & \centering\textbf{\#} & \centering\textbf{Image} & \centering\arraybackslash\textbf{Instruction} \\
\midrule
1 & \centering\includegraphics[width=2.1cm]{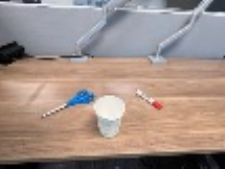} & \scriptsize{Move the cup forward then put the marker inside the cup} & 11 & \centering\includegraphics[width=2.1cm]{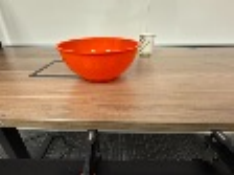} & \scriptsize{Move the bowl on the left to the right side of the table.} \\
\midrule
2 & \centering\includegraphics[width=2.1cm]{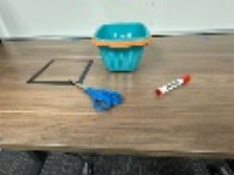} & \scriptsize{Put the marker in the blue box} & 12 & \centering\includegraphics[width=2.1cm]{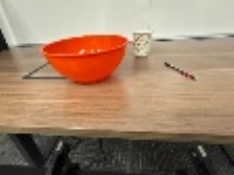} & \scriptsize{Pick up the pencil and put it on the bowl} \\
\midrule
3 & \centering\includegraphics[width=2.1cm]{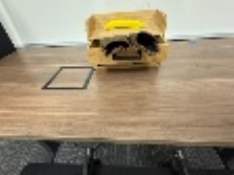} & \scriptsize{Remove the pair of gloves from the open drawer and put it on the table} & 13 & \centering\includegraphics[width=2.1cm]{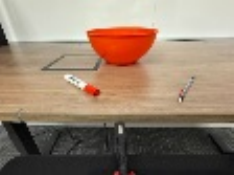} & \scriptsize{Pick the marker up from the table and put it in the bowl} \\
\midrule
4 & \centering\includegraphics[width=2.1cm]{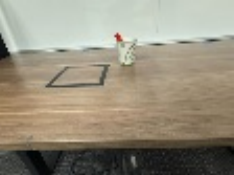} & \scriptsize{Put the marker on table} & 14 & \centering\includegraphics[width=2.1cm]{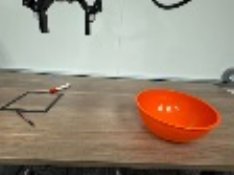} & \scriptsize{Place the bowl next to the marker} \\
\midrule
5 & \centering\includegraphics[width=2.1cm]{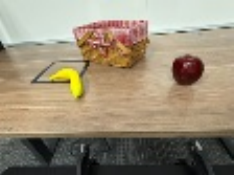} & \scriptsize{Pick up the apple and put it in the basket} & 15 & \centering\includegraphics[width=2.1cm]{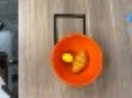} & \scriptsize{Remove a lemon from the bowl} \\
\midrule
6 & \centering\includegraphics[width=2.1cm]{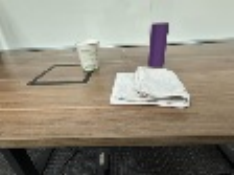} & \scriptsize{Put the towel on the white cup} & 16 & \centering\includegraphics[width=2.1cm]{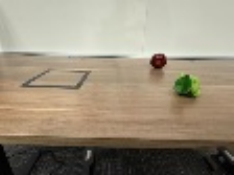} & \scriptsize{Move the grapes to the left} \\
\midrule
7 & \centering\includegraphics[width=2.1cm]{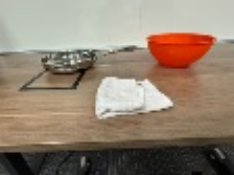} & \scriptsize{Put the towel in the pan} & 17 & \centering\includegraphics[width=2.1cm]{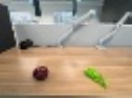} & \scriptsize{Move the green grapes backwards} \\
\midrule
8 & \centering\includegraphics[width=2.1cm]{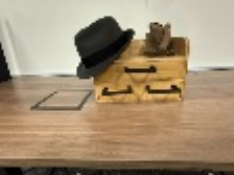} & \scriptsize{Put the hat on the table} & 18 & \centering\includegraphics[width=2.1cm]{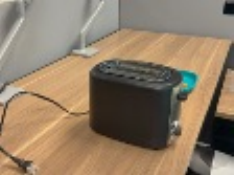} & \scriptsize{Put the bread inside the toaster} \\
\midrule
9 & \centering\includegraphics[width=2.1cm]{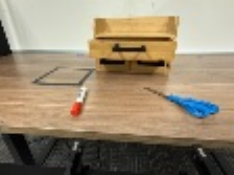} & \scriptsize{Put the pair of scissors into the drawer} & 19 & \centering\includegraphics[width=2.1cm]{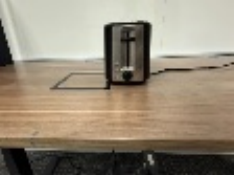} & \scriptsize{Push the lever on the bread toaster downwards} \\
\midrule
10 & \centering\includegraphics[width=2.1cm]{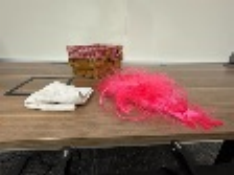} & \scriptsize{Put the towel in the basket} & 20 & \centering\includegraphics[width=2.1cm]{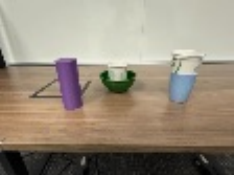} & \scriptsize{Pick up the cup from the bowl and put it in the other cups} \\
\bottomrule
\label{tab:droid_seen}
\end{tabular}
}
\end{subtable}
\hfill
\begin{subtable}[t]{0.49\textwidth}
\centering
\caption{\textbf{Tasks with unseen verbs on DROID}}
\label{tab:droid_unseen_tasks}
\resizebox{\linewidth}{!}{%
\begin{tabular}{c m{2.1cm} >{\centering\arraybackslash}m{3.6cm} c m{2.1cm} >{\centering\arraybackslash}m{3.6cm}} 
\toprule
\centering\textbf{\#} & \centering\textbf{Image} & \centering\textbf{Instruction} & \centering\textbf{\#} & \centering\textbf{Image} & \centering\arraybackslash\textbf{Instruction} \\
\midrule
1 & \centering\includegraphics[width=2.1cm]{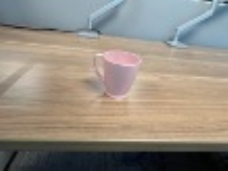} & \scriptsize{\textbf{\underline{Orient}} the mug so the handle is to the right} & 11 & \centering\includegraphics[width=2.1cm]{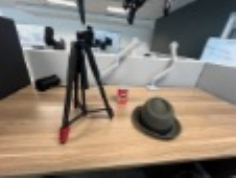} & \scriptsize{\textbf{\underline{Hook}} the hat onto the tripod} \\
\midrule
2 & \centering\includegraphics[width=2.1cm]{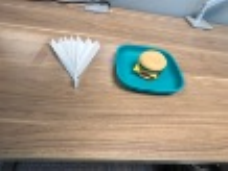} & \scriptsize{\textbf{\underline{Fan}} the burger} & 12 & \centering\includegraphics[width=2.1cm]{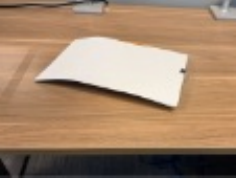} & \scriptsize{\textbf{\underline{Pinch}} the binder clip to release the papers} \\
\midrule
3 & \centering\includegraphics[width=2.1cm]{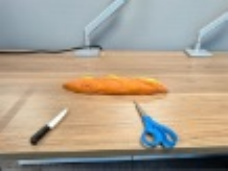} & \scriptsize{\textbf{\underline{Slice}} the bread with the knife} & 13 & \centering\includegraphics[width=2.1cm]{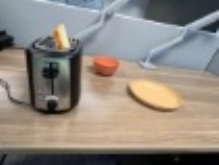} & \scriptsize{\textbf{\underline{Withdraw}} the bread from the toaster and place on the plate} \\
\midrule
4 & \centering\includegraphics[width=2.1cm]{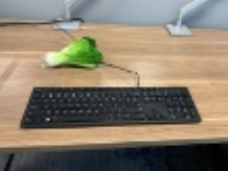} & \scriptsize{\textbf{\underline{Type}} 'hi' on the keyboard} & 14 & \centering\includegraphics[width=2.1cm]{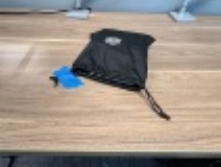} & \scriptsize{\textbf{\underline{Cinch}} the drawstring of the bag} \\
\midrule
5 & \centering\includegraphics[width=2.1cm]{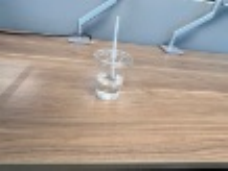} & \scriptsize{\textbf{\underline{Extricate}} the straw from the cup} & 15 & \centering\includegraphics[width=2.1cm]{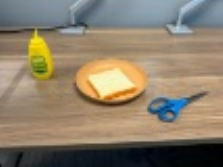} & \scriptsize{\textbf{\underline{Dispense}} the mustard onto the bread} \\
\midrule
6 & \centering\includegraphics[width=2.1cm]{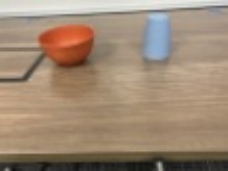} & \scriptsize{\textbf{\underline{Reveal}} the object under the cup} & 16 & \centering\includegraphics[width=2.1cm]{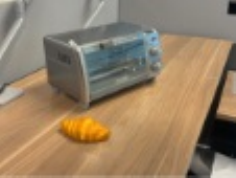} & \scriptsize{\textbf{\underline{Bake}} the croissant in the oven} \\
\midrule
7 & \centering\includegraphics[width=2.1cm]{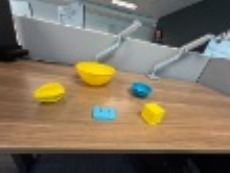} & \scriptsize{\textbf{\underline{Match}} the objects to their corresponding bowl} & 17 & \centering\includegraphics[width=2.1cm]{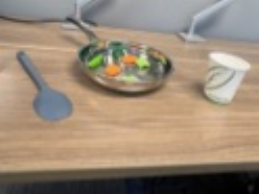} & \scriptsize{\textbf{\underline{Fry}} the vegetables in the pan with the spatula} \\
\midrule
8 & \centering\includegraphics[width=2.1cm]{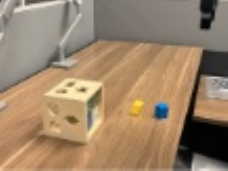} & \scriptsize{\textbf{\underline{Maneuver}} the blocks through the matching hole} & 18 & \centering\includegraphics[width=2.1cm]{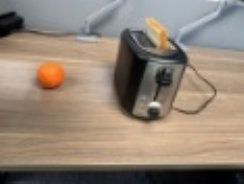} & \scriptsize{\textbf{\underline{Depress}} the lever on the toaster} \\
\midrule
9 & \centering\includegraphics[width=2.1cm]{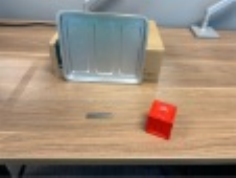} & \scriptsize{\textbf{\underline{Affix}} the magnet to the tray} & 19 & \centering\includegraphics[width=2.1cm]{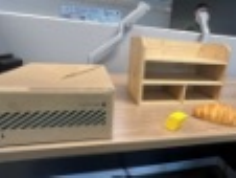} & \scriptsize{\textbf{\underline{Elevate}} the yellow block to the highest platform} \\
\midrule
10 & \centering\includegraphics[width=2.1cm]{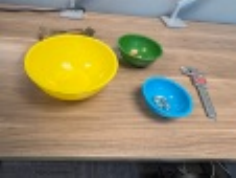} & \scriptsize{\textbf{\underline{Combine}} the nuts and batteries} & 20 & \centering\includegraphics[width=2.1cm]{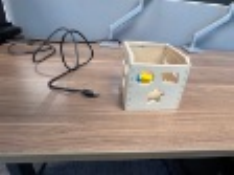} & \scriptsize{\textbf{\underline{Weave}} the wire through the holes of the box} \\
\bottomrule
\label{tab:droid_unseen}
\end{tabular}
}
\end{subtable}
\end{table*}

\section{Failure Case Analysis}
\label{appen:hallucination}
\begin{figure}[htbp]
    \centering
    \includegraphics[width=1\textwidth]{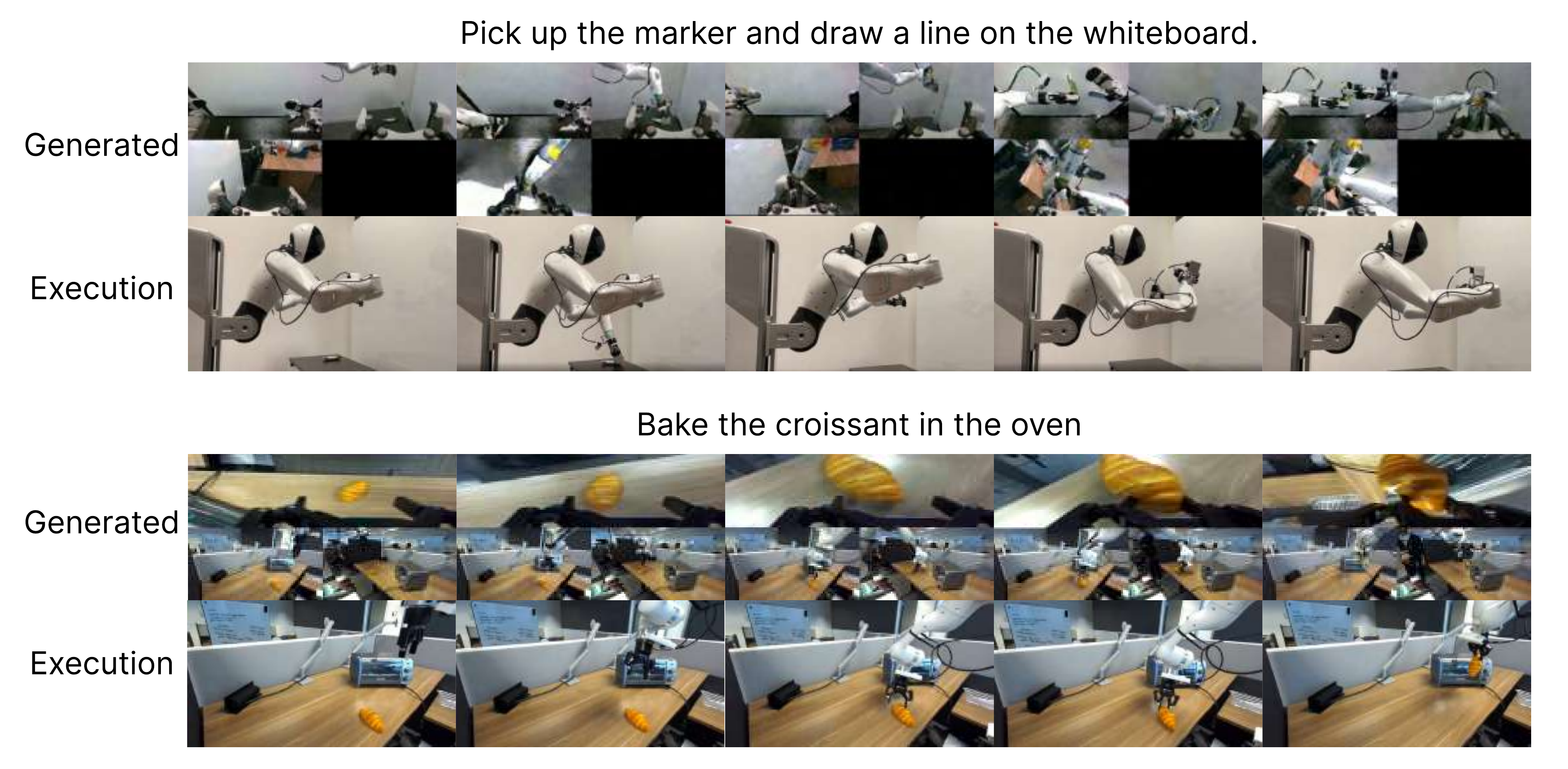}
    \caption{\textbf{Illustration of generated and executed pair.} We illustrate the generated video and action execution pair. These two examples show scenarios where the video prediction \textit{failed} and the robot followed the failed plan. \ourmethod failed to generate video of the Agibot G1 robot drawing in the whiteboard and failed to generate video of the Franka robot opening the oven first.}
    \label{fig:video_alignment}
\end{figure}

In Figure~\ref{fig:video_alignment}, we illustrate the generated video by \ourmethod and execution rollout for both AgiBot and DROID. Overall, the robot execution follows the visual plan generated on the video modality side. For AgiBot video generated by \ourmethod, the robot picks up the marker with left arm and passes the marker to the right arm. Consistent with the generated video, for the execution rollout, the robot picks up the top part of the marker, but instead of drawing a line on the whiteboard, the left arm passes the marker to the right arm. For DROID video generated by \ourmethod, the robot picks up the bread instead of opening the oven first. Aligned with the generated video, for the execution rollout, the robot picks up the bread first instead of opening the oven, leading to the rollout being stuck after the robot has reached to the oven with bread held. This implies that improving the language following and visual planning capability of WAMs could potentially lead to better action execution. 

\clearpage
\setcitestyle{numbers}
\bibliographystyle{plainnat}
\bibliography{main}

\end{document}